\documentclass[runningheads]{llncs}
\usepackage[T1]{fontenc}

\usepackage[a4paper,twoside=false]{geometry} 
 \usepackage{booktabs}
\usepackage{tikz}
\usepackage{tikz-3dplot}
\usetikzlibrary{positioning, calc}

\usepackage{graphicx}
\usepackage{afterpage}

\usepackage[table,dvipsnames]{xcolor}
\usepackage{colortbl}
\usepackage{amsmath,amsfonts}
\usepackage{pifont}
\usepackage{amssymb}
\usepackage{array}
\usepackage{multicol}
\usepackage{multirow}
\usepackage{graphicx}
\usepackage{enumitem}
\usepackage{float}
\usepackage{hyperref}

\usepackage{mfirstuc}

\hypersetup{
    colorlinks=true,
    citecolor=black,
    linkcolor=black,
    urlcolor=blue
}

\definecolor{colcolor}{gray}{0.96}
\colorlet{mixedcolor}{gray!60!ForestGreen!20}
\colorlet{mixedcolorred}{gray!40!BrickRed!15}

\newcolumntype{P}[1]{>{\raggedright\arraybackslash}p{#1}}

\newcommand{\dimname}[1]{{\bf\color{ForestGreen!50!black}\capitalisewords{#1}}}

\newcommand{\tablenote}[2][0.32in]{\par\vspace{-5pt} \makebox[#1]{\tiny #2}\vspace{-3pt}}

\definecolor{provencolor}{RGB}{220,240,220}
\definecolor{partialcolor}{RGB}{255,243,205}
\definecolor{designcolor}{RGB}{220,230,245}
\definecolor{futurecolor}{RGB}{245,220,220}

\newcommand{\statusdem}{{\color{green!60!black}\ding{52}}}    %
\newcommand{\statusdes}{{\color{blue!70!black}\ding{72}}}     %
\newcommand{\statusfut}{{\color{red!70!black}\ding{55}}}      %
\newcommand{\statuspar}{{\color{orange!80!black}\bf$\approx$}}          %

\makeatletter
\renewcommand\subsubsection{%
 \@startsection{subsubsection}{3}{0pt}%
 {2.25ex plus 1ex minus .2ex}%
 {0pt}%
 {\normalfont\bfseries}%
}
\makeatother

\newcommand{\dimension}[3][]{%
 \subsubsection{#2{\normalfont #1}: }%
 #3%
}

\begin{document}
\title{The MMM Data Model\\ \large\normalfont A Normative Specification for Knowledge Interoperability in a Decentralisable Knowledge Commons  }
\titlerunning{The MMM Data Model}
\author{Mathilde Noual\inst{1,2}
\orcidID{0000-0001-7943-6795} 
}
\authorrunning{M. Noual} %
\institute{
    Aix Marseille Univ, CNRS, LIS, Marseille, france
    \and
 Centre Européen de sociologie et de sciences politiques (CESSP)\\ UMR8209. CNRS, Université Paris 1 Panthéon-Sorbonne 
 }
\maketitle 

\begin{abstract}
    Many information systems are built around documents: self-contained units optimised for print production and linear reading. While effective for large-scale dissemination, the document-centric organisation constrains how knowledge can be structured, updated, shared, and reused. Formal approaches address some of these limitations but struggle to achieve widespread contribution and adoption
    due to their prioritisation of formal structure over other system properties such as human usability and scope. 
    AI systems are reshaping document production, but without providing a unified portable alternative to traditional documents  for humans' expression and exchange of knowledge.
    This paper presents MMM, a data model for knowledge documentation that emerged from the practical needs of interdisciplinary collaborative research,  and positioned here within a comparative analysis of the design space of information systems. MMM combines a small set of normative constraints with the expressive freedom of free-text labels. It is designed for interoperability across disciplines, applications and deployments with minimal semantic agreement requirements.
    A reference implementation and pilot deployment data demonstrate implementability and early usability.

     \end{abstract}
\vspace{1cm}

\section{Introduction}

The technologies through which we capture, structure, and share knowledge reflect different design priorities. Some tend to prioritise flexibility and low-friction entry, allowing users to organise information as they see fit -- e.g. personal knowledge management tools like Obsidian and Roam. Some tend to prioritise large-scale collaboration -- e.g. OpenStreetMap and Wikipedia. Some tend to prioritise rigorous structure to enable systematic, fine-grained querying and reuse -- e.g. relational databases and RDF-based systems. These priorities manifest in what each system requires of its users and contributors, what it forbids, and what it especially facilitates. 

In 1945, Vannevar Bush observed a misalignment between prevailing traditional modes of information organisation and how
scientific knowledge needs to be produced and used in practice
\cite{bush1945we}. This misalignment remains relevant in contemporary   day-to-day collaborative interdisciplinary scientific research workflows. The comparative framework developed in Section 2 is motivated by these difficulties, and aims to highlight capabilities and design choices that existing systems have already shown to be achievable in practice, as well as the spaces that remain between them.

We have OpenStreetMap for continuously and collectively curating, at planetary scale, a structured corpus of geographic observations and data valuable to many independent systems. 
We have Wikipedia for collaboratively documenting encyclopedic knowledge in unstructured prose. 
We have Argument Mapping systems for capturing the epistemic structure of certain discussions. 
We have the Semantic Web for representing knowledge in machine-friendly forms that can be exchanged and processed across systems. 
And we have Git for decentralised version control of the documents and code we write together.
In 2026, what, if anything, stands in the way of having an OpenStreetMap-like system for documenting and exchanging ideas and general human knowledge?

Section 3 introduces MMM, a data model motivated by how researchers interact with evolving knowledge in practice, and informed by combinations of capabilities present in existing systems.

\def\adoptionrow{\rowcolor{Blue!10}\cellcolor{Blue!40!black!36}
{\bf Adoption \&  Accessibility} &
    \yes & %
    \yes & %
    \yes &  %
    \limited &  %
    \yes & %
    \yes & %
    \yes & %
    \yes & %
    \nope & %
    \yes  %
    \\\hline\hline
} %
\def\coherencerow{\rowcolor{mixedcolorred}  \cellcolor{mixedcolorred!90!black}{\bf Global Consistency} &
  \nope & %
  \nope & %
  \nope &  %
  \nope & %
  \nope & %
  \nope & %
  \nope & %
  \nope & %
  \cellcolor{mixedcolorred}\yes & %
  \cellcolor{mixedcolorred}\nope %
  \\\hline

}

\def\commonprodrow{\rowcolor{Blue!10}\cellcolor{Blue!40!black!36}
    {\bf Enclosure Resistance} &
    \nope & %
    \nope & %
    \limited &  %
    \nope & %
    \nope & %
    \nope & %
    \yes & %
    \yes & %
    \nope & %
    \yes %
    \\\hline\hline
}

\def\commonrulesrow{\cellcolor{gray!35}
    {\bf Strong Common Rules} &
    \nope & %
    \nope & %
    \nope & %
    \yes & %
    \nope & %
    \nope & %
    \yes & %
    \yes & %
    \yes & %
    \yes  %
    \\\hline
}

\def\contextrow{\cellcolor{gray!35}
    {\bf Contextual Enrichment} &
    \nope & %
    \limited & %
    \nope &  %
    \yes & %
    \limited & %
    \nope & %
    \limited & %
    \yes & %
    \yes & %
    \yes %
    \\\hline

    \cellcolor{gray!35}
    {\bf Knowledge System} &
    \nope & %
    \nope & %
    \nope &  %
    \yes  & %
    \nope & %
    \yes \tablenote{implicit} & %
    \nope & %
    \limited & %
    \yes & %
    \yes %
    \\\hline\hline
}

\def\datalevelsRow{\cellcolor{gray!35}{\bf Homogeneous Space}  &
    \nope & %
    \nope & %
    \nope &  %
    \yes &  %
    \nope & %
    \nope & %
    \nope & %
    \nope & %
    \nope & %
    \nope  %
    \\\hline\hline
}

\def\datamodelrow{\cellcolor{gray!35}
    {\bf Normative Data Model} &
    \nope & %
    \nope & %
    \nope &  %
    \yes &  %
    \nope & %
    \nope & %
    \nope & %
    \yes & %
    \yes & %
    \yes  %
    \\\hline
}

\def\decentralrow{\rowcolor{mixedcolor}\cellcolor{mixedcolor!90!black}  {\bf Write Decentralisability} &
    \nope & %
    \nope & %
    \nope &  %
    \nope & %
    \nope & %
    \nope & %
    \nope & %
    \nope & %
    \limited & %
    \nope %
    \\ \hline}

\def\disagreerow{
   
    \cellcolor{gray!35} {\bf Disagreement } &
    \yes & %
    \yes & %
    \yes &  %
    \yes & %
    \yes & %
    \yes & %
    \yes & %
    \yes & %
    \nope& %
    \yes %
    \\\hline

    \cellcolor{gray!35} {\bf Continual Improvement } &
    \yes & %
    \yes & %
    \yes &  %
    \yes & %
    \yes & %
    \nope & %
    \yes & %
    \yes & %
    \nope& %
    \yes %
    \\\hline
}

\def\emergentrow{\rowcolor{Blue!10}\cellcolor{Blue!40!black!36}
    {\bf Emergent Coll. Benefits} &
    \nope & %
    \nope & %
    \nope & %
    \nope & %
    \yes & %
    \limited & %
    \yes & %
    \yes & %
    \intheory  \vspace{2pt} \tablenote{theory} & %
    \yes %
    \\\hline} %

\def\expressrow{\cellcolor{gray!35}
    {\bf Expression Intent} &
    \yes & %
    \yes & %
    \yes &  %
    \yes & %
    \yes & %
    \yes & %
    \nope & %
    \nope & %
    \nope & %
    \yes  %
    \\\hline
}

\def\humanrow{\cellcolor{gray!35}
    {\bf Human-Primacy} &
    \yes & %
    \yes & %
    \yes &  %
    \yes & %
    \yes & %
    \yes & %
    \yes & %
    \nope & %
    \nope & %
    \yes  %
    \\\hline

}

\def\indivvaluerow{\rowcolor{Blue!10}\cellcolor{Blue!40!black!36} {\bf Immediate \& Local %
} &
    \yes & %
    \yes & %
    \yes &  %
    \yes & %
    \nope & %
    \yes & %
    \limited & %
    \nope & %
    \nope & %
    \yes %
    \\\hline
}

\def\modelrow{ \rowcolor{mixedcolorred} \cellcolor{mixedcolorred!90!black}
    {\bf {Model Intent}} &
    \nope & %
    \nope & %
    \nope &  %
    \nope & %
    \nope & %
    \nope & %
    \yes & %
    \yes & %
    \yes & %
    \yes  %
    \\\hline
}

\def\convergencerow{ \rowcolor{mixedcolorred} \cellcolor{mixedcolorred!90!black}
    {\bf Convergence Intent} &
    \nope & %
    \nope & %
    \nope &  %
    \nope & %
    \nope & %
    \nope & %
    \yes & %
    \yes & %
    \yes & %
    \yes  %
    \\\hline
}

\def\externalKrow{
    \cellcolor{gray!35}{\bf Persistent Production} &
    \yes & %
    \yes & %
    \yes &  %
    \yes & %
    \yes & %
    \nope & %
    \yes & %
    \yes & %
    \yes & %
    \yes  %
    \\\hline
}

\def\persistentRow{\cellcolor{gray!35}
    {\bf Persistent Artefacts} &
    \yes & %
    \yes & %
    \yes &  %
    \yes & %
    \yes & %
    \nope & %
    \yes & %
    \limited & %
    \limited & %
    \yes  %
    \\\hline
}

\def\questionsRow{\cellcolor{gray!35}  {\bf Questions} &
  \yes & %
  \yes & %
  \limited & %
  \nope & %
  \yes & %
  \yes & %
 \nope & %
  \nope & %
  \nope %
  \\\hline
}

\def\ontolcomrow{
  \rowcolor{mixedcolorred}  \cellcolor{mixedcolorred!90!black} {\bf
    Ontological Commitment} &
  \nope & %
  \nope & %
  \nope &  %
  \nope &  %
  \nope & %
  \nope & %
  \nope & %
  \limited & %
  \cellcolor{mixedcolorred}\yes & %
  \cellcolor{mixedcolorred}\nope  %
  \\\hline\hline
}

\def\redundancyRow{\cellcolor{gray!35}{\bf Redundancy-Friendly}  &
    \yes & %
    \yes & %
    \yes &  %
    \yes &  %
    \yes & %
    \yes & %
    \nope & %
    \yes & %
    \nope & %
    \nope  %
    \\\hline
}

\def\collabrow{\rowcolor{Blue!10}\cellcolor{Blue!40!black!36}{\bf Wide-Scale Collab.} &
    \nope & %
    \nope & %
    \yes &  %
    \nope & %
    \yes & %
    \nope & %
    \yes & %
    \yes & %
    \par \intheory \vspace{2pt} \tablenote{experts} & %
    \yes  %
    \\\hline
}
\def\scoperow{
    \cellcolor{gray!35}
    {\bf Universal Scope} &
    \yes & %
    \yes & %
    \yes &  %
    \nope &  %
    \yes & %
    \yes & %
    \nope  \tablenote{encycl.}  & %
    \nope \tablenote[2.5cm]{\hspace{-20pt}propositional} & %
    \nope & %
    \nope  \tablenote{geogr.} %
    \\\hline\hline

}

\def\typingrow{\cellcolor{gray!35}
    {\bf Formal Typing} &
    \nope & %
    \nope & %
    \nope &  %
    \yes & %
    \nope & %
    \nope & %
    \nope & %
    \yes & %
    \yes & %
    \yes %
    \\\hline
} %
\def\explicitlinksrow{\cellcolor{gray!35}
    {\bf 1st-Class Relationships} &
    \nope & %
    \nope & %
    \nope &  %
    \yes  & %
    \nope & %
    \nope & %
    \nope & %
    \limited & %
    \yes & %
    \yes %
    \\\hline

} 
\def\interopRow{
    \rowcolor{mixedcolor}\cellcolor{mixedcolor!90!black} {\bf Interoperability} &
    \nope & %
    \nope & %
    \nope &  %
    \nope & %
    \yes & %
    \nope & %
    \nope & %
    \yes & %
    \yes & %
    \yes %
    \\\hline
}

\def\addressandreuserow{\cellcolor{gray!35}
    {\bf Post-Document
    } &
    \nope & %
    \nope & %
    \nope &  %
    \yes & %
    \nope & %
    \nope & %
    \nope & %
    \yes & %
    \yes & %
    \yes  %
    \\\hline
}

\section{Related Work and Positioning}
\label{framework}

This section provides a provisional contextual basis for understanding the MMM data model presented in Section \ref{mmm}. It compares a range of information systems and representational frameworks such as Personal Knowledge Management tools (PKM, e.g. Obsidian, Roam, Notion, Logseq, SiYuan), Argument Mapping systems (e.g. Argdown, Arguman, Compendium, DebateGraph, DebateMap, Kialo, Rationale) \cite{chesnevar2006aif,ibis1970,scheuer2010argument},  the World Wide Web (WWW),  Large Language Models (LLMs) \cite{zhao2023survey}, OpenStreetMap (OSM) and its broader ecosystem \cite{haklay2008osm}, Wikipedia (not reduced to Wikidata) \cite{Wikipedia}, knowledge graphs like Wikidata,  and  RDF/OWL \cite{RDF11,OWL2W}  -- all broadly construed as systems for the production, organisation, and exchange of information.
The selection of systems is illustrative with no intention of being exhaustive.
Protocols (e.g. ActivityPub, IPFS, Solid) that provide technical infrastructure (storage, messaging, access control) delegating knowledge-level concerns to applications built on top of them, are not considered.

Selected systems are informally examined
along a set of dimensions (cf \S \ref{dimensions})
reflecting a particular point of view -- one motivated by the needs of collaborative scientific research.
The dimensions are not mutually independent: some tend to co-occur, others are near opposites.
Some concern the intrinsic properties of a format or data model specification, reflecting architectural design choices. Others concern the intended use or aspirations of a deployment. And others depend on contingent social conditions, governance structures, and community practices that develop around a system.
Dimensions are defined as  binary properties to facilitate comparison.
In reality, they rather correspond to continuous spectra along which information systems may occupy intermediate positions.
The positions assigned in Table~\ref{taxonomytable} are therefore  deliberate simplifications. Each system has been placed at the end of the spectrum that most closely reflects its dominant characteristics.

\subsection{Dimensions} 
\label{dimensions}

\afterpage{\clearpage
\newcommand{\yes}{\textcolor{ForestGreen}{\ding{52}}}
\newcommand{\intheory}{\textcolor{BrickRed}{\ding{52}}}
\newcommand{\bydesign}{\textcolor{YellowOrange}{\ding{52}}}
\newcommand{\wanted}{\textcolor{YellowOrange}{\ding{52}}}
\newcommand{\nope}{\textcolor{BrickRed}{\ding{56}}}
\newcommand{\nearnope}{\textcolor{BrickRed}{(\ding{56})}}
\newcommand{\limited}{\textcolor{ForestGreen}{(\ding{52})}}

\newgeometry{left=4cm,right=3.6cm} 

\begin{table}[htbp]
    \centering
    \setlength{\parskip}{-4pt}
    \setlength{\extrarowheight}{0pt}
    \renewcommand{\arraystretch}{1.5}%
        \begin{tabular}{|>{\columncolor{gray!20}\bfseries\arraybackslash\vspace{2pt}}m{1.61in}|
            >{\centering}p{0.31in}| %
            >{\setlength{\parskip}{-4pt}\centering}p{0.32in}| %
            >{\setlength{\parskip}{-4pt}\centering}p{0.32in}| %
            >{\setlength{\parskip}{-4pt}\centering}p{0.32in}| %
            >{\centering}p{0.32in}| %
            >{\centering}p{0.32in}| %
            >{\centering}p{0.32in}| %
            >{\centering}p{0.32in}| %
            >{\setlength{\parskip}{-4pt}\centering}p{0.32in}| %
            >{\centering\arraybackslash}p{0.31in}| %
            }
            \rowcolor{gray!25} \hline
            \multicolumn{1}{|>{\cellcolor{gray!35}\bfseries}m{1.61in}|}{Dimension} &
            \multicolumn{1}{m{0.31in}|}{\centering Docs}                           &
            \multicolumn{1}{m{0.32in}|}{\centering PKM}                            &
            \multicolumn{1}{m{0.32in}|}{\centering Wiki}                           &
            \multicolumn{1}{m{0.32in}|}{\centering Arg\par Map}                    &
            \multicolumn{1}{m{0.32in}|}{\centering \scalebox{0.75}[1]{WWW}}        &
            \multicolumn{1}{m{0.32in}|}{\centering \scalebox{0.95}[1]{LLMs}}       &
            \multicolumn{1}{m{0.32in}|}{\centering Wiki-\par pedia}                &
            \multicolumn{1}{m{0.32in}|}{\centering Wiki-\par data}                 &
            \multicolumn{1}{m{0.32in}|}{\centering RDF/\par OWL}                   &
            \multicolumn{1}{m{0.31in}|}{\centering OSM}
            \\
            \hline
            \hline
            \scoperow

            \humanrow
            \expressrow
            \redundancyRow
            \disagreerow
            \datalevelsRow
            \convergencerow

            \coherencerow
            \ontolcomrow

            \indivvaluerow
            \collabrow
            \commonrulesrow

            \emergentrow
            \externalKrow
            \commonprodrow

            \datamodelrow
            \typingrow
            \addressandreuserow
            \explicitlinksrow
            \contextrow
            \adoptionrow

            \decentralrow
            \interopRow
        \end{tabular}
        \smallskip
        \caption{{\bf An exploration of the design space of information systems.} Rows refer to dimensions introduced in  \S\ref{dimensions}. Satisfaction indicators (\yes = fully satisfies, \limited = partially satisfies, \nope = does not satisfy) are indicative rather than definitive. Given the intentionally informal definitions, reasonable disagreement on individual cells is expected. Docs stands for Documents, OSM for OpenStreetMap, PKM for Personal Knowledge Management tools, and Arg Map for Argument Mapping tools. 
        RDF/OWL represents the W3C semantic web standards stack as predominantly deployed in practice.
          }
        \label{taxonomytable}

\end{table}
\clearpage
\restoregeometry

}

\dimension{Universal Scope}{Whether the system imposes no {\it a priori} restriction on the topics, epistemic forms, or degrees of formalisation that can be expressed, and thus allows to {naturally} document established knowledge, knowledge in the making, mathematical definitions, empirical observations, partial and tentative understanding, open questions, drafts, half-formed intuitions, and standalone claims. 
Arguably, only documentation systems supporting free text contribution satisfy this dimension.}

\dimension{Human-Primacy}{Whether the system is designed primarily for humans to write and read information in natural language or familiar notations, rather than requiring contributions to be expressed in formal, machine-oriented representations for machine processing, as in commonsense reasoning systems \cite{davis2015commonsense} and formal knowledge representation frameworks such as RDF/OWL \cite{semwebhard,OWL2W}  and knowledge graph systems \cite{hogan2021knowledge}. Argument Mapping, Wikipedia, and OSM provide examples of systems satisfying this dimension without satisfying the previous.  }

\dimension{Expression Intent}{Whether the system is primarily designed as a space for contributors to {\it express} their own possibly subjective or situated ideas, observations, viewpoints, interpretations  -- rather than requiring contributions to conform to a neutral, encyclopedic, or authoritative standard of truth.                Despite its aspiration to model an accurate map of the world, OSM arguably satisfies this property: it encourages contributors to align with existing tagging conventions and consider existing data, but it prioritises the direct expression of local observations, e.g. of  contributor mapping their neighbourhood. Interestingly, OSM also satisfies the next dimension.}

\dimension{Convergence Intent}{Whether the system aims to produce a single authoritative, consensual, or accurate account of its subject domain, treating conflicting contributions as (temporary) conditions requiring resolution rather than as permanently coexisting perspectives. Systems without convergence intent may leave contradictions unresolved because no convergent target state exists that can be undermined by them.}

This dimension is about the system's end goal (which may  differ across deployments and governances, cf Wikidata, RDF/OWL).
Wikipedia aims to produce an account of something external, namely, established knowledge. This is different from documenting the internal process of knowledge production or of argumentative discourse itself (as in the corpus of scientific publications and  Argument Maps, cf e.g. \cite{jeuris2020socratreesexploringdesignargument}).
In Wikipedia (and Wikidata likewise), this exclusion of knowledge in the making follows directly from \dimname{Convergence Intent} and explains \dimname{Universal Scope} failure. 
Argument Mapping systems, who  stand as a clear counter-example to \dimname{convergence intent}, actively embracing conflicts, illustrate that  this is  however  not the only reason to fail \dimname{Universal Scope}.
 
In systems with \dimname{convergence intent}, what counts as a conflict is determined by the kind of \dimname{convergence intent} the system seeks.
A common kind of convergence is semantic. In systems that have {\bf Semantic Convergence Intent}, contributions about the same subject are expected to converge toward a single agreed-upon meaning or interpretation. Contributors are expected not merely to produce structurally compatible data, but   to mean the same things by their terms, using concepts, entities, and relations  consistently and in alignment with a shared understanding maintained across the contributor community, and, sometimes, across independently governed deployments.
By contrast, the shared structural vocabulary of Argument Mapping systems (e.g. claim, pro-argument, con-argument)  functions as an organising layer for the map and is not part of the account it represents.

A stronger form of \dimname{convergence intent}, implying semantic convergence, is {\bf Model Intent} satisfied by systems  designed to model an external reality (e.g. an accurate map of the world as with OSM, an accurate formal representation of gene functions as with the Gene Ontology) or fiction (e.g. a film fandom wiki) or an established consensus of facts (e.g.  encyclopedic knowledge in Wikipedia).
Contributions are then evaluated for accuracy  against a referent  external to the contributor community (e.g. external sources, executed code behaviour).

Systems with \dimname{Semantic Convergence Intent} usually  have {\bf Canonical Intent}: they expect or impose   a single preferred representation of each entity or concept they cover, with a tendency to actively deduplicate, merge, or redirect equivalent contributions toward a single authoritative one. For instance,   even if multiple Wikipedia articles can cover related topics from different angles (this reflects editorial granularity in deciding what counts as a distinct topic), at the level of articles, within the scope of a given topic, Wikipedia   enforces a one-canonical-article-per-topic principle.

\dimension{Redundancy-Friendliness {\normalfont (Support for Epistemically Productive Redundancy)}}{Whether the system preserves multiple expressions of equivalent knowledge, entities, or perspectives as potentially useful rather than treating redundancy as a defect to eliminate through normalisation or deduplication.   This often implies  a trade-off: accepting ambiguity   in order to gain inclusiveness, accessibility, and usability across different communities. 
\dimname{Redundancy-friendliness} thus tends to be in tension with \dimname{convergence intent}. However, a system can satisfy both and possibly thereby mitigate the trade-off (e.g. Wikidata and SKOS). }

\dimension{Expression of Disagreement}{Whether epistemic disagreement is a normal and expected form of contribution within the system's knowledge or associated discussion space (e.g. Argument Maps, Wikipedia talk pages). This tends to imply \dimname{Human Primacy}.
But the content domain or purpose of some human friendly systems such as library records or crowd-sourced and citizen science observation reporting apps (e.g.  weather, pollution, biodiversity) exclude  disagreement. }
 
\dimension{Support for Continual Improvement} {Whether the system tolerates imperfections and is designed to support the progressive improvement of its knowledge   over time, providing mechanisms for contributors to revisit, correct, extend, or supersede prior work incrementally and at low cost -- rather than treating contributions as fixed, final deposits, and imperfections as bugs.   This dimension is about the lifecycle of contributions: whether the system is built around the assumption that contributions will need to improve, and provides a socially-mediated framework to surface, identify, and address imperfections through continuous community effort. }
\medskip

The preceding dimensions -- \dimname{Expression Intent}, \dimname{Convergence Intent}, \dimname{Redundancy Friendliness}, \dimname{Accommodation of Disagreement} and \dimname{Continual Improvement} -- apply differently depending on whether a system maintains a strict separation between  a clean  knowledge  layer and a discussion layer. A system may have an {expression intent}, be redundancy friendly and support disagreement  in its discussion layer (e.g. Wikipedia's talk pages) while enforcing convergence and canonical representation in its official destination   layer (e.g. Wikipedia's articles).

\dimension{Homogeneous Data Space {\normalfont (First-Class Discussion / Work Space)}}{
Whether knowledge contributions and discussion contributions coexist within a single contribution space as contributions of the same kind, rather than being separated into content and discussion layers. Traditional document systems like Google Docs, as well as PKM tools  tend to fail this property: comments, tracked changes, and review annotations   are usually structurally and  presentationally distinct from the document content or note they concern. A comment is not a document contribution of the same kind as the text it annotates. }

\dimension{High Ontological Commitment}{Whether contributors must rely on a predefined ontology or schema specifying what kinds of entities and relationships exist, rather than introducing new kinds during contribution.  This is one formal, top-down approach  among others toward \dimname{Semantic Convergence Intent}. 
Knowledge graph systems vary on this dimension, from Freebase at the low end of commitment, Wikidata in the middle, to typical RDF/OWL deployments at the high end, including biomedical ontologies.}
 
Since meaningful additions  and improvement must either fit within an existing vocabulary or trigger costly structural change,  \dimname{High ontological commitment} tends to exclude lightweight, incremental, and tentative   contributions.   It also narrows the space of expressible disagreement to what can be expressed in the terms provided by the predefined vocabulary.
\medskip

\dimension{Global Formal Consistency  {\normalfont (Contribution Dependence)}}{Whether formal constraints apply across the entire system such that the validity or consequences of a contribution depend on the state of the system as a whole rather than on local conditions alone. Contradictions are either prevented from entering the system through constraint enforcement (e.g. relational databases) or they are tolerated global effects that compromise intended functionality until resolved (e.g. automated reasoning over inconsistent OWL ontologies), even if the knowledge base as a store remains readable and queryable.  This deployment property requires both  
(i)~formal constraints or axioms declared in advance within the system  that define what counts as a contradiction before any contribution is made,  and (ii)~the formal mechanism that enforces them. }

While \dimname{High Ontological Commitment} raises the cognitive entry cost of contribution, \dimname{Global Formal Consistency} raises the computational cost of validating updates. Each change must be checked for compatibility against the global state of the knowledge base. As the knowledge base grows, this increases the cost of maintaining consistency, and may discourage casual, partial, or exploratory contributions, and
concentrate curation among users able to navigate the formal requirements.
\medskip

\dimname{Global Formal Consistency} separates systems that formally treat contradictions as system-level failures from those that do not. \dimname{Convergence Intent} separates systems that treat contradictions as local quality problems to be resolved (OSM, Wikipedia, Wikidata) from systems that tolerate them permanently as normal and informative (Argument Mapping, the WWW, documents, personal knowledge management).
\medskip

\dimension{Immediate Individual or Local Value}{ Whether a single user obtains clear, direct benefit from the system from day one  -- independently of any other user's participation  -- such that the system is worth adopting for personal use before any network or community exists, as is typically the case with personal knowledge management, reference management (e.g. Zotero) and note-taking tools. When the system supports a shared resource like OSM's map, contributing to it must serve the contributor's own local interests (e.g. getting their own neighbourhood charted).  Wikipedia satisfies this property for readers but only marginally for contributors \cite{teblunthuis2018revisiting}.  }

\dimension{Wide-Scale Collaborative Usage}{Whether the system is designed to coordinate contributions over time from large numbers of participants across geographic, institutional and community boundaries, typically resulting in worldwide-scale deployment and the accumulation of a shared growing corpus (e.g. the WWW, Wikipedia, OSM, Freebase and Wikidata \cite{bollacker2008freebase,wikidata}), as opposed to systems designed primarily for individual or small-team use typically operating at a small scale (e.g. PKM).}

\dimension{Common Contribution Rules}{Whether the system defines explicit system-level rules governing contribution structure, validity, coordination, epistemic standards  (e.g., neutrality, verifiability, original research policies), and dispute resolution, rather than leaving these entirely to local convention.    Wikipedia, OpenStreetMap and Argument Mapping tools impose such frameworks, but they differ in implementation. Wikipedia uses natural-language rules. Both OSM and Argument Mapping tools encode part of their code of conduct into their data model or schema.  }

\dimension{Emergent Collective Benefits}{Whether the system produces a qualitatively different thing that couldn't exist without collective production, some emergent value that no individual could produce alone, and whether this emergence requires scale to materialise. For instance, an OpenStreetMap contributor may find some immediate personal benefit from mapping their local neighborhood but the map's primary value materialises when enough of the surrounding world has been mapped by enough other contributors, enabling continuous navigation from place to place. Open  knowledge graphs (e.g., Wikidata \cite{wikidata}) also satisfy this property, for instance by offering semantic search at scale, as do  Large language models,   albeit as black boxes whose emergent properties don't always serve the  collective benefit \cite{bommasani2022opportunitiesrisksfoundationmodels,huang2023generativeaidigitalcommons,krakauer2025emergence,noroozian2025generative,wei2022emergent}. 
The Semantic Web was also designed around this dimension (federated queries across independently published datasets), but in practice adoption remains confined to individual organisations, rather than a collective commons \cite{BernersLeeVision,semwebadoption,hogan2021knowledge,SemwebVision}.   }

\dimension{Persistent Knowledge Production Site}{Whether the system serves as a site of original knowledge production and documentation,  producing a new lasting organised knowledge resource in its own right, that accumulates and organises contributions as durable, referenceable artefacts that can be built upon over time -- as opposed to systems like search engines existing primarily to be queried (rather than contributed to), whose value proposition primarily depend on consuming (retrieving, ranking, synthesising) knowledge produced and documented elsewhere and systems generating outputs that are typically ephemeral, session- or system-bound.}

\dimension{Enclosure Resistance}{Whether the system incorporates mechanisms (e.g.  open content licensing or public institution governance) that prevent accumulated contributions from being unilaterally enclosed, ensuring the continued existence of openly accessible deployments and shared resources \cite{benkler2006commons,lindemann2025chatbots,ostrom1990governing,noroozian2025generative,del2024large,sun2026aiimprovesanswersslows}, even if private deployments remain possible. } 

\dimension{Normative Data Model}{Whether the system relies on a formal, explicit data model that normatively constrains how data is structured across all instances, independently of any specific deployment or user configuration -- such that information structure is enforced by the data model itself, as opposed to being left to UI conventions, user behaviour, or per-workspace configuration, or to presentation-oriented formats such as HTML or Markdown -- with most or all content being natively expressed in that data model. }

\dimension{Formal Typing}{Whether units of {\it information} documented in the system can be qualified or categorised  into either semantic types (indicating what kind of worldly entity a unit represents, as in knowledge graph class membership) or epistemic roles (declaring what kind of contribution a unit constitutes, such as OSM's distinction between nodes, ways, and relations, and Argument Mapping distinctions between claims, premises and objections).  These type annotations should be formally expected if not predefined by the data model specification. They should be embedded in the data itself as a first-class property of every data unit, making them portable across any conforming implementation of the data model which can then read and interpret them independently of any specific deployment's conventions -- as opposed to being confined to a specific tool's interface, workspace configuration, or user convention. }

\dimension[ (Fine-Grained Addressability \& Reuse Without Copy-Paste)]{Post-document Organisation}{Whether the system's primary unit of organisation is the individual {\it information} unit (e.g. a specific concept, question or argument) conceived as a semantic entity -- as opposed to the document, section, or {\it text} block; and whether those units are independently addressable by stable, persistent identifiers and reusable across multiple contexts, tools, workspaces, by reference rather than by copy.  Following Nelson's principle of transclusion \cite{derose1997problems,nelson1981literary}  -- though applied here to semantic units rather than text segments  -- reference-based reuse facilitates information maintenance, ensuring updates propagate and avoiding WWW-style meaning drift and fragmentation \cite{zhu2025linkrot,zittrain2021paper}.}

\dimension{First-class Relationships}{ Whether the connections between information units are themselves first-class information units satisfying the previous two dimensions as any other information unit in the system -- as opposed to being mere pointers, associative or navigational devices such as hyperlinks or Luhmann's Zettelkasten cross-references \cite{Zettelkasten} as implemented in most PKM tools. }
\medskip

From the previous four dimensions follows {\bf fine-grain queryability} of content within the system (via a standardised query engine, simple key-value lookup, or graph traversal algorithm): the system supports precise, structured queries over individual information units  -- beyond full-text search  -- enabling reliable and reproducible retrieval of units matching any combination of specified criteria (type, content, type of relationship to other units, etc.), with results determined deterministically by the data model,
rather than by user convention.
\medskip

Systems that satisfy the previous four dimensions also provide a basis to support   {\bf redundancy reduction} mechanisms. Contributions, being addressable and queryable,
facilitates the detecting of equivalent contributions, asserting equivalence through typed edges, and filtering redundant results.

\dimension{Contextual Enrichment}{Whether the system is primarily designed so that contributions (information units) gain structured meaning from explicit relationships to other contributions. Systems where links are not first-class information units (e.g., the WWW, Wikipedia, PKM tools) satisfy this property only marginally.}

\dimension{Epistemic Structure {\normalfont (Knowledge System Organisation)}}{Whether the system represents the epistemic structure of knowledge, making it navigable, and making individual knowledge contributions recombinable,  rather than merely accumulating, and retrieving contributions.  A knowledge system is organised around knowledge acts -- distinguishing claims, questions, arguments, perspectives, and subjects --  rather than undifferentiated text or documents.   It represents or   exposes relationships between them (disagreement, support, refinement, restatement, exemplification), and makes those relationships available for navigation, querying, comparison, and recombination.}

Satisfying \dimname{Post-Document Organisation} and \dimname{First-Class Relationships} is one way to satisfy \dimname{Knowledge System Organisation} -- the explicit structural way. LLMs represent a fundamentally different, implicit statistical way.

\dimension{Adoption Ease {\normalfont (Low Contribution Barrier)}}{Whether users can begin contributing effectively without specialised training in formal knowledge representation, either because complexity is hidden by tools (a.g. Notion, Roam,  GitHub, VS Code) or because the data model aligns with ordinary human intuitions (e.g. Markdown,  OpenStreetMap).} 

\dimension{Content-Based Write Decentralisability}{Whether the system's content can
  reside and be maintained across multiple independent locations, exchanged via any transport mechanism including offline channels,   and concurrent contributions can be merged for the significant majority of natural use cases,  based on   content alone -- independently of contributor   identification,   role,  trust, and   commitment --  without requiring a shared platform, central broker, or coordination infrastructure.  This requires:  (1) a shared normative data model ensuring contributions from different sources are structurally compatible, (2) stable portable identifiers ensuring contributions with the same identifier are recognised as the same contribution, and (3) a merge mechanism that handles the significant majority of natural use cases without human coordination, either (3a) because union-based merging suffices  for typical cases, with edge cases possibly remaining unresolved (when contributions are additive by nature or by data model design, e.g. RDF/OWL without constraining axioms, under the open world assumption),  and their union produces a coherent, navigable, cumulatively improving whole
   -- or (3b) because a deterministic mechanism handles all conflict cases (e.g. CRDTs \cite{shapiro2011crdt}, Git-like version control). }
\medskip

\dimname{Convergence intent}'s conflict intolerance tends to make decentralisation difficult.  Git's  \dimname{syntactic Convergence Intent} is nonetheless compatible  with decentralisation  for most non-conflicting commits. But once conflicting edits require human judgement, additional coordination becomes necessary (cf the Linux kernel's hierarchical maintainer structure). \dimname{Semantic Convergence Intent} intensifies this tension by adding the problem of {\it semantic} conflicts. And \dimname{Global Formal Consistency} raises the difficulty further since merging individually consistent datasets may itself produce a globally inconsistent result. When the domain is narrow and stable enough to be fully specified in advance through a fixed vocabulary (e.g. Schema.org, Gene Ontology), the semantic convergence can be built into a frozen semantic frame within which contributors operate  without any on-going agreement to entertain. But as scope broadens, unanticipated cases emerge and meanings may need to be  revised over time.
As participation scales,  meaning risks drifting across the contributor community, and semantic conflicts risk multiplying.
Maintaining semantic convergence therefore requires negotiation of meaning to be systematically supported as a load-bearing part of the contribution process.
Consequently, for any broad-scope system    seeking to maintain shared meaning at scale, fully decentralising the contribution work requires write decentralisability not only of the primary contribution space but also of a discussion space  in which contributors can  express their interpretations and disagreements in view of negotiating and refining meanings (\dimname{Human Primacy}, \dimname{Expression of Disagreement}, \dimname{Expression Intent}, and \dimname{Continual Improvement}). In the case of a \dimname{Homogenous Data Space}, these two spaces coincide.
The linked open data landscape \cite{bizer2009linkeddata,semwebhard} confirms that  relaxing semantic constraints alone is not enough to save \dimname{write decentralisability} under \dimname{Semantic Convergence Intent} without a decentralised \dimname{First-class Discussion Space}.  Most knowledge graphs ended up centralised  (e.g., DBpedia, Wikidata, Freebase, Google Knowledge Graph and other enterprise
Knowledge Graphs, RDF/OWL biomedical ontologies)  and   attempts at decentralising them 
tend to rely on trust, provenance, and cryptographic identity rather than deterministic content-based merge \cite{underlay,originTrail}. 

OSM illustrates another reason to sacrifice \dimname{write decentralisability}: ensuring the data remains accessible under an open license
(\dimname{Enclosure Resistance}), while maintaining read decentralisability, which is the essential part in OSM's case given the  heavy usage of its data by downstream applications \cite{haklay2008osm,mooney2012characteristics,osm_api}.

\dimension{Interoperability}{Whether multiple independent applications serving functionally distinct purposes  -- rather than merely competing implementations of the same application type -- can share, parse, process, combine and possibly merge each other's data, {\it without pairwise conversion}, based on  a common normative data model and stable identifiers, whether through direct file exchange or via a shared infrastructure (e.g. APIs, central servers).}

\dimname{Write decentralisability}, as defined above, provides the basis for   interoperability, but is not necessary: Wikidata provides  interoperability despite centrally coordinating all writes.

A more demanding form of interoperability than the basic, structural form, is {\bf Semantic Interoperability} requiring independent applications to agree on the meaning of the data they share a structure for.  Its weakest, widely achieved form  is {\bf Terminological interoperability}:  independently developed systems use the same symbols to refer to the same entities, concepts, and properties (e.g. Wikidata, SKOS, Schema.org, Dublin Core,  ORCID, DOI, ISO standard codes). One application using data from two separate applications  can recognise when they both refer to the same thing. This requires some shared definition of vocabulary terms, whether through a controlled vocabulary, a SKOS concept scheme, or even a well-maintained natural-language specification (low or moderate ontological commitment). A stronger, more demanding form is {\bf Inferential Interoperability}, the original ambition of the Semantic Web, requiring  both \dimname{High Ontological Commitment} and \dimname{Global Formal Consistency} \cite{BaaderDL,rdfsemantics}.  The Semantic Web sought to achieve this across a global, decentralised community of contributors.
But \dimname{Semantic Interoperability} (whether  terminological or inferential) inherits the decentralisability limitations of \dimname{Semantic Convergence Intent}.  In practice, interoperability has been achieved by relaxing at least one of four constraints: narrowing the data domain, settling for structural or terminological rather than inferential interoperability, restricting the contributor community, or abandoning decentralisation altogether.

\subsection{The Design Space}

\begin{figure}[htbp]
  \centering
  \includegraphics[width=0.4\textwidth]{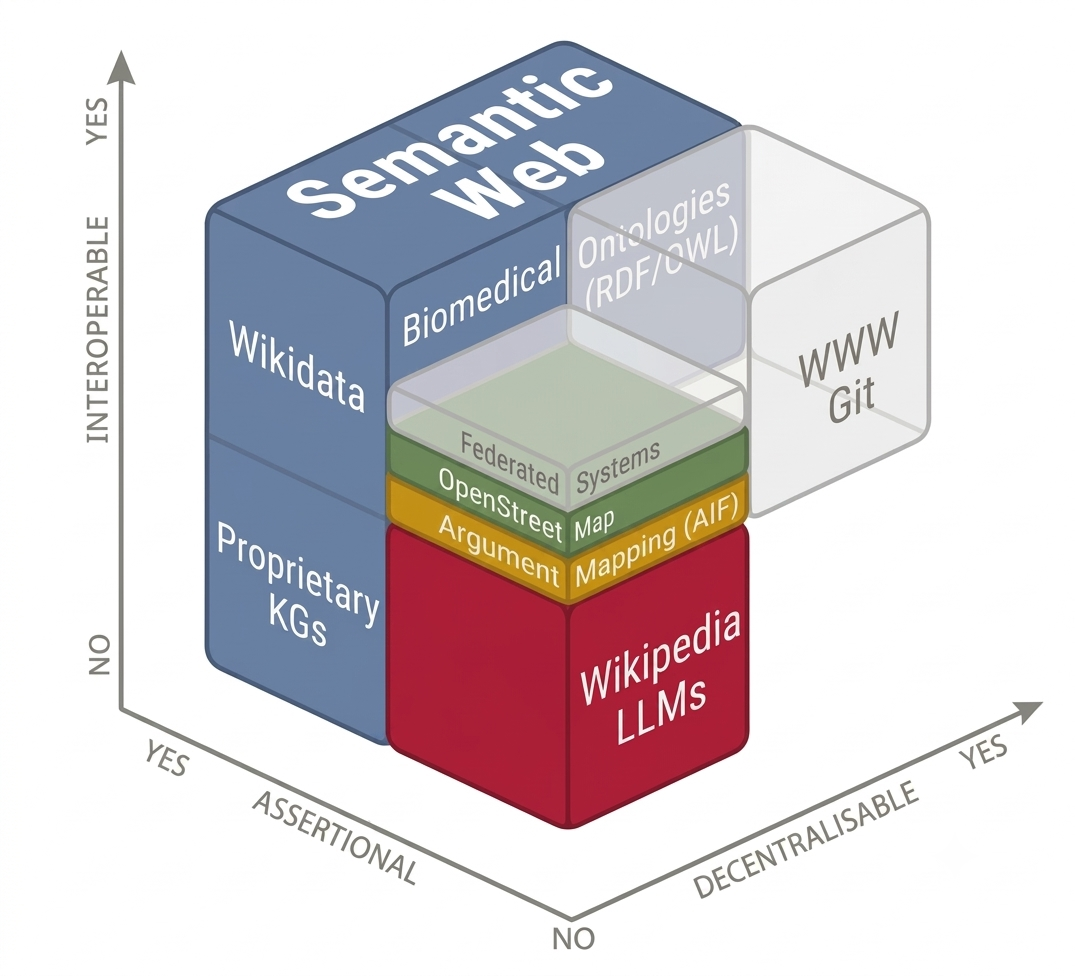} 
  \caption{The design space of information systems across three architectural  dimensions: \dimname{interoperability}, \dimname{write decentralisabil}ity, and epistemic breadth (the non-assertional axis). Among the systems considered here, the only ones that satisfy all three  are not \dimname{knowledge systems}.}
  \label{fig:gap}
\end{figure}

Recent work integrating LLMs and knowledge graphs -- using LLMs to automate KG construction and KGs to ground LLM outputs --  makes population of knowledge graphs easier and cheaper \cite{hogan2021knowledge,zhu2024llms}. 
However, this does not change a fundamental limitation of Semantic Web technologies: they deal predominantly with formally assertable, verifiable facts about  identifiable, objectively existing or widely recognised  entities. Contested claims, tentative hypotheses, evolving knowledge --  which constitute  the primary content in scientific research -- remain largely outside their scope. By contrast, the WWW provides a  structural framework that allows any independently produced content to be exchanged and accessed without pairwise conversion across the full range of human knowledge expression.
Git similarly achieves broad epistemic scope (any file content), decentralised contribution, and structural interoperability through its data model and merge mechanism. 
But neither is a \dimname{knowledge system}.
To our knowledge, no existing \dimname{knowledge system} combines \dimname{universal scope} with \dimname{interoperability} and \dimname{write decentralisability}  (cf Fig.~\ref{fig:gap}).

\medskip

\begin{figure}[htbp]
    \centering
    \includegraphics[width=0.45\textwidth]{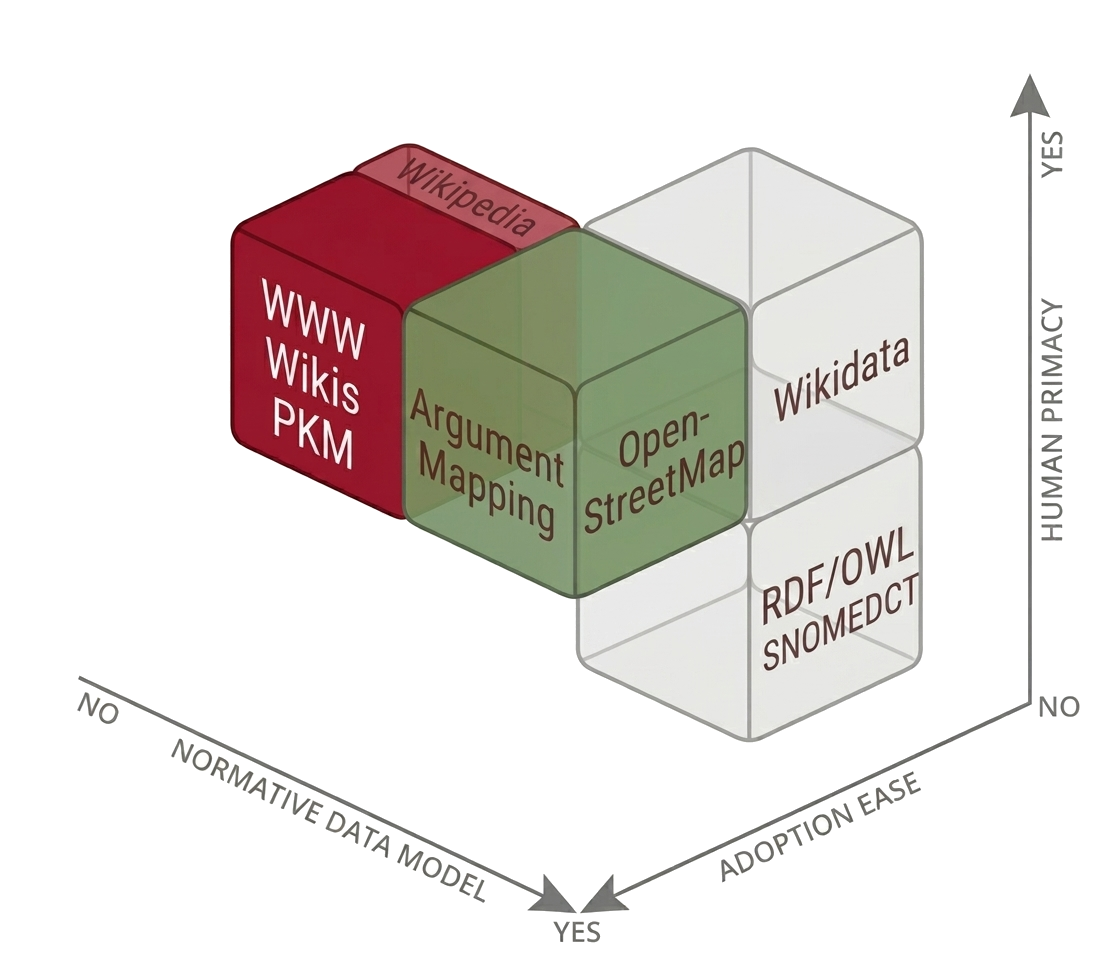} %
    \caption{\dimname{human primacy},  \dimname{adoption ease} and \dimname{accessibility} (whether the contribution barrier is low), and \dimname{normative data model} across the design space. Semi-transparent: non-\dimname{Universal Scope}.
    }
    \label{fig:dual}
\end{figure}

A complementary
way of looking at the design space is simply through the two dimensions:
\dimname{Human Primacy} and the presence of a \dimname{Normative Data Model}.
On one side are highly permissive systems  -- the WWW, PKM tools, Wikipedia  -- which favour \dimname{Human Primacy} but lack a \dimname{normative data model}. These systems have broad adoption, but their content is mostly unstructured prose that is not interoperable and only queryable through text search. On the other side are highly formal systems like RDF/OWL which enforce a \dimname{normative data model} but give up on \dimname{Human Primacy} to also enforce strict global constraints (e.g. \dimname{Global Formal Consistency}, \dimname{High Ontological Commitment}). These systems trade adoptability for theoretical guarantees  -- e.g. fine-grained queryability, consistency checking, automated logical inference  -- that deliver limited benefit until enough knowledge has been encoded.
Intermediate systems such as OpenStreetMap and Argument Mapping tools combine both dimensions -- thereby demonstrating that \dimname{Human Primacy} and a \dimname{normative data model} are not mutually exclusive -- although they do that within restricted scopes (cf Fig.~\ref{fig:dual}).

While all systems considered in Table \ref{taxonomytable} sacrifice one of the 3 dimensions, \dimname{Human Primacy}, \dimname{normative data model}, \dimname{Universal Scope},  no obvious
reason emerges from Table \ref{taxonomytable} as to why all three could not be satisfied simultaneously {\it i.e.}, why universal-scope systems could not exist that are more constrained than the WWW but less than RDF/OWL -- more structured than free text but still tolerant of ambiguity, falsehood, and disagreement -- combining low enough contribution barriers to enable broad population of the system, with sufficient structure to organise collective documentation and support \dimname{interoperability}.

The historically contingent evolution of permissive, document-centric platforms has led to the introduction of structuring layers drawn from the opposite, formal end of the design space, in an attempt to bridge a large gap. Wikidata extends Wikipedia. Semantic web technologies extend the WWW. Both these layers remain partial and marginal, covering only a small (assertional) fraction of otherwise largely unstructured prose content.

OSM challenges the assumption that the level of formalisation
required by the Semantic Web stack \cite{BaaderDL,rdfsemantics,semwebhard} is a precondition for \dimname{interoperability} in general, rather than one powerful implementation suited specifically for use cases requiring automated reasoning.
OSM has achieved widespread adoption, with a large and sustained contributor base and diverse interfaces for editing and consuming data, resulting in a large-scale, continuously populated database \cite{osmcomplete,haklay2008osm,neis2012analyzing}. Its \dimname{interoperability} rests on a lightweight \dimname{normative data model} (nodes, ways, relations; stable IDs). This structural spine supplies just enough canonical structure for large-scale coordination among contributors and consistent consumption by renderers, without paying the \dimname{accessibility and adoption} costs of full semantic ontological commitment and global consistency. For more semantic expressivity and flexibility, OSM's data model pairs this structural spine with a folksonomic tag system. With referential and structural stability ensured by the spine, OSM can afford to let tags be messy, inconsistent, or ambiguous  -- contributors agree on what is being tagged, even if they disagree on how to tag it  -- and where tags may break \dimname{interoperability} (e.g., through synonymy or polysemy), social conventions and community processes are expected to compensate. External research projects (e.g., LinkedGeoData) have demonstrated that OSM data can be lifted into RDF/OWL, suggesting that such an intermediate architecture is compatible with gradual formalisation \cite{osm_rdf}.

While OSM is characterised by the geographic nature of its content, Table~\ref{taxonomytable} shows that it is also characterised by domain-agnostic architectural design choices that are implemented in other, non-geographic systems. OSM exemplifies a balance between constraint and flexibility that is compatible with both large-scale \dimname{adoption} and \dimname{interoperability}.
Rather than giving up on \dimname{scope universality}, a way of replicating this balance for general information, may be to give up on the \dimname{global consistency} dimension -- which is anyway   in tension with the \dimname{expression intent}. The existence of scientific literature -- where contradictory and competing claims coexist -- suggests that useful collective documentation is possible without enforcing \dimname{global consistency} prior to publication \cite{fleck1981genesis,kuhn1962structure,latour1987science,latour1986laboratory,story2015contradiction}. Documentation itself can serve as the medium in which inconsistencies are made explicit and worked through over time, rather than resolved upfront.

\subsection{A Practice-Driven Response to the Design Gap}
\label{whiteboard}

The next section introduces a data model, MMM, designed to satisfy a similar architectural balance as OSM, with the aim of enabling an intermediate knowledge system in the design space discussed above.

\afterpage{ \clearpage
\newcommand{\yes}{\textcolor{ForestGreen}{\ding{52}}}
\newcommand{\intheory}{\textcolor{BrickRed}{\ding{52}}}
\newcommand{\bydesign}{\textcolor{YellowOrange}{\ding{52}}}
\newcommand{\wanted}{\textcolor{YellowOrange}{\ding{52}}}
\newcommand{\nope}{\textcolor{BrickRed}{\ding{56}}}
\newcommand{\nearnope}{\textcolor{BrickRed}{(\ding{56})}}
\newcommand{\limited}{\textcolor{ForestGreen}{(\ding{52})}}
\newgeometry{left=3cm,right=3cm} %

\begin{table}[htbp]
        \centering
        \setlength{\parskip}{-4pt}
        \setlength{\extrarowheight}{0pt}
        \renewcommand{\arraystretch}{1.5}%
        \scalebox{0.9}{\begin{tabular}{|m{0.77\textwidth}|m{0.33\textwidth}|}
                        \hline\rowcolor{gray!30}
                        \textbf{Practical Need / Design Constraint}
                                                                                         & \textbf{Dimension} (cf. \S~\ref{dimensions})
                        \\\hline \hline

                        Scientific research generates and operates on fragments (open questions, definitions, theorems, specific implications, etc),
                        not whole documents.
                                                                                         & \dimname{Post-Document Organisation}
                        \\\hline

                        Scientific research  consists largely in establishing and characterising relationships between previous results, proofs, definitions, formalisms
                                {\it etc}.
                                                                                         & \dimname{First-Class Relationships}
                        \\\hline

                        Researchers progress by expressing their situated, partial
                        understanding as it emerges  \cite{noualpersp}, before it has been formalised into established results.
                                                                                         & \dimname{Expression Intent}
                        \\\hline

                        Early whiteboard work, precisely because understanding is partial, records,  as they emerge, clarifying  structural relations between ideas and pieces of knowledge.
                                                                                         & \dimname{Epistemic Structure}
                        \\\hline

                        In practice, there is no sharp divide distinguishing  final results from intermediary steps: both are
                        work material that researchers seamlessly navigate, and reuse.
                        
                                                                                         & \dimname{homogeneous data space}
                        \\\hline

                        Contributions must outlast individual work sessions and accumulate into a common durably accessible resource  so they can be built upon.  
                       
                                                                                         & \dimname{Persistent Production} \par \dimname{Enclosure Resistance}
                        \\\hline

                        Scientific research
                        (e.g. mathematical proof building) involves periods of focused inquiry by single researchers working on a narrow line of reasoning.
                        
                                                                                         & \dimname{Local \& Individual  Value}
                        \\\hline

                       The primary medium of scientific thought and communication is natural language, produced by and addressed to human researchers. & \dimname{Human primacy}
                        \\\hline

                        Expert practitioners with  effective formalisms,  workflows and tools shouldn't be  diverted from their actual research work to learn new, generic ones.

                                                                                         &  \dimname{Adoption ease \& accessibility}
                        \\\hline

                        Without requiring of researchers to converge on  a single formalism or tool, documented understanding needs to circulate through the  diversity of existing disciplines,  workflows,  cultures and systems already in use. 
                         & \dimname{Interoperability}
                        \\\hline

Scientific understanding deepens as ideas travel across scientific communities and disciplines, are re-understood and re-applied in different contexts, requiring their re-expression in adapted formulations.

                                                                                         & \dimname{Redundancy friendliness}
                        \\\hline

                        Scientific knowledge spans heterogeneous subjects, and its building blocks take many forms: informal notes and conversation records, natural-language interpretations, bibliographic references,
                        mathematical objects, and entire articles.

                                                                                         & \dimname{Universal Scope}
                        \\\hline

                        Scientific research aims to move from idiosyncratic reasoning and ad-hoc documentation toward shared, explicit, reproducible, combinable structures of thought.

                                                                                         & \dimname{Formal typing}
                        \\\hline

                        Scientific research is a highly collective effort  that is
                        geographically and institutionally distributed worldwide.
                        Individual researchers and teams  coordinate through shared scientific method,
                        and disciplinary  conventions rather than centralised authority.
                        
                                                                                         &
                        \dimname{Worldwide collaboration}\par 
                        \dimname{Common Rules}\par \dimname{Normative Data Model}\par \dimname{Write Decentralisability}
                        \\\hline

                        Scientific research  is an ongoing
                        effort  driven by well identified imperfections and contradictions.
                       
                                                                                         & \dimname{Disagreement Expression} \&\par \dimname{Continual Improvement}
                        \\\hline

                        Scientific knowledge
                        is inherently evolutive, incomplete, and inconsistent.
                        Convergence in scientific research is local and  temporary,  open to challenge, refinement, and overturning.

                                                                                         & No \dimname{Convergence intent}\par No \dimname{Global Formal Consistency}\par No \dimname{High Ontol. Commitment}
                        \\\hline

                        Research work is highly context-dependent, relying on %
                        definitions, notations, and baseline results 
                        that are costly to redocument, especially 
                        during live reasoning, 
                        yet are still typically reintroduced
                        in final documents, often with slight variations that make individual pieces of knowledge difficult to trace.

                                                                                         & \dimname{Contextual Enrichment}
                        \\\hline

                     Collective documentation grows richer and more rigorous as contributors build on each other's work   (add precisions, connect to other work, surface implicit assumptions, fill gaps, etc), reducing the documentation burden on each individual.

                                                                                         & \dimname{Collective Emergent Benefit}
                        \\\hline
                \end{tabular}}
        \smallskip
        \caption{Practical needs from collaborative mathematical research -- in particular interdisciplinary work on Boolean automata networks as formalisations of systems of interacting objects with biological applications~\cite{noualBAN} -- and the MMM dimensions they motivated or excluded.  }
        \label{tab:mmm-constraints}
\end{table}
\clearpage
\restoregeometry
}

Despite this architectural similarity between MMM and OSM, the MMM design has a practice-driven origin, independent of OSM. It emerged from recurring frustrations encountered while doing collaborative mathematical research work about "Boolean Automata Networks"    \cite{noualBAN}, at the whiteboard.
Traditional document-based ways of recording research progress, whether informal (drafts, notes) or formal (peer-reviewed publications), proved inadequate for two related reasons. First, they are difficult to retrieve and navigate on the fly at the whiteboard.
Second, they are ill-suited to capturing the local, fine-grained, incremental, and heterogeneous character of knowledge in the making: open questions, partial results, fragments of proofs, evolving conjectures, links between ideas  -- ranging from highly contextual and imperfect formulations mixing natural language and ad hoc notation to publishable theorems and their proofs.

The MMM data model was conceived as a practical response to this inadequacy of existing document-based solutions for the day-to-day reality of scientific research. Because research is inherently document-heavy, its aim was not to replace documents, but to generalise them while remaining compatible with existing scientific records and practices.
The design constraints that shaped MMM are summarised in Table \ref{tab:mmm-constraints}. MMM's position along the dimensions it was designed to address or exclude is assessed in Table \ref{tab:dimensions-evidence} on page~\pageref{tab:dimensions-evidence} following the data model definition in Section~\ref{mmm} and the pilot deployment described in  Section~\ref{sec:pilot}.
 
 \section{The MMM data model} \label{mmm}

This section defines the MMM data model abstractly. MMM stands for "Mutable Mutual Meaning" or "Mutable Mutual Matrix".

A reference implementation (\href{https://myrmex.app}{myrmex.app}) and research prototype applications demonstrate the practical implementability of the data model, described in Section 4. The MMM data model itself is independent of implementation maturity and is not tied to any particular application, serialisation format, or storage backend.

The core syntax and semantics of the MMM data model are presented below. Some components, such as the set of edge types, are intentionally underspecified  and will be subject to a subsequent stabilisation process based on community feedback and empirical usage. This phased approach separates foundational design decisions presented here, from elements that require broader consensus to ensure durable interoperability.\medskip

The atomic element of MMM formatted content is the {\bf typed contribution}, a.k.a. "{\it MMM landmark}". The core of the MMM data model proposition lies in the typing of contributions which are otherwise deliberately very unconstrained.\medskip

Below, for any set $X$, we let $X_\mathtt{NULL}=X\cup\{\mathtt{NULL}\}$, and we denote the power set of $X$ by $\mathcal{P}^X$.

\subsection{Formal definition} \label{contributions}
To define the set of MMM contributions, let:
\begin{itemize}
    \item $\mathcal{I}$ be a set of universally unique identifiers (UUID), identifying existing MMM landmarks, and containing a special reserved identifier $i_0\in\mathcal{I}$ (the nil UUID '{\tt 00000000-0000-0000-0000-000000000000}').
    \item $\mathcal{D}$ be the set of {dates}.
    \item $\mathcal{S}=\{\mathtt{Private}, \mathtt{Public}\}\cup\{\mathtt{Shared}(x) | x\in R\}\cup\{\mathtt{Licensed}(l) | l\in L\}$  be the set of visibility/access statuses. $\mathtt{Private}$ restricts access to the contribution's authors. $\mathtt{Public}$ grants unrestricted access  (analogous to \href{https://creativecommons.org/publicdomain/zero/1.0/deed.en}{CC0} public domain). $\mathtt{Shared}(x)$ limits access to the  entities identified by $x$ (e.g.   users, teams,   organisations). $\mathtt{Licensed}(l)$ grants global unrestricted read access but restricts use under  licence $l$.
    \item $\mathcal{A}= \mathcal{P}^\mathcal{N}\times \mathcal{D} $ be the set of {\bf authorships}. An authorship is given by {\it (i)} a team which is a set of author names from the set $\mathcal{N}$ of author names, and {\it (ii)} a date in $\mathcal{D}$ .

    \item $\mathcal{K}=\{\mathtt{Edge}, \mathtt{Vertex},\mathtt{Pen},\mathtt{Pit}\}$ be the set of {\bf contribution kinds}. The kind field determines which other fields are required or allowed. It defines the structural role of an MMM landmark, analogous to OpenStreetMap's nodes, ways and relations. Only MMM's $\mathtt{Pit}$ kind representing absurdity has no analogue in OSM.
    \item $\mathcal{T}=\mathcal{T}_\mathtt{Edge}\cup\mathcal{T}_\mathtt{Vertex}\cup\mathcal{T}_\mathtt{Pen}$ be the set of {\bf contribution types} -- the core of the MMM data model, detailed below in \S\ref{mmmtypes}.

    \item $\mathcal{M}$ be the set of {\bf marks} (see \S\ref{marks} below).
    \item $\mathbb{T}$ be the set of text {\bf labels} (character strings). Labels may include formatted text (Markdown, LaTeX), hyperlinks, or  external resource identifiers (e.g., OSM node IDs, Wikidata Q identifiers, DOIs).
\end{itemize}
MMM contributions/landmarks are defined as follows:
\begin{equation*}
    \begin{aligned}
        \mathcal{C} & = \mathcal{C}_\mathtt{Vertex} \cup \mathcal{C}_\mathtt{Edge} \cup \mathcal{C}_\mathtt{Pen} \cup \mathcal{C}_\mathtt{Pit} \\[10pt] \mathcal{C} &\subseteq \underbrace{ \mathcal{I} \times \mathcal{D}_\mathtt{NULL} \times \mathcal{S}_\mathtt{NULL} \times\mathcal{P}^\mathcal{A}_\mathtt{NULL} }_{\text{id, date, status, authorships}} \times \underbrace{ \mathcal{P}^\mathcal{M}_\mathtt{NULL}}_{\text{marks}} \times \underbrace{\mathcal{K}}_{\text{kind}} \times \underbrace{\mathcal{T}_\mathtt{NULL}}_{\text{type}} \times \underbrace{\mathbb{T}_\mathtt{NULL}}_{\text{label}} \times \underbrace{\mathcal{I}^2_\mathtt{NULL}}_{\text{endpoints}} \times \underbrace{\mathcal{P}^\mathcal{I}_\mathtt{NULL}}_{\text{contents}}
    \end{aligned}
\end{equation*}
where
\begin{equation*}
    \begin{aligned}
        \mathcal{C}_\mathtt{Vertex} & = \mathcal{I} \times \mathcal{D}  \times \mathcal{S} \times \mathcal{P}^\mathcal{A} \times \mathcal{P}^\mathcal{M}_\mathtt{NULL} \times \{\mathtt{Vertex}\} \times \mathcal{T}_\mathtt{Vertex} \times \mathbb{T} \times \{\mathtt{NULL}\}^3 \\[6pt] \mathcal{C}_\mathtt{Edge} &= \mathcal{I} \times \mathcal{D}  \times \mathcal{S}\times \mathcal{P}^\mathcal{A} \times \mathcal{P}^\mathcal{M}_\mathtt{NULL} \times \{\mathtt{Edge}\} \times \mathcal{T}_\mathtt{Edge} \times \mathbb{T}_\mathtt{NULL} \times \mathcal{I}^2 \times \{\mathtt{NULL}\} \\[6pt] \mathcal{C}_\mathtt{Pen} &= \mathcal{I} \times \mathcal{D}  \times \mathcal{S}\times \mathcal{P}^\mathcal{A} \times \mathcal{P}^\mathcal{M}_\mathtt{NULL} \times \{\mathtt{Pen}\} \times \mathcal{T}_\mathtt{Pen} \times \{\mathtt{NULL}\}^3 \times \mathcal{P}^\mathcal{I} \\[6pt] \mathcal{C}_\mathtt{Pit} &=  \underbrace{\{i_0\} \times \{\mathtt{NULL}\} \times \{\mathtt{Public}\} \times \{\mathtt{NULL}\}}_{\text{Pit's id, date, status, authorships}} \times \mathcal{P}^\mathcal{M}_\mathtt{NULL} \times \{\mathtt{Pit}\} \times \{\mathtt{NULL}\}^5
    \end{aligned}
\end{equation*}
For convenience, we refer to collections of MMM contributions as "MMM graphs." Strictly speaking, since edges may have other edges as endpoints (the node set includes the edge set), this departs from the standard graph-theoretic definition. 

Informally, there is only one $\mathtt{Pit}$ landmark but possibly multiple versions of it with differing mark sets. The $\mathtt{Pit}$ landmark has no authorships set and cannot be created by users.  It  represents logical absurdity.
It is intended  to serve as a basis for a form of censureless quality control. Linking a  contribution to the $\mathtt{Pit}$ landmark is the closest MMM provides to flagging the contribution as false. It allows users to signal that they consider a contribution, or its  relation to other contributions, to be  structurally unsound.
 For example, an {\tt Answers Edge} between a {\tt Question Vertex} about the biological mechanisms of RNA vaccines and a {\tt Narrative Vertex} claiming that 5G technology is murderous could be flagged in this way, signalling the irrelevance of the {\tt Narrative} with respect to the {\tt Question}, without taking a position on the semantics of the {\tt Narrative}.

 Every other kind of MMM contribution ({\tt Vertex}, {\tt Edge}, {\tt Pen})  -- we will call those {\bf user contributions} or {\bf user landmarks} as opposed to the pit -- has an identifier in $\mathcal{I}$, a timestamp in $\mathcal{D}$, a non-empty set of authorships in $ \mathcal{P}^\mathcal{A} $, an optional set of marks in $ \mathcal{P}^\mathcal{M} $, and a kind in $\mathcal{K}\setminus\{\mathtt{Pit}\}$.

{\bf \texttt{Vertex} contributions} additionally have a mandatory label in $\mathbb{T}$. {\bf \texttt{Edge} contributions} have an optional label in $\mathbb{T}$ and mandatory endpoints, the edge {\bf source} and edge {\bf target}, both in $\mathcal{I}$. {\bf \texttt{Pen} contributions}, like OSM relations, are container contributions grouping other contributions by reference. They have no label and no endpoints, but have an optional set of {\bf contents} in $\mathcal{P}^\mathcal{I}$. \medskip

MMM contributions thus comprise several fields: up to 8 non-null fields for vertices and pens, and up to 10 for edges. Their epistemic role and semantic content are however primarily determined by a minimal subset of these fields: {\it kind, type} and {\it label} for vertex contributions, {\it kind, type} and {\it contents} for pen contributions, and {\it kind, type} and endpoints ({\it source} and {\it target}) and optional {\it label} for edge contributions.
The {\it kind} field enforces which other fields are required or allowed. The {\it type} field is detailed in \S\ref{mmmtypes}.
The remaining fields support identification, provenance (e.g., authorship and timestamp), and extensibility (cf \S \ref{marks}). Those remaining fields are generally not required to interpret the core informational content of a contribution. In this sense, MMM separates common meaning from coordination metadata. \medskip

Labels of MMM contributions may include formatted text (e.g., Markdown or rudimentary LaTeX), whose optional rendering is left to consuming applications. A label may also consist of a hyperlink to an external document. Any information expressible in free text (including structured and styled text such as Markdown or LaTeX) can thereby be represented within a MMM contribution, as its label. MMM content can thus span a continuum from unstructured to structured. This establishes a baseline level of interoperability with document-centric systems. At the limit, entire documents (e.g., articles or books) can be represented as single MMM contributions (directly or via hyperlinks). Such contributions  do not exploit the structural capabilities of MMM. But as they  get used  -- commented on, questioned, referred to  -- they may be incrementally decomposed into smaller, interrelated MMM contributions that make the internal structure of the initial coarse-grained contribution more explicit.
\medskip

In consideration of decentralisability, external resources (e.g. DOI, OSM node identifiers, Wikipedia articles, Wikidata Q identifiers, RDF/URIs) are not represented as first-class MMM contributions. MMM identifiers apply exclusively to MMM-native contributions. This means that edge endpoints and pen contents can't be external identifiers. At the MMM level, no mechanisms are defined for resolving, synchronising, or tracking the state of external resources. Interoperability with external systems is limited to  referencing external resources as  opaque values within MMM contribution labels (or possibly marks).
It is the responsibility of applications  built on top of MMM to decide whether and how to interpret, resolve, enrich, or jointly present external resources alongside MMM contributions, and possibly  maintain external synchronisation layers.
\medskip

MMM authorships generalise the traditional notion of authorship since a single MMM contribution may be associated with multiple authorships. The rationale is to treat authorship as accountability rather than ownership \cite{foucault2003author,authorship}, with the intent that an author's responsibility and recognition should be a function of the overall body of their work -- patterns of contributions across the system -- rather than any single contribution.
The definition of metrics to qualify patterns of contributions in a typed MMM graph is planned research work. Elementary metrics (e.g. for “usefulness” and “depth”) have been implemented in our reference application and represented using heat maps and 3D (cf Fig.~\ref{fig:metrics}).
\medskip

\afterpage{
    \begin{figure}[htbp]
        \centering
        \includegraphics[width=0.7\textwidth]{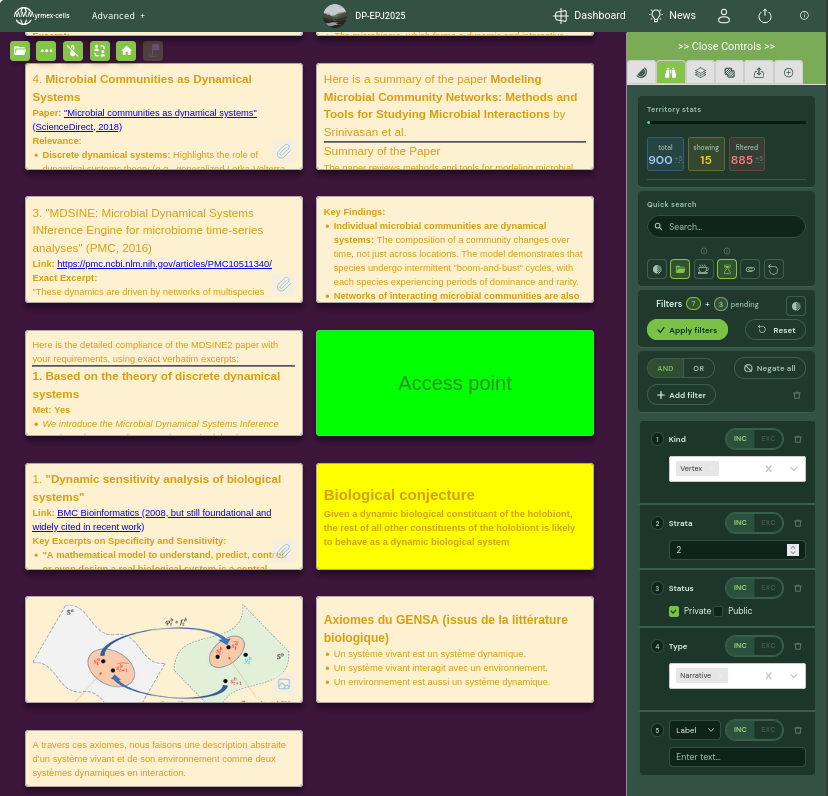}
        \caption{Illustration of MMM's fine-grained filtering capabilities using a real researcher workspace on a MMM-based note-taking prototype application. Courtesy of D. Pastor.}
        \label{fig:filters}
    \end{figure}
}

Note that by providing a semi-structured representation of knowledge, the  MMM data model enables fine-grained filtering and selection of contributions (implemented in the reference application, cf Fig~\ref{fig:filters}).
\medskip

A JSON schema  of the MMM data model will be published as part of the forthcoming stable specification, settling the normative names and formats of all fields defined in the present document. 

\subsection{Formal Typing} \label{mmmtypes}

The type field is mandatory for all user-created MMM contributions.  However, applications may supply a default type when the user does not provide one. For users, typing should be like hashtags in social media and like punctuation  in real-time messaging: it structures meaning but remains {optional and forgiving} (\dimname{adoption ease \& accessibility}). When a user does not provide a type, the application supplies a default from the set of predefined ones.
 \medskip
 
The set of vertex types is fixed: 
$$ 
\mathcal{T}_\mathtt{Vertex}=\{\mathtt{Narrative}, \mathtt{Question},\mathtt{Existence},\mathtt{Instruction},\mathtt{Data}\} 
$$
$\mathtt{Question}$ is the recommended type for vertex contributions whose label constitutes a question. $\mathtt{Instruction}$ is the recommended type for vertex contributions whose label constitutes an instruction --e.g. "Be nice.". $\mathtt{Narrative}$ is the recommended type for vertex contributions whose label constitutes a (series of) statements, an opinion, a story. $\mathtt{Existence}$ is the recommended type for contributions that are not full sentences and that denote or describe a concept, a property, something that exists, or at least matters to the person contributing it. Finally, $\mathtt{Data}$ is the recommended type for data -- e.g. "42", "42km", "01-42-92-81-00". Data landmarks may be links to entries in external databases. Contrary to the first four vertex types, data landmarks with identical labels are not, by default, candidates for merge. Efforts to reduce  semantic redundancy among MMM contributions will focus on questions, instructions, narratives and existences.
\medskip

The set of pen types proposed here is provisional: 
$$ 
\mathcal{T}_\mathtt{Pen}=\{\mathtt{Default}, \mathtt{Definition},\mathtt{Folder},\mathtt{Territory}\} 
$$

The set of edge types  
can be organised into three broad categories:
$$\mathcal{T}_\mathtt{Edge}=\mathcal{T}_{-}\cup\mathcal{T}_\leftrightarrow\cup\mathcal{T}_\rightarrow$$
\begin{itemize}[label={\textbullet}]
\item The set of {\bf adirectional edge types} $\mathcal{T}_{-}=\{\mathtt{Relate}\} $ contains a single type which may be used as a fallback but is generally to be avoided.
\item The set of {\bf bidirectional edge types} $\mathcal{T}_{\leftrightarrow}=\{\mathtt{Equate},\mathtt{Differ}\} $ contains two  types,  one to express various forms of  similarity (e.g. equivalence, equality, synonymy)    and the other to express various forms of difference or opposition. 
\end{itemize}
Adirectional and bidirectional types constitute the "horizontal" types. 
\begin{itemize}[label={\textbullet}]
\item A  provisional set of {\bf unidirectional} ("vertical") {\bf edge types} $ \mathcal{T}_\rightarrow\approx \{\mathtt{RelatesTo},$ $\mathtt{Answers},$ $\mathtt{Questions},$ $\mathtt{Pertains}, $ $\mathtt{Instantiates},$ $\mathtt{Substantiates},$ $\mathtt{Precedes},$ $\mathtt{Challenges},\ldots\}$ containing  a fallback unidirectional edge type, $\mathtt{RelatesTo}$,  generally to be avoided. All other unidirectional edge types orient edges from  more specific or concrete source landmark to   more general or abstract target landmark. 
Unidirectional edge types, like other MMM types, are deliberately broad and semantically permissive. They define broad categories of relationships rather than precise semantic predicates. In particular, \texttt{Pertains}, the most common epistemically specific  edge type,  covers a deliberately wide range of contextually determined meanings: a contribution   may \texttt{Pertain} to another   in the sense of belonging to it, characterising or specifying it, supporting it, being mentioned or involved in it, being a detail of it, being involved in its proof, its definition or simply its wording, being presupposed by it\ldots{} Until possibly undergoing the promotion process described in \S\ref{evolution} to become a native MMM edge type, even logical implication is to be expressed using a {\tt Pertains} edge with a specifying semantic mark (cf \S\ref{marks}).
The underlying rationale for this intentional breadth is that conflating several loosely  related meanings under a single well-understood edge type,  that is easy to distinguish from  other types, reduces the cognitive burden of choosing the "right" type, increasing the likelihood that contributors will use it meaningfully and as expected, rather than hesitating or misusing a more finely grained alternative with subtle contextual semantics. Since MMM does not target automated reasoning,  semantic specificity is less important than structural regularity and ease of use.
\end{itemize}
This last category of edge types remains open-ended and subject to refinement but is intended to become fixed in the near future. 
The complete list of edge type semantics will be published subsequently.

MMM imposes a small fixed set of edge types, deliberately permissive and tolerant of ambiguity.
The intended meaning of an edge is left to context, or can  be specified through an optional free text label and/or semantic marks  (see \S~\ref{marks}).
This design choice reflects the trade-off between establishing a shared common ground (through the small set of fixed native MMM types) and preserving flexibility. 
  \medskip

\subsection{Marks} \label{marks}

An MMM contribution's meaning is normally conveyed by its core components: {\it type, label, source, target,} and {\it contents}. MMM {\it marks} are primarily intended for metadata, display, or administrative purposes. They can, however, also, locally, be used to refine or extend native MMM semantics in a lightweight contextual way. They can serve as a form of meta-typing that contextualises or specialises the meaning of native MMM types. In this sense, {\it marks}  enable semantic refinement to emerge from practice rather than being fully fixed in advance. Groups of collaborators may gradually develop shared "meta-types" that capture recurring patterns of meaning and interpretation for them. These can be implemented informally, locally, through semantic {\it marks}, without changing the underlying data model. When such meta-types become sufficiently stable and broadly useful, they may later be promoted to native MMM edge types (see \S\ref{norms}).\medskip

A {\it mark} is defined by {\it (i)} a set of situations in which the mark applies, {\it (ii)} a timestamp in $\mathcal{D}$, {\it (iii)} a value and {\it (iv)} a mark type in:
\begin{align*}
    \mathcal{T}_{\mathcal{M}} = \{ &
    \mathbf{Stratum},
    \mathbf{Semantic},
    \mathbf{Obsolete},
    \mathbf{Reference},
    \mathbf{Attachment},
    \mathbf{Cosmetic},               \\ &
    \mathbf{Position} \ \}.
\end{align*}
A situation
is represented as a set of key-value pairs (e.g. \texttt{appid}, \texttt{userid}, \texttt{deviceid}). MMM does not prescribe a fixed set of situation keys. Implementations may define their own.  Future versions may introduce conventions or restrictions to improve consistency across implementations.\medskip

Let  $\mathfrak{S}$ be the set of {situations}, let  $\mathbb{B}=\{0,1\}$, let $\mathbb{T}$ be the set of text labels as before, and let $\mathcal{V}=\mathbb{T}\cup\mathbb{B}\cup\mathbb{N}\cup\mathcal{V}_\mathbf{Cos} \cup\mathcal{V}_\mathbf{Pos}\cup\mathcal{V}_\mathbf{Att}$ be the set of mark values.  The set of marks is:
\begin{align*}
    \mathcal{M} & = \mathcal{M}_\mathbf{Str}\cup
    \mathcal{M}_\mathbf{Sem}\cup
    \mathcal{M}_\mathbf{Obs}\cup
    \mathcal{M}_\mathbf{Ref}\cup
    \mathcal{M}_\mathbf{Att} \cup
    \mathcal{M}_\mathbf{Cos}\cup
    \mathcal{M}_\mathbf{Pos}                                                                                                 \\
                & \subset   \mathcal{P}^\mathfrak{S}\times\mathcal{D}\times  \mathcal{V}   \times  \mathcal{T}_{\mathcal{M}}
\end{align*}
where:
\begin{itemize}[label={\textbullet}]
    \item {\bf Stratum marks} of $\mathcal{M}_\mathbf{Str}   =   \mathcal{P}^\mathfrak{S}\times\mathcal{D}\times   \mathbb{N}   \times   \{\mathbf{Stratum}\} $ provide a way of quantitatively classifying MMM contributions.

    \item {\bf Semantic marks} of $ \mathcal{M}_\mathbf{Sem} = \mathcal{P}^\mathfrak{S}\times\mathcal{D}\times   \mathbb{T}  \times  \{\mathbf{Semantic}\} $ can serve as user defined tags. They can also have more application-level roles. For example, in the reference implementation (\href{https://myrmex.app}{myrmex.app}), the semantic mark with value "{\tt home}" has the fixed application-level meaning of  workspace starting point.

    \item  {\bf Obsolete  marks} of $ \mathcal{M}_\mathbf{Obs}= \mathcal{P}^\mathfrak{S}\times\mathcal{D}\times  \{0,1\}  \times   \{\mathbf{Obsolete}\}$ enable pseudo-deletion of contributions in a way that compares to tombstones in CRDTs \cite{kleppmann2017conflict,shapiro2011crdt}.

    \item  {\bf Reference marks} of $\mathcal{M}_\mathbf{Ref}  =  \mathcal{P}^\mathfrak{S}\times\mathcal{D}\times  \mathcal{V}_\mathbf{Ref}  \times  \{\mathbf{Reference}\}$  are for external resource identifiers  (e.g. URIs, Wikidata Q identifiers, DOIs).

    \item  {\bf Attachment marks} of $ \mathcal{M}_\mathbf{Att}  =  \mathcal{P}^\mathfrak{S}\times\mathcal{D}\times  \mathcal{V}_\mathbf{Att}  \times  \{\mathbf{Attachment}\} $ are for attaching a file to a contribution. File metadata, including access information, is stored in the mark's value.

    \item  {\bf Cosmetic marks} of $\mathcal{M}_\mathbf{Cos}  =  \mathcal{P}^\mathfrak{S}\times\mathcal{D}\times  \mathcal{V}_\mathbf{Cos}  \times  \{\mathbf{Cosmetic}\}$ are for customising the rendered style of a contribution in a given software application.

    \item  {\bf Position marks} of $\mathcal{M}_\mathbf{Pos}  =  \mathcal{P}^\mathfrak{S}\times\mathcal{D}\times  \mathcal{V}_\mathbf{Pos}  \times  \{\mathbf{Position}\}$  are for graphical software applications that need to display contributions geographically on the viewport.

\end{itemize}
Software applications have a degree of liberty in the implementation of $\mathcal{V}_\mathbf{Ref}$, $\mathcal{V}_\mathbf{Cos}$, $\mathcal{V}_\mathbf{Pos}$, and $\mathcal{V}_\mathbf{Att}$. However, for the sake of interoperability, standardised data structures are recommended: standard bibliographic formats (e.g. BibTeX, RIS) and identifier schemes  (e.g. DOI, ISBN, arXiv ID) for $\mathcal{V}_\mathbf{Ref}$, CSS property styles for $\mathcal{V}_\mathbf{Cos}$, x-y coordinates for $\mathcal{V}_\mathbf{Pos}$ and certain file properties for $\mathcal{V}_\mathbf{Att}$ that will be standardised in the future.\medskip

To illustrate the role of the {\it situations} field in {\it marks}, consider {\bf Position  marks}   that specify x-y coordinates. Since the MMM data model is meant to support interoperability, each MMM contribution could be read by multiple software applications. One software application $S_1$ may not implement any form of graphical layout of contributions. For $S_1$, all position marks of all contributions are useless. Another application $S_2$, may want to display a specific contribution $c$ at the center of the viewport for all users, while application $S_3$, may want to display the same contribution $c$ at the top left of the viewport for user $u_1$ and at the bottom right for all other users. In this case, contribution $c$ will need three  position  marks, one with value $(x_{center},y_{center})$ that applies in situations where the application is $S_1$, one position mark with value $(0,0)$ that applies when the application is $S_2$ and the user is $u_1$, and a third position mark with value $(x_{max},y_{max})$ that applies for all other users of $S_2$. These three position marks of contribution $c$ will each be associated with a different situations set, depending on the software application and users they concern.

\subsection{Normative requirements  toward decentralised coordination} \label{norms}

Certain properties of MMM contributions are normatively fixed by the data model specification and must be respected by all conforming implementations (\dimname{strong common contribution rules}). These requirements are preconditions for \dimname{decentralisable contribution}. 

\begin{itemize}[label={\textbullet},leftmargin=*]
  \setlength\itemsep{0.5em}
  \item {\bf Endpoint preexistence:} an edge contribution may only be created between landmarks that   already exist and are identified at the time of the edge's creation. The creation timestamp of each endpoint must therefore be anterior or equal to that of the edge, never posterior. And  an edge can't have itself as one of its own endpoints. 
  \item {\bf Identifier and creation date immutability:} The identifier and creation date of a MMM contribution are assigned at creation and are permanently immutable. They must never be modified, reassigned, or recycled, even after a contribution is deleted or marked obsolete. Implementations that import MMM contributions from other deployments must preserve all identifiers and creation dates exactly as received.
  \item {\bf Status monotonicity:} The status of an MMM contribution may only be relaxed, never tightened: $\mathbf{Private} \rightarrow    \mathbf{Shared}(x)    \rightarrow \mathbf{Licensed}(l)  \rightarrow    \mathbf{Public}$, where  $\mathbf{Shared}(x)$ may relax to $\mathbf{Shared}(y)$ if $x \subseteq y  $, and  $\mathbf{Licensed}(l)$ may relax to $\mathbf{Licensed}(l')$ if $l'$ is demonstrably less restrictive than $l$. The {\bf Public} status is terminal.
  \item {\bf Status and sharing:} The data model does not enforce access control -- this is an application-layer responsibility.
        For sharing across independent deployments, globally meaningful identifiers (e.g. ORCID iDs, DIDs) should be used in the  $\mathbf{Shared}(x)$ status.
        Applications must respect the licence attached to $\mathbf{Licensed}(l)$ status of a contribution and make licence terms visible to users. Compliance is an application responsibility.
        Applications must not expose  contributions with status  {\bf Private} or $\mathbf{Shared}(x)$  to unauthenticated or unauthorised users (i.e. users not identified as being among the authors for {\bf Private}  contributions, or part of $x$ for $\mathbf{Shared}(x)$ contributions).
        Applications must not support any form of public query interface that returns such non-public contributions.

  \item {\bf Limited styling for Label portability:} The label field of any MMM contribution must contain only plain text, Markdown, and KaTeX formulae, encoded in UTF-8.  Applications must not embed proprietary styling, application-specific markup,  or custom formatting conventions within the label field of contributions that may be communicated externally. Presentational customisation belongs exclusively in Cosmetic marks (cf \S\ref{marks}).
        Any conforming application must be able to render any label intelligibly using only Markdown and KaTeX.
        Should additional formatting capabilities beyond Markdown and KaTeX prove necessary in practice, their inclusion will be considered through the standard MMM evolution process defined in \S\ref{evolution}.

  \item {\bf Type semantics:} Conforming applications must respect the intended semantics of native MMM types 
  --   vertex types, edge types, and pen types -- as provided in §\ref{mmmtypes} and in the forthcoming stable specification (cf.\ §\ref{evolution}). 
        For example, if a contribution represents a question, it must be typed as a $\mathtt{Question}$ vertex, not as a $\mathtt{Narrative}$ vertex or any other type. Applications should be designed so that any automatic typing happens in good faith, and where possible they should help users publish well-typed contributions, e.g., by suggesting appropriate types or providing clear defaults.\medskip

        However, users themselves may type contributions poorly out of convenience or misunderstanding. This is permitted. The requirement applies to applications, not to end users.

  \item {\bf Robustness to poor typing:} Applications should expect that poorly typed contributions may appear in content sourced from other applications. Applications remain free to reject or ignore such content based on their own quality criteria and purpose.

  \item {\bf Handling unstructured content:}  Free-text or multi-media contributions should be typed as  \texttt{Narrative}  vertices when no other type applies.

  \item {\bf Default edge type:}  Edges whose relationship type cannot be determined or specified at contribution time
        should be typed \texttt{Relate} (adirectional) if they have no direction, or \texttt{RelatesTo} (directional) if they have.
        These two types are the fallback for unqualified relationships.
        \item {\bf Mandatory Pit:} All shared MMM graphs must contain the \texttt{Pit} landmark, representing logical absurdity.
        
        \item {\bf Pit-based Quality control:} Conforming applications are expected to implement quality control mechanisms based on the \texttt{Pit} landmark as described in \S\ref{contributions}. 
        Applications may apply default filters based on \texttt{Pit} connectivity -- for instance by visually distinguishing or deprioritising \texttt{Pit}-connected contributions  -- and may treat \texttt{Pit} connectivity as a validity constraint or use it to block contributions from entering the system. But this is an application design choice. 

  \item {\bf Situations-based filtering of marks:}
        When sharing a contribution, applications must strip any mark whose {\it situations set} does not match the recipient's context (i.e. user, application, device, workspace, etc).  Recipients must never receive marks not intended for them.

  \item {\bf Attachment marks:} Attachment marks whose value references a local filesystem path must have their situations set restricted to the specific device or user to which that path is meaningful.
        A contribution should normally have at most one attachment mark. Multiple attachment  marks are allowed only if: their {\it value} fields provide different access paths (e.g. public URL vs local copy), and the files accessible at those locations are identical (same content hash), and their {\it situations sets} are distinct  (e.g. empty set vs device-restricted set). Otherwise,   applications must use separate  connected contributions, each with its own attachment mark.

  \item {\bf Obsolete mark semantics:}  The Obsolete semantic mark has a fixed and reserved meaning: it signals
        that a contribution is to be considered deleted in the situational context defined by the mark's situations field.
        This meaning must not be repurposed or overloaded for any other meaning  by conforming implementations. The Obsolete mark is comparable to a CRDT tombstone \cite{shapiro2011crdt}: it delays actual destruction of deleted contributions,  allowing deletions to propagate  across independent deployments without central coordination, and avoiding silent resurrections.  A full CRDT-based merge behaviour for Obsolete marks, with a formal merge operator,  is deferred to future work  (cf \S \ref{evolution}).

  \item {\bf Field immutability upon publication:}  Once an MMM landmark is Public, its core fields ({\it kind, type, label, source, target, contents}) become immutable. After publication, only the {\it authorships} and {\it marks} fields may be modified.
        This constraint is motivated by the same concern as stable identifiers: a public contribution that has been referenced, reused, or built upon by other contributions in other deployments must not change its informational content, as such changes would propagate meaning drift to all contexts that reference it.
        Corrections are expressed by creating a new contribution linked to the old (obsoleted) one via an edge  indicating that the new contribution replaces the old.  It is left to the application layer to decide which contributions to show and how to present them.
        Authorships and marks are exempted because they are metadata and quality control fields whose evolution should not substantially and globally affect the informational content of the contribution.

  \item {\bf Authorship monotonicity upon publication:} Once Public, within a given replica, a landmark's {\it authorships} set may only grow  (no global obligation to incorporate authorships from  untrusted sources). New authorship records may be added to acknowledge additional contributors, but existing authorship records cannot be removed or modified.
        Published contributions continue to be annotatable and (re)contextualisable via new, connected contributions.

  \item  {\bf Authorships and author teams:}     The {\it authorships} field is an unordered set of authorships. The team component of any authorship is an unordered set of author names. No significance is attached to any implied order. When displayed, applications should default to  ordering authors alphabetically by name.  Applications must not interpret any order as indicating relative importance, contribution level, or responsibility.

  \item  {\bf AI-generated content:} Contributions generated wholly or substantially by an AI system rather than by a human must include the name and version identifier of the AI system (e.g., "GPT-4 (OpenAI)", "LLaMA-3-70B (Meta)") as an author in the {\it authorships} field. Applications that generate MMM content via automation are responsible for ensuring this requirement is met.
        If multiple AI systems contribute to a single contribution (e.g., a cascade), each must be included as a co-author.
        When a human edits an AI-generated contribution, they should be added as a co-author, alongside the existing AI author.
        A human editor may manually remove the AI author entry if, in good faith, they deem that their editing has substantively transformed the contribution.
        Systematic or mass production of AI-generated contributions attributed solely to human authors is forbidden. Conforming applications must implement safeguards 
        to detect and prevent such abuse.

  \item  {\bf No universal semantics:}  MMM has no \dimname{convergence Intent}. Contributions are recorded expressions of their authors, not assertions of universal meaning. Conforming applications must not assume any external or globally synchronised semantics. MMM has no native notion of truth and makes no claim about a contribution's accuracy or consensuality.   Qualification of contributions (e.g., plausibility, corroboration, bias) is delegated to appropriately defined graph-theoretic metrics.

  \item {\bf External resource references:} When a contribution  refers to an external resource, the source should be recorded (e.g. via a {\bf Reference} mark), preserving the possibility of contributing back. Applications may process external references freely, but must not treat any MMM contribution as guaranteeing universal semantics by virtue of external provenance -- even if the external source has \dimname{Convergence intent} -- and should not rely on a {\bf Reference} mark to remain current beyond the contribution's date.
        {External resources (e.g., DOI, URIs) are not first-class MMM contributions. They may be referenced only as opaque values in labels or marks. MMM defines no mechanisms for resolving or synchronising external state -- those are application responsibilities.}
\end{itemize}

Much collaborative knowledge work, especially in exploratory phases, does not require simultaneous editing at the level of characters or sentences --with the exception of final document polishing where fine-grained editing becomes more common. Even in real-time collaborative tools such as Google Docs, which resolve conflicts at the character level via Operational Transform \cite{OT}, collaborators often coordinate externally to avoid fine-grained interference with each other's edits (e.g. they partition work so that they each edit different sections of the document, or they use comments, suggestions, and annotations rather than directly modifying the same text). As a result, a large share of collaborative contributions can be expressed by adding new content rather than modifying existing material. Accordingly, MMM is designed so that core contribution operations -- new contributions, marks, and authorships -- are primarily additive. For non-public contributions, different mutable editing policies may be freely implemented locally (e.g. last-write-wins).
\medskip

This list of normative requirements is subject to future refinement. A more comprehensive specification may be published separately.

 \subsection{Evolution of the MMM data model} \label{evolution}

The present work defines the MMM data model core, while leaving certain aspects   (e.g.  the exact set of vertical edge types) deliberately underspecified for a short transition period. During this period, emerging practices and use cases will inform how these definitions are finally settled. After this transition, a stable complete specification of the data model will be published. 

Beyond this point, the data model is intended to remain largely unchanged, and to evolve only conservatively, with changes introduced sparingly, with careful attention to stability and backward compatibility, and only when strongly justified by practice as opposed to speculative needs or theoretical considerations.

This controlled evolution relies on a usage-driven extension mechanism, whose governance is a subject for future work. Before proposed changes to MMM can be considered for standardisation, they must first be implemented within the existing data model, using the MMM data model's own marking system (cf \S \ref{marks}). For instance, a new candidate edge type "is-a-proof-of" can initially be represented as a non-native "meta-type", using a semantic mark whose value is "is-a-proof-of". Only those extensions that have been tested and shown to be durably useful across implementations and contexts may later be considered for incorporation into the core MMM specification.
 \medskip

One planned refinement of the MMM data model (prior to stabilisation) concerns decentralisability (cf. \S~\ref{dimensions}). The constraints defined above (cf \S \ref{norms}) -- in particular immutability of core fields and predominantly monotone, append-only, irreversible operations  -- already align the data model with CRDT design principles \cite{shapiro2011crdt}. During stabilisation, merge behaviour will be formalised as a CRDT in support of  deterministic decentralised synchronisation.

\subsection{Interoperability}
\label{mmminterop}

Essentially, the present proposal is to base interoperability on 11 fields: {\it identifier, timestamp, status, authorships, marks, kind, type, label, source, target}, and {\it contents}. This small fixed set of structural fields is sufficient to wrap any information  a human can express as text or multimedia.
Expressivity of the MMM data model means many applications can represent their data in it. To have interoperability, applications must also agree on how to interpret and update representations. MMM ensures this through a set of normative constraints: stable identifiers, a fixed field set, a small set of native types with fixed semantics, and a planned CRDT-oriented merge design. This combination of expressive freedom and structural discipline means that any two conforming applications can exchange contributions without pairwise conversion, but still with some level of epistemic agreement.

\afterpage{
\begin{figure}[htbp]
    \centering
    \includegraphics[width=1\textwidth]{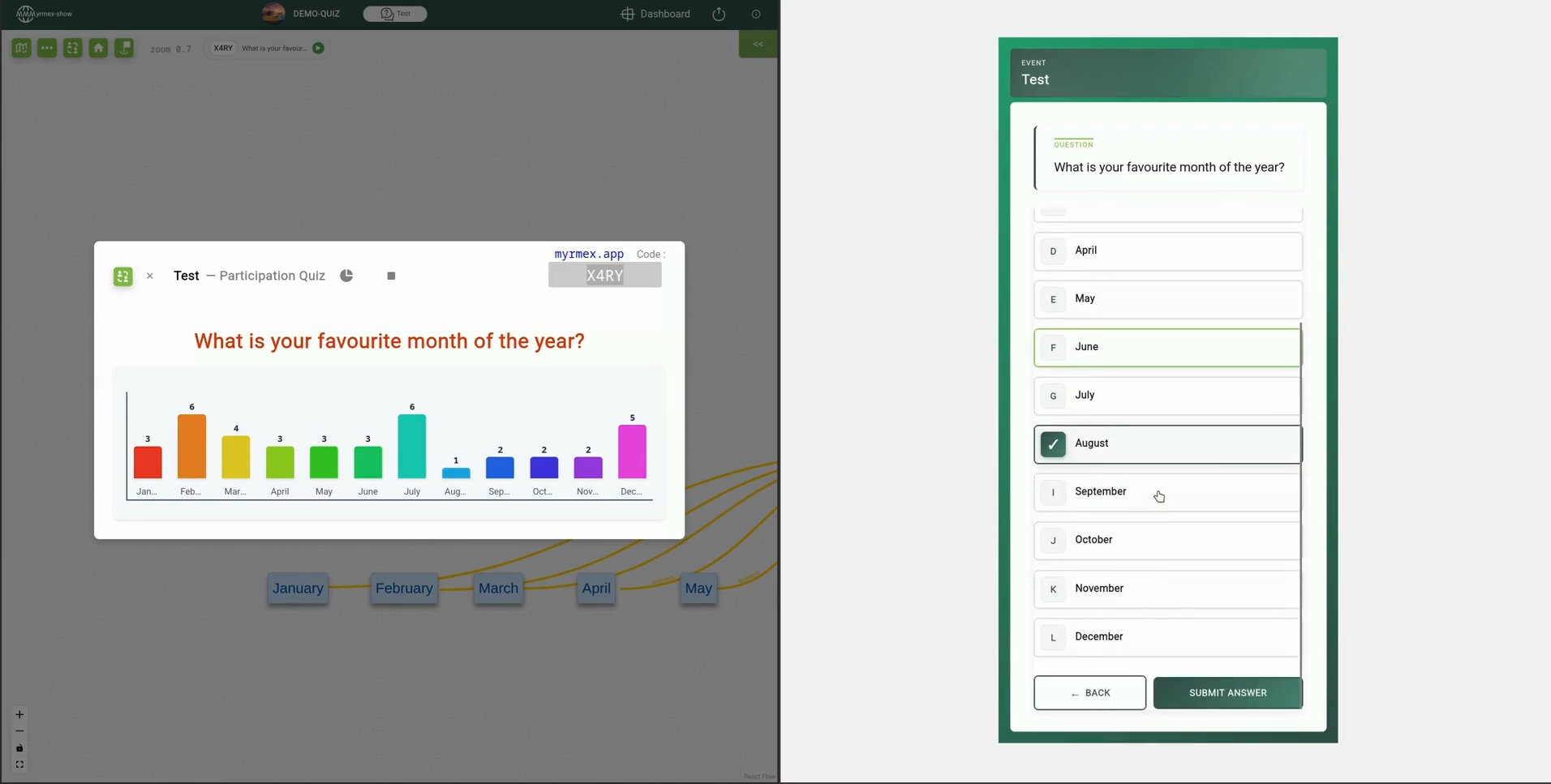}\\[-1.5cm]
    \includegraphics[width=0.46\textwidth]{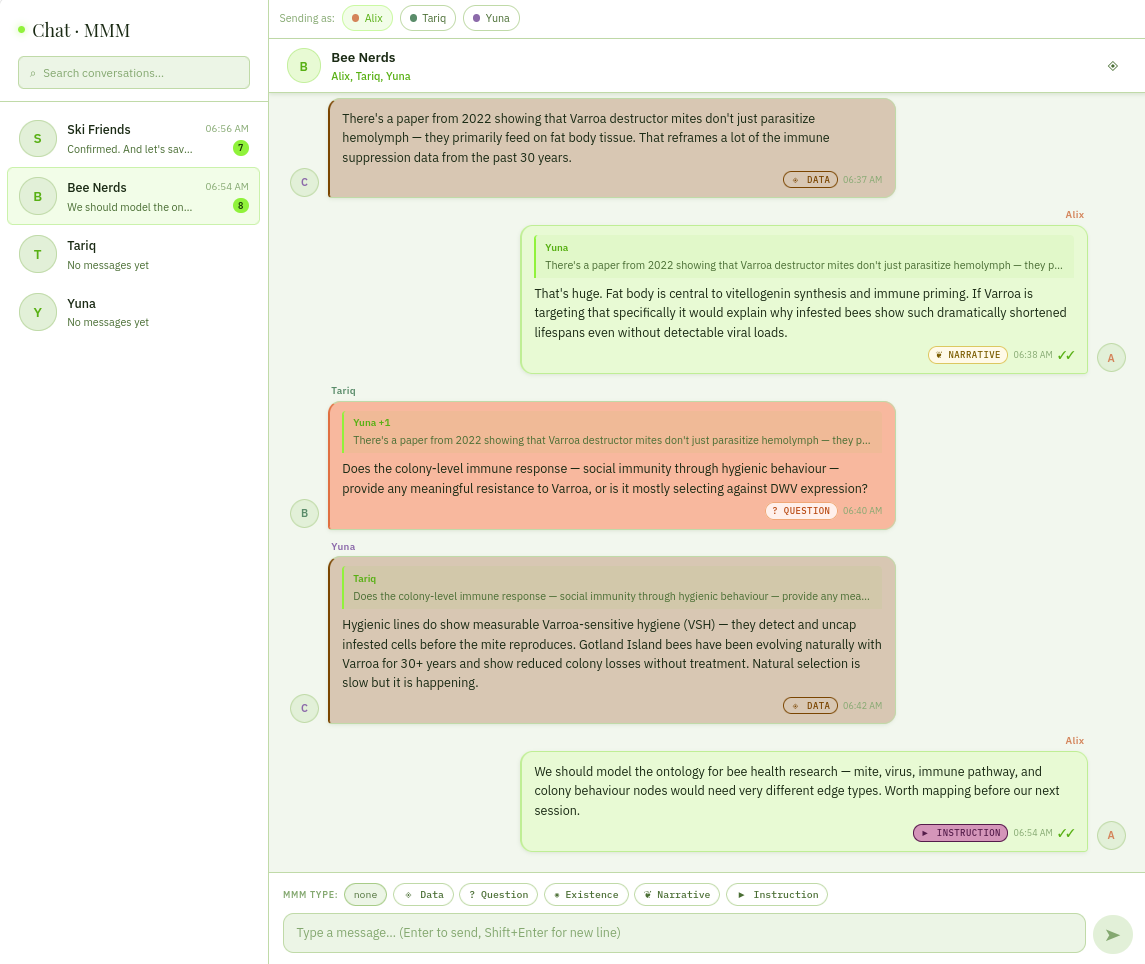}\includegraphics[width=0.45\textwidth]{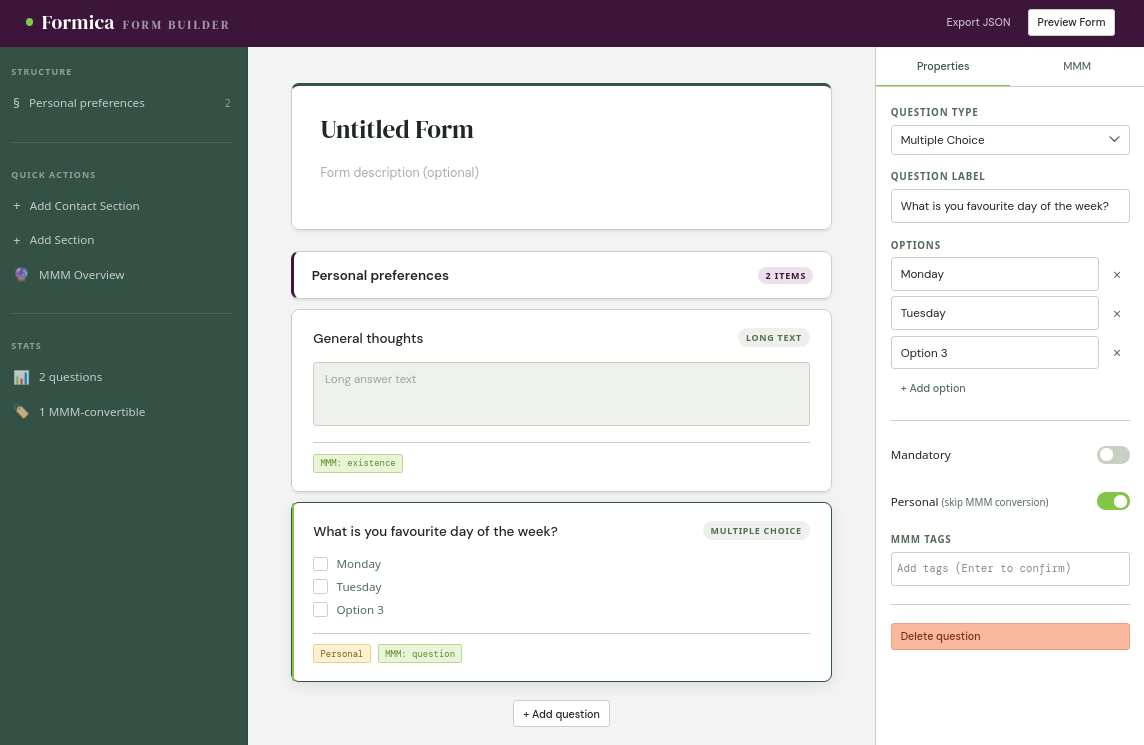}
    \caption{MMM-based prototype applications.
        Top: a functional research prototype for interactive presentation  (see also Fig~\ref{fig:filters} for another  functional research prototype, of a note-taking application).
        Bottom: messaging interface (left) and form builder (right) prototypes currently under development.
    }
    \label{fig:interop-apps}
\end{figure}}

In this respect, MMM's approach to interoperability is analogous to OpenStreetMap's: a lightweight normative spine (the 11 fields) that guarantees structural regularity, paired with flexible mechanisms (marks and free text) that accommodate local conventions and evolving semantics. Like OSM, the data model commits to a {\it formal container structure} fixing how information is {\it packaged} (structural commitment) without fixing the interpretation of the content (no semantic commitment). This makes the system loose enough to accommodate diverse applications, yet tight enough to support interoperability across any conforming implementation.

MMM is thus compatible with a wide range of application goals (without constraining persistence policies). Here is an illustrative, non-prescriptive list:
\begin{itemize}[label={\textbullet}]
 \item Traditional document editors -- handling single coarse-grained contributions, while allowing sections, comments or suggested edits to be decomposed into smaller linked contributions;

 \item PKM applications and mind-mapping tools -- persisting individual notes as separate typed contributions linked by typed relations;

 \item Communication and content platforms (social media, messaging, collaboration, email) -- optionally persisting posts, messages, comments, replies, and attachments as MMM contributions, linked to \texttt{Existence} vertices labelled with the discussion topic or email subject, or to \texttt{Question} vertices as responses, with typed relationships capturing replying, quoting, and cross-referencing, allowing communication streams to be partially persisted and progressively structured;

 \item Bibliography reference management tools -- associating notes, annotations, and comments to \texttt{Existence} vertices representing specific articles, and interrelating those notes across publications;

 \item Productivity and project management tools -- representing tasks, decisions, objectives, project names and rationale as connected MMM contributions;

 \item Educational tools -- capturing student questions, feedback, references to learning resources, and assignments as interconnected typed MMM contributions, and making the pedagogical relationship between concepts explicitly structured;

 \item Specialised professional (medical, legal, scientific) domain tools -- defining their own contribution workflows while still relying on the same underlying structural primitive.
\end{itemize}
Several of these application goals  have been prototyped on the MMM data model  (cf Fig.~\ref{fig:interop-apps}).

Importantly, it is up to each application to decide, among the information it processes, what is to be persisted in MMM form, and how  -- through the choice of contribution kinds and types, and interlinking strategies  -- according to its own domain and use case.

Applications conforming to the MMM data model then share a common underlying structure. This makes exchange and reuse of content between them more predictable than free-text-based exchanges. This holds provided implementations consistently apply the data model and maintain identifier stability across creation, import, export, and migration.

The MMM data model's deliberately unopinionated design is intended as a satisfactory baseline for interoperability. Should it prove, in practice, insufficiently constrained to support meaningful coordination across interfaces, the refinement mechanism described in Section \ref{evolution} provides a path for progressively strengthening the model and addressing those limitations based on observed usage.  

\section{Pilot Deployment}
\label{sec:pilot}

This section describes two distinct academic usage
contexts of the same MMM deployment: an interdisciplinary research setting,
and a repeated pedagogical exercise within the CapECL formation at
École Centrale de Lyon (projects CREPE 2025 and CREPE 2026). Both ran on the reference
implementation, Myrmex (\href{https://myrmex.app}{myrmex.app}), described below.
The Crepe exercise was designed and run independently by the instructors. The researcher usage was self-directed.

No active user recruitment was conducted. The deployment reflects organic adoption 
emerging from the author's academic and scientific network, which explains its current small scale.

\subsection{Reference Implementation and Other Prototypes}
\label{refapp}

Myrmex is a web application implementing the full MMM data model as specified in
Section~\ref{mmm}. It is built on a TypeScript/React frontend using ReactFlow for
graph-based visualisation of MMM contributions, and a Node.js/Express backend persisting
data in a MariaDB relational database. Contributions can be imported and exported as JSON.
 These implementation choices -- relational storage, JSON serialisation -- are specific to Myrmex and not requirements of the data model itself, which is compatible with any storage backend provided content conforms to the MMM data model specification. 
 Determining optimal persistence layer design is planned research work, to be informed by observed patterns of use.
\medskip

To enable third party experimentation and development, the reference codebase
is under active refactoring in preparation for open-source release. A public API conforming to the MMM data model will be published alongside the codebase, enabling third-party implementations and independent deployments. 
\medskip

Beyond the primary Myrmex knowledge mapping application,
several additional MMM-based prototype and demonstrator applications have been developed on the same data model, including
a note-taking tool (cf Fig~\ref{fig:filters}),
an interactive presentation tool,
a messaging interface,
and a form builder 
(cf Fig.~\ref{fig:interop-apps}),
demonstrating that the reference application is only one possible implementation: functionally heterogeneous applications can 
be built on a single MMM substrate, all of them producing MMM formatted data they can parse and use without pairwise conversion.   
At this small scale of development, we have found that relying on a common single fixed underlying substrate has simplified  the design and development of  independent, diverse-purpose, interoperable applications.
Frontend development has become a separate concern, decoupled from the underlying data management logic, which can be the same for all applications. This has not yet been verified at scale.

\paragraph{Prototype maturity.}
The reference implementation is under active development and should be understood as a
research prototype rather than a production system. Usage data reflects interactions with
a system that has known limitations in performance, user experience, and features.
Observed usage patterns described below
may in part reflect interface friction.

\begin{table}[htbp]
    \centering
    \begin{tabular}{|l|l|r|r|}
        \hline
        \textbf{User}      & \textbf{Kind} & \textbf{Count} & \textbf{\% of user total} \\
        \hline
        \multirow{3}{*}{Researcher 1}
                           & Edge          & 2043           & 53.9                      \\
                           & Vertex        & 1554           & 41.0                      \\
                           & Pen           & 195            & 5.1                       \\
        \rowcolor{gray!15} & Total         & 3792           & 100                       \\
        \hline
        \multirow{3}{*}{Researcher 2}
                           & Edge          & 566            & 52.0                      \\
                           & Vertex        & 515            & 47.3                      \\
                           & Pen           & 8              & 0.7                       \\
        \rowcolor{gray!15} & Total         & 1089           & 100                       \\
        \hline
    \end{tabular}\smallskip
    \caption{MMM contribution counts in total and by kind for two research users}
    \label{tab:user-kinds}
\end{table}

\subsection{Research usage setting} 
\label{pastor}
The first usage context involved two researchers affiliated to different French CNRS laboratories: Researcher~1, specialised in applied mathematics and signal processing, and Researcher~2, specialised in systems biology and immunology -- both working on questions related to the interdisciplinary field of Complex Systems.
The two researchers interacted over an extended period, with a plan to co-publish. Researcher~1 used Myrmex regularly to document his work and its relations to Researcher 2's -- translating joint discussions, results, and evolving understanding into MMM contributions -- while Researcher~2's direct usage of the platform was more episodic, primarily in preparation of presentations and of a joint article.
Researcher~1's largest graph (905 landmarks) served as the primary MMM documentation space for this collaborative research thread and eventually fed directly into a peer-reviewed publication~\cite{dp2026}. Myrmex has a "Scenarios" feature that allows the user to define lists of contributions/subgraphs and to name and display them sequentially. Researcher~1 reports having used his big MMM graph, and in particular the Scenarios feature, to organise and filter information, and to structure the final article and decide what content went into it.
Researcher~2 engaged with this workspace as part of the collaboration, demonstrating that MMM-documented epistemic reasoning is legible and usable by a collaborator beyond its primary author. The published article constitutes direct evidence that MMM-formatted research documentation is compatible with and directly  supportive of
standard scientific output.

Tables~\ref{tab:user-kinds} and \ref{tab:user-types} detail the per-user distributions for
the two research users. Both users made substantive use of \texttt{\texttt{Existence}} vertices
alongside \texttt{\texttt{Narrative}}
(19.5\% and 20.3\% of vertices respectively), indicating that
the concept-level granularity of the data model was actively exploited rather than flattened
into undifferentiated prose.

\begin{figure}[htbp]
    \centering
    \begin{tabular}{cc}
        \includegraphics[width=0.5\textwidth]{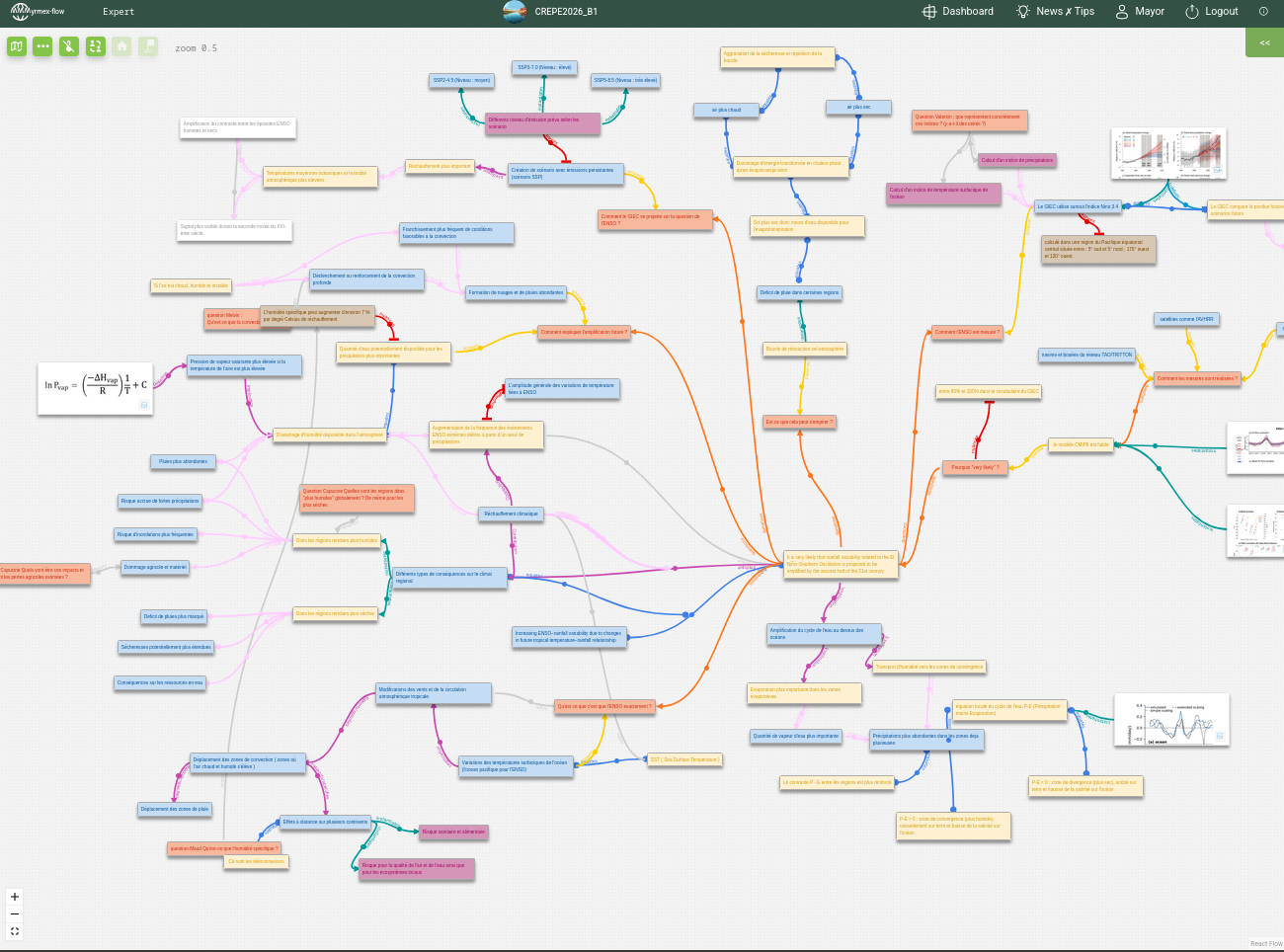} &
        \includegraphics[width=0.5\textwidth]{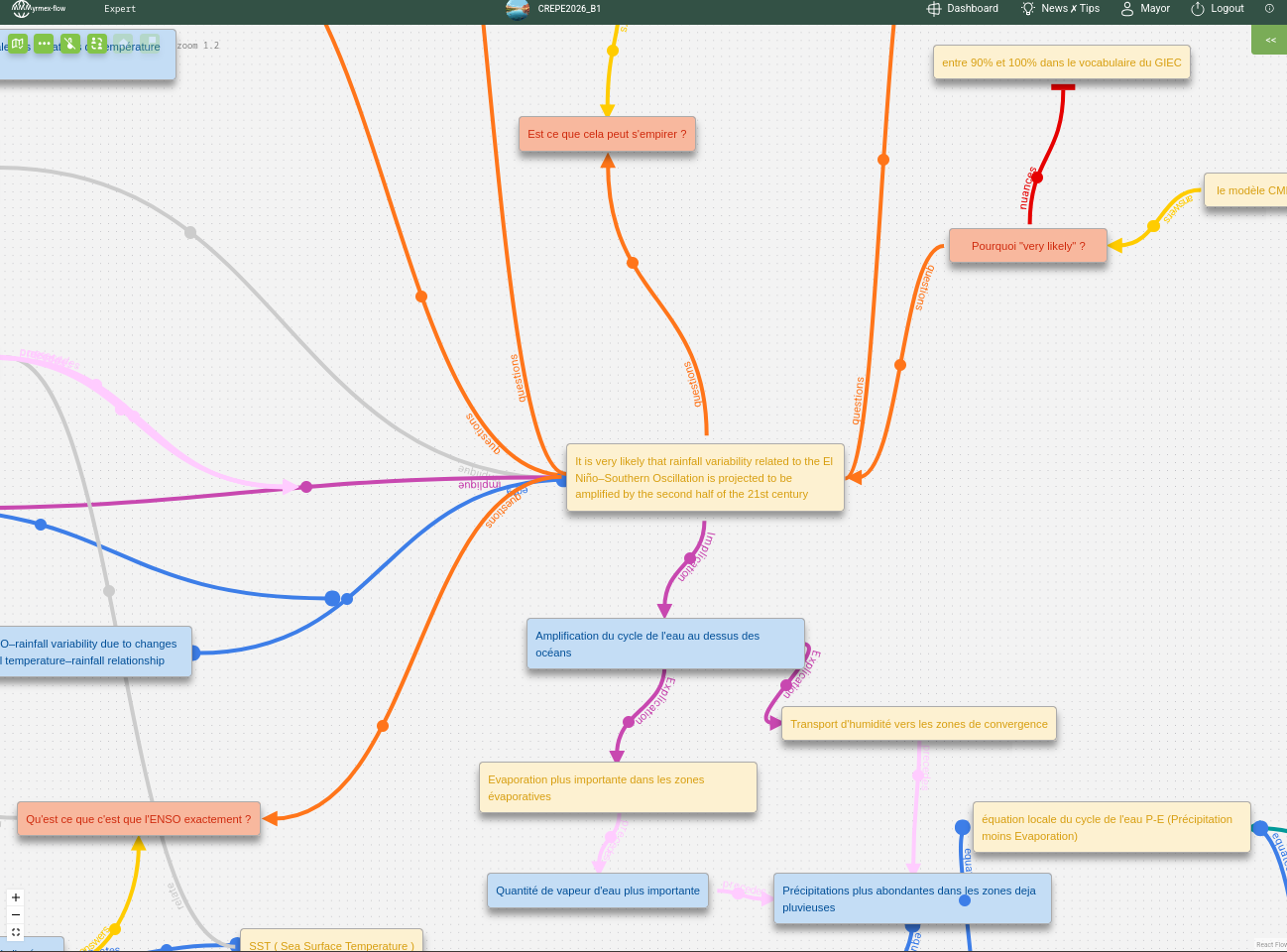}
    \end{tabular}
    \caption{An example of CREPE 2026 student graph strongly relying on MMM edge types although not consistently relying on the normative semantics of typed edge directions.}
    \label{fig:B1}
\end{figure}

\subsection{Pedagogical usage setting}
\label{crepe}

The second usage context is a repeated educational exercise designed by higher education physics instructors, conducted over two consecutive years (2025 and 2026) within CapECL a satellite engineering programme of École Centrale de Lyon based in Saint-Étienne, France. The project is called "CREPE".  The exercise consists of a conceptual and bibliographic mapping of claims taken from the IPCC Sixth Assessment Report Summary for Policymakers (2021) \cite{RN2}. Both iterations involved approximately 50 students divided into 16 groups of three to four, each combining first-year (L1) and second-year (L2) students. Each group was tasked with selecting an assertion from the report and collectively mapping their conceptual and bibliographic understanding of it (its related questions, answers, {challenges}, and supporting sources) as an MMM graph. An example assertion studied by one group is the following:

\begin{quote}
    \textit{"Emissions reductions in 2020 associated with measures to reduce the spread of COVID-19 led to temporary but detectable
        effects on air pollution (high confidence) and an associated small, temporary increase in total radiative forcing, primarily
        due to reductions in cooling caused by aerosols arising from human activities (medium confidence)."  {\normalfont  -- IPCC AR6, D.2.1
        \cite{RN2}}}
\end{quote}

\begin{figure}[htbp]
    \centering
     \includegraphics[width=1\textwidth]{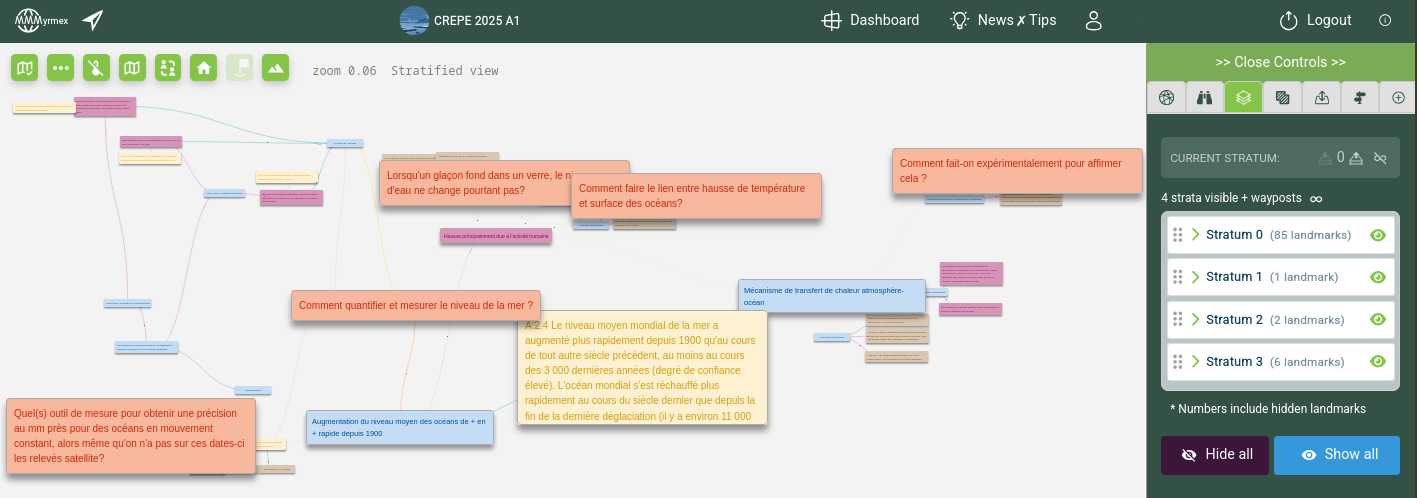}
    \caption{The stratified zooming in Myrmex makes use of {\tt Stratum} marks (cf~\S\ref{marks})}
    \label{fig:strata}
\end{figure}
The exercise has three phases. In the first, each group documents their chosen assertion. In the second, each group is assigned one or two other groups' graphs and must formulate \texttt{Question}s about that work based solely on what they can read in the MMM graph
(a test of the \dimname{Epistemic Structure} and \dimname{Post-Document Organisation} properties of the data model, cf~\S~\ref{dimensions}).
Instructors also contributed \texttt{Question}s in MMM form directly into student graphs.
In 2025, teachers assigned their own \texttt{Question}s higher-valued stratum marks to help students identify them.
A mouse wheel controlled functionality of myrmex.app allows users to reveal different strata, with contributions of different strata rendered at different sizes (cf Fig.~\ref{fig:strata}).
In the third phase, prepared by the \texttt{Question}s on their graph that were asked by outsiders, students make an oral presentation of their work to the entire class.

The first-year cohort of  2025  became the second year students of  2026,  who were joined by a  new first-year cohort, such that approximately half of the 2026 participants had prior exposure to the tool, albeit in an earlier version. 
Three students from the 2025 cohort independently chose to adopt myrmex.app for their own work during the intervening year, without any assignment requiring them to do so, and despite its
status as an early-stage working prototype with known limitations, functional instability, and unresolved bugs
-- a form of voluntary re-engagement that, while anecdotal, aligns with Researcher~1's own unprompted regular use, suggesting the tool provided sufficient standalone value to motivate reuse by some  (\dimname{Immediate Individual Value}).

Sizes of individual graph  created by students range from 51 to 221 landmarks in 2025 (mean: 104) and from 96 to 216 in 2026 (mean: 155). The 45\% increase in mean graph size from 2025 to 2026 may reflect  greater familiarity with the tool among instructors and returning students, as well as increased tool maturity.

The 2026 exercise was still under way at the time of writing: the first phase had been completed by all groups, but the second phase   (the cross-group questioning)   just started. This explains the slightly lower proportion of \texttt{Question} vertices
in 2026 relative to 2025 (13.2\% of vertices in 2025, 10.9 in 2026, cf Table~\ref{tab:user-types}).

\begin{table}[htbp]
    \centering
    \begin{tabular}{|l|l|r|r|}
        \hline
        \textbf{Year}                    & \textbf{Kind} & \textbf{Count} & \textbf{\% of year total} \\
        \hline
        \multirow{2}{*}{2025}
                                         & Edge          & 873            & 53.1                      \\
                                         & Vertex        & 772            & 46.9                      \\
        \hline
        \rowcolor{gray!15}
        \multicolumn{2}{|l|}{Total 2025} & 1{,}645       & 100.0                                      \\
        \hline
        \multirow{2}{*}{2026}
                                         & Edge          & 1{,}247        & 52.8                      \\
                                         & Vertex        & 1{,}116        & 47.2                      \\
        \hline
        \rowcolor{gray!15}
        \multicolumn{2}{|l|}{Total 2026} & 2{,}363       & 100.0                                      \\
        \hline
        \multirow{2}{*}{Combined}
                                         & Edge          & 2{,}120        & 53.0                      \\
                                         & Vertex        & 1{,}888        & 47.0                      \\
        \hline
        \rowcolor{gray!15}
        \multicolumn{2}{|l|}{Total}      & 4{,}008       & 100.0                                      \\
        \hline
    \end{tabular}
    \smallskip
    \caption{Landmark counts by kind in the CREPE pedagogical setting, 2025 and 2026.
        No pen contributions were recorded in CREPE workspaces.}
    \label{tab:crepe-kinds}
\end{table}

Students received  minimal instructions on how to use the application toward achieving  the exercise's pedagogical and scientific goals, and no tutorial on the application itself -- a level of guidance that some students found insufficient. Within the first two hours of the documentation session, 33 students had contributed 350 landmarks, some of them with no prior experience of the application. A similar effect was observed in 2025.
This rapid uptake among first- and second-year engineering students -- a population partially accustomed to mental mapping tools according to their instructors -- suggests that the contribution model presents a low barrier for users already familiar with graph-based knowledge organisation
(\dimname{Adoption Ease} and \dimname{Human Primacy}).
\smallskip

\subsection{Observations}
\label{observations}

\begin{table}[h]
    \centering
    \begin{tabular}{|l|r|r|}
        \hline
        \textbf{Kind}  & \textbf{Count} & \textbf{\% of total} \\
        \hline
        Edge           & 5704           & 53.1                 \\
        Vertex         & 4754           & 44.2                 \\
        Pen            & 292            & 2.7                  \\
        \hline
        \textbf{Total} & \textbf{10750} & \textbf{100.0}       \\
        \hline
    \end{tabular}
    \smallskip
    \caption{Landmark counts by kind across the full Myrmex deployment.}
    \label{tab:full-kinds}
\end{table}

\paragraph{Overall scale.}
As of the writing of this paper, the Myrmex deployment contains 10\,750 MMM landmarks across
all users and workspaces, comprising 51 real-world workspaces (graphs) with at least 25 landmarks and
a mean size of over 210 landmarks per graph (including edges), spanning domains from signal processing, to arithmetic, systems biology,   climate science,  mechanics, and European asylum law.

Across the 51 workspaces, contributors expressed the diversity  of knowledge forms that the data model is designed to support -- informal natural-language observations in free text labels, structured mathematical results and formulae in KaTeX, bibliographic references, and pedagogical annotations and questions -- without requiring any extension to the native vocabulary (\dimname{Universal Scope}).

\paragraph{Mathematical notation.} 
Thirty-three of the 51 workspaces contain at least one vertex whose label contains a KaTeX mathematical formula. Overall, 288 of the 4\,754 vertex contributions (6.1\%) contain such formulae, including 156 authored by CREPE students, 52 by Researcher~1, and 51 by physics and mathematics instructors preparing mechanics, thermodynamics, and arithmetic lessons with Myrmex.
 The reference implementation's fixed node display size causes long formulae to be truncated, which has been reported as discouraging the documentation of complex mathematical expressions.

\begin{figure}
    \centering
      \includegraphics[width=0.7\textwidth]{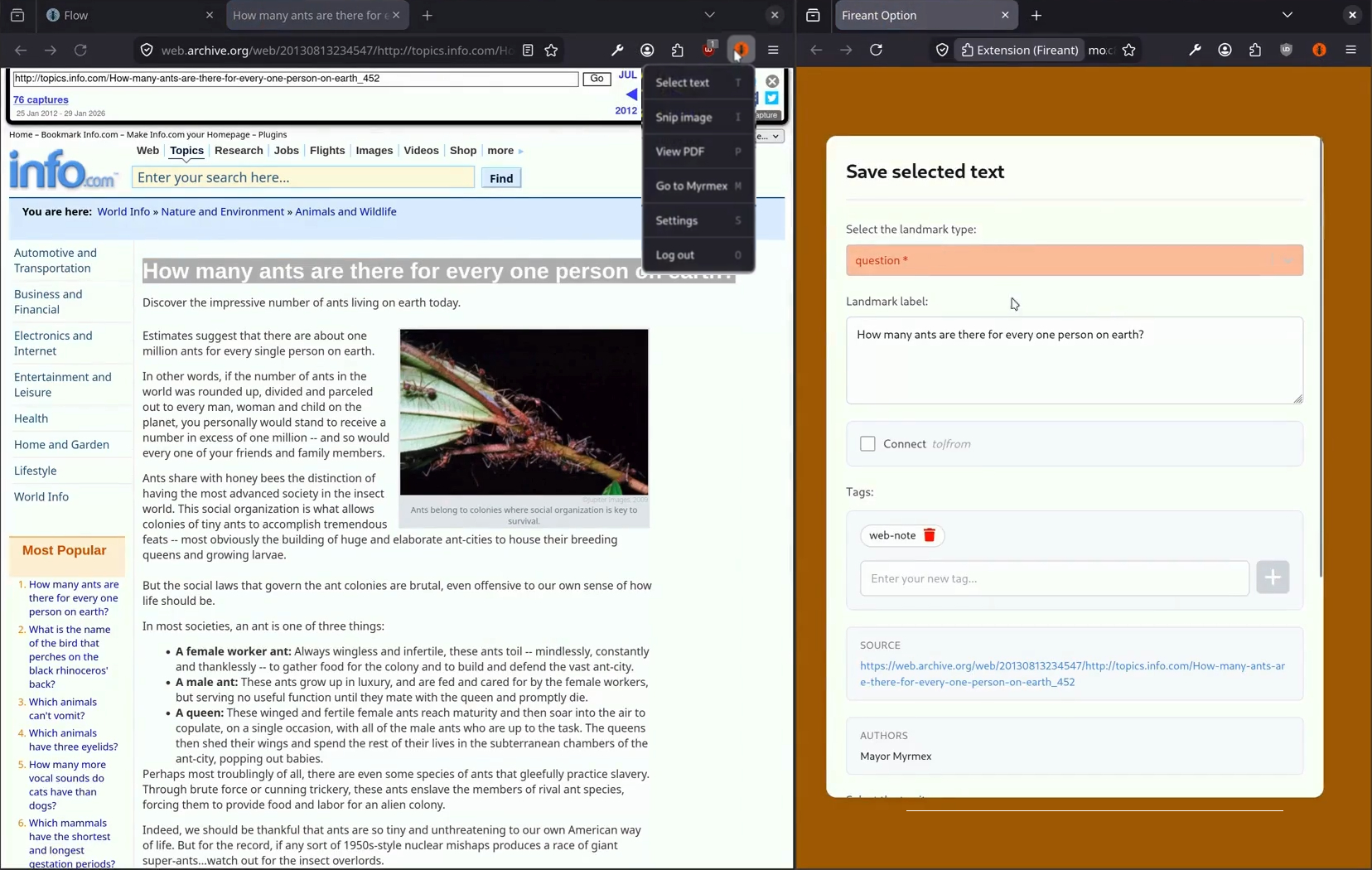}
    \caption{An MMM browser extension to  extract text and images from external sources, keeping references to original sources embedded in the new MMM contributions.}
    \label{fig:fireant}
\end{figure}

\paragraph{External resources and attachments.}
Of the 4\,754 vertex contributions, 544 (11.4\%) carry an attachment or hyperlink: 236 are embedded images (43.4\%), 247 are hyperlinks to external web resources including online documents and PDFs (45.4\%), and 61 are other attached files (e.g. PDF). 
8.4\% of CREPE student vertices, and 11.3\% of Researcher 1's vertices have attachments or hyperlinks,
indicating active use of the provided mechanisms to ground claims in external documentary sources.
Among the embedded images, 15 were captured via "Fireant", a momentarily available prototype Firefox browser extension enabling users to select text and take snapshots from web pages and PDF files and send them directly to a Myrmex graph (cf Fig. \ref{fig:fireant}). A further 72 are named  "{\it Capture d'écran <timestamp>.png}" (screenshot). This suggests %
a practical need for seamless capture of external content and the feasibility of browser-level interoperability with MMM -- a possibility the existence of "Fireant" demonstrates.

\paragraph{Questions.} 
The CREPE student {\tt Questions}, including those produced during the cross-group phase illustrate the epistemic range the data model accommodates. They span formal mathematical reasoning ("\textit{Why use a Poisson distribution here, and how is $\lambda$ calculated?}"), physical mechanism ("\textit{What physical mechanism explains this inertia?}"), and epistemological reflection on scientific language and method ("\textit{Why `very likely'?}")
 all coexisting as first-class \texttt{Question}  vertices within the same MMM graph
(\dimname{Universal Scope}). 

Unlike neutral information requests, {\tt Question}s such as "\textit{But AR6 says 7\%; why?}", "\textit{Yet theoretical wind speeds are almost never reached in reality.}", or "\textit{You say that even if emissions stopped immediately, the thermal expansion of the oceans would continue for several centuries, but then what physical mechanism explains this inertia?}" only exist because a specific student noticed a specific discrepancy, felt a specific doubt, or wanted to challenge a specific formulation from their own reasoning (\dimname{Expression Intent}). Some questions explicitly solicit the other group's own interpretation rather than a factual or authoritative answer: "\textit{In your opinion, why do articles alternate between these two quantities? What might be the benefit of keeping two quantities that are so similar yet distinct to describe the same situation?}" These are expressions of individual epistemic states
that no encyclopedic or convergent system would accommodate as valid contributions  in its primary knowledge space.

\begin{figure}
    \centering
      \includegraphics[trim={9mm 18mm 1mm 3cm},clip,width=1\textwidth]{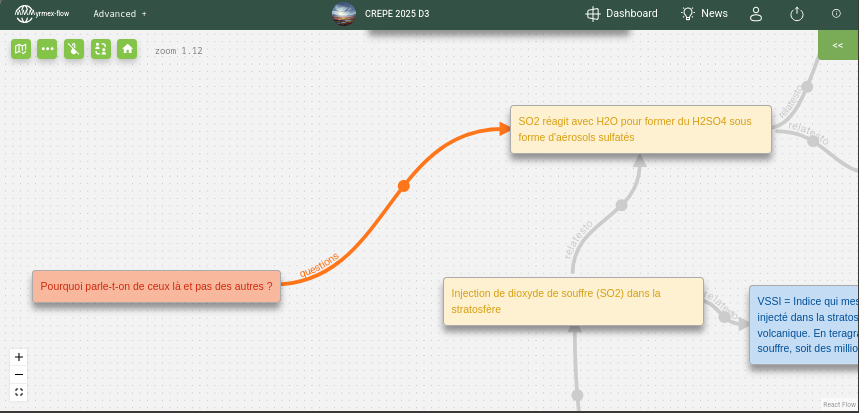}
    \caption{A student question asking "{\it Why are we discussing those ones and not the others?}".}
    \label{fig:pourquoi}
\end{figure}

\paragraph{Redundancy.} 
Two groups independently formulated near-identical {\tt Question}s  about ice cube melting from slightly different angles ("\textit{When an ice cube melts in a glass, the water level doesn't change, does it?}" / "\textit{And yet, when an ice cube melts in a glass of water, the water level doesn't rise from start to finish, does it?}") (\dimname{Redundancy Friendliness}).
The same pattern appears at the concept level: four students across different groups and years independently created \texttt{\texttt{Existence}} vertices expressing equivalent concepts -- "\textit{ERF Négatif}", "\textit{Forcage radiatif (ERF) négatif}", "\textit{influence humaine}", "\textit{Human influence}" -- in different terms and different languages.

\begin{figure}[htbp]
    \centering
    \includegraphics[trim={0 3cm 5mm 4.5cm},clip,width=1\textwidth]{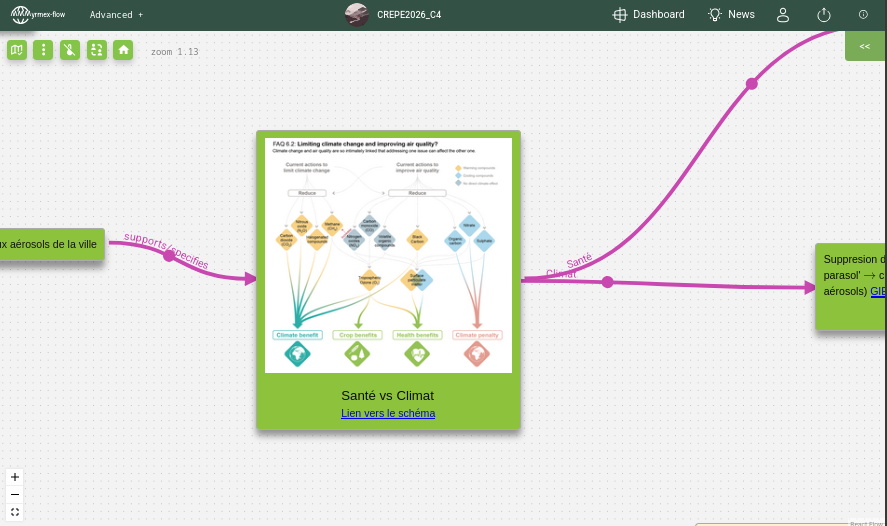}
    \caption{Detail of a CREPE 2026 student workspace illustrating voluntary enrichment beyond the assignment requirements: image attachments and hyperlinks to original web sources added spontaneously by students.   }
    \label{fig:links}
\end{figure}

\begin{table}[h]
    \centering
    \begin{tabular}{|l|l|r|r|}
        \hline
        \textbf{Kind}      & \textbf{Type}           & \textbf{Count} & \textbf{\% of kind} \\
        \hline
        \rowcolor{gray!15}\multirow{6}{*}{Vertex}
                           & \texttt{Narrative}               & 2513           & 52.9                \\
                           & \texttt{Existence}               & 949            & 20.0                \\
                           & \texttt{Data}                    & 533            & 11.2                \\
                           & \texttt{Question}                & 433            & 9.1                 \\
                           & \texttt{Instruction}             & 326            & 6.9                 \\
        \hline
        \rowcolor{gray!15}\multirow{16}{*}{Edge}
                           & \texttt{RelatesTo}               & 1659           & 29.1                \\
                           & \texttt{Pertains}                & 1623           & 28.5                \\
                           & \texttt{Instantiates}            & 553            & 9.7                 \\
                           & \texttt{Answers}                 & 538            & 9.4                 \\
                           & \texttt{Equates}                 & 390            & 6.8                 \\
                           & \texttt{Questions}               & 300            & 5.3                 \\
                           & \texttt{Substantiates}  & 261            & 4.6                 \\
                           & \texttt{Challenges}                 & 187            & 3.3                 \\
                           & Precedes                & 88             & 1.5                 \\
        \rowcolor{gray!15} & \texttt{Relate}                  & 77             & 1.4                 \\
                           & \texttt{Differ}             & 28             & 0.5                 \\
        \hline
    \end{tabular}\smallskip
    \caption{Landmark counts by kind and type across the full Myrmex deployment.}
    \label{tab:full-types}
\end{table}

\paragraph{Individual value \& Collaboration.} 
The deployment illustrates MMM's ability to support both individual and collaborative work within the same system.
Of the 51 graphs, 8 were produced by a single author -- including  Researcher~1's two largest workspaces (905 and 464 landmarks) produced over an extended period of regular use -- demonstrating that the system delivers standalone value without requiring any prior community or network, capable of sustaining Researcher~1's deep individual knowledge documentation work over time (\dimname{Immediate Individual Value}). 

The remaining 43 workspaces involve between 2 and 8 contributors, including a three-author conceptual mapping of a pseudo-scientific article on climate systems in which the author was involved (442 landmarks).

\paragraph{(Co-)Authorship.}
Beyond their primary authoring teams, most workspaces in the deployment received contributions from at least one external 'commentator' -- a user who either only added new edges to the graph,
or  only added edges and  \texttt{Question} vertices. Commentators used \texttt{Questions}, \texttt{Challenges}, and \texttt{\texttt{RelatesTo}} edge types.
In the CREPE exercise, workspaces have 
up to 9 co-authors  actively contributing to a single shared graph: 3 primary authors and 6 external commentators.
In line with the \dimname{Homogeneous data space} dimension, this shows that in practice, {challenges}, questions, and annotations coexist as first-class MMM landmarks alongside the content they concern, navigable and queryable by the same mechanisms, with no structural boundary between the knowledge layer and the discussion layer.

\paragraph{Edge-to-vertex ratio.}
The edge-to-vertex ratio is strikingly stable across use cases: stabilising around 53/47 regardless of user expertise, usage context, or year.
This consistency   suggests a possible structural balance as a natural outcome of MMM-formatted knowledge work.

\begin{table}[h]
    \centering
\begin{tabular}{|l|l||r|r|r|r|r|r|r|r|}
        \hline
        \textbf{Kind} &
        \textbf{Type} &
        \multicolumn{2}{c|}{\textbf{CREPE 2025}} &
        \multicolumn{2}{c|}{\textbf{CREPE 2026}} &
        \multicolumn{2}{c|}{\textbf{Researcher 2}} &
        \multicolumn{2}{c|}{\textbf{Researcher 1}} \\
        \cline{3-10}
        & &
        \textbf{Count} & \multicolumn{1}{c|}{\textbf{\%}} &
        \textbf{Count} & \multicolumn{1}{c|}{\textbf{\%}} &
        \textbf{Count} & \multicolumn{1}{c|}{\textbf{\%}} &
        \textbf{Count} & \multicolumn{1}{c|}{\textbf{\%}} \\
        \hline

        \rowcolor{gray!15}\multirow{5}{*}{Vertex}
        & \texttt{Narrative}      & 419 & 54.3 & 557 & 49.9 & 375 & 72.8 & 757 & 48.7 \\
        & \texttt{Existence}      & 105 & 13.6 & 182 & 16.3 & 115 & 20.3 & 303 & 19.5 \\
        & \texttt{Data}           & 95  & 12.3 & 219 & 19.6 & 4   & 0.7  & 168 & 10.8 \\
        & \texttt{Instruction}    & 51  & 6.6  & 36  & 3.2  & 9   & 1.6  & 197 & 12.7 \\
        & \texttt{Question}       & 102 & 13.2 & 122 & 10.9 & 12  & 2.1  & 129 & 8.3 \\
        \hline

        \multirow{12}{*}{Edge}
        & \texttt{Pertains}       & 87  & 10.0 & 139 & 11.1 & 5   & 0.9  & 1208 & 59.1 \\
        & \texttt{Equates}        & 15  & 1.7  & 88  & 7.1  & -   & -    & 246  & 12.0 \\
        & \texttt{Answers}        & 118 & 13.5 & 195 & 15.6 & 4   & 0.7  & 135  & 6.6  \\
        & \texttt{Questions}      & 65  & 7.4  & 37  & 3.0  & 26  & 4.6  & 130  & 6.4  \\
        & \texttt{Instantiates}   & 61  & 7.0  & 272 & 21.8 & 23  & 4.1  & 101  & 4.9  \\
        & \texttt{Challenges}        & 9   & 1.0  & 55  & 4.4  & 3   & 0.5  & 86   & 4.2  \\
        & \texttt{Substantiates}  & 39  & 4.5  & 56  & 4.5  & 2   & 0.4  & 72   & 3.5  \\
        \rowcolor{gray!15} & \texttt{RelatesTo}     & 437 & 50.1 & 298 & 23.9 & 499 & 88.2 & 48   & 2.4 \\
        & \texttt{Differ}    & 1   & 0.1  & 7   & 0.6  & 1   & 0.2  & 10   & 0.5 \\
        \rowcolor{gray!15} & \texttt{Relate}        & 38  & 4.4  & 27  & 2.2  & -   & -    & 4    & 0.2 \\
        & Precedes       & 3   & 0.3  & 73  & 5.9  & 3   & 0.5  & 3    & 0.2 \\
        \hline
    \end{tabular} \smallskip
    \caption{Landmark counts by kind and type for two research users and for  2025 and 2026 iterations of the CREPE pedagogical exercise. Percentages are relative to number of landmarks of same kind.  }
    \label{tab:user-types}
\end{table}

\paragraph{Fallback edge type usage as an engagement indicator.}
The proportion of \texttt{\texttt{RelatesTo}} and \texttt{Relate} edges -- the two fallback types for
unqualified relationships (cf~\S~\ref{mmmtypes}) -- varies substantially across users
and contexts,  revealing users' engagement with the epistemic
typing system. Across the full deployment, these two types together account for 30.5\% of
all edges. Researcher~1, the most experienced user, uses them for only 2.5\% of his edges,
relying instead on epistemically specific types throughout. His primary research workspace
(905 landmarks) contains no fallback edges at all. Researcher~2's rate of
88.2\% likely reflects primarily her use of the automatic Markdown-to-MMM conversion functionality
in the reference implementation, which imports unstructured text as \texttt{\texttt{Narrative}} and
\texttt{\texttt{Existence}} vertices connected by \texttt{RelatesTo} edges. In the CREPE student setting, the rate drops from 54.5\% in 2025 to 26.1\% in 2026, possibly due to familiarity with MMM typing of half of the students or to varying wording of instructions by instructors.

The per-workspace distribution further illustrates this pattern. With the exception of two
CREPE 2026 groups, all student workspaces exhibit less than 46\% fallback edges, and more
than half fall below 23\%. Since the MMM typing system is intentionally non-coercive (cf \S\ref{mmmtypes} and \S~\ref{norms}),
these figures suggest that the type vocabulary is  accessible and that users engage with it voluntarily and substantively rather than systematically defaulting to the least constrained option available.

\afterpage{
\begin{figure}[htbp]
    \centering
    \includegraphics[width=1\textwidth]{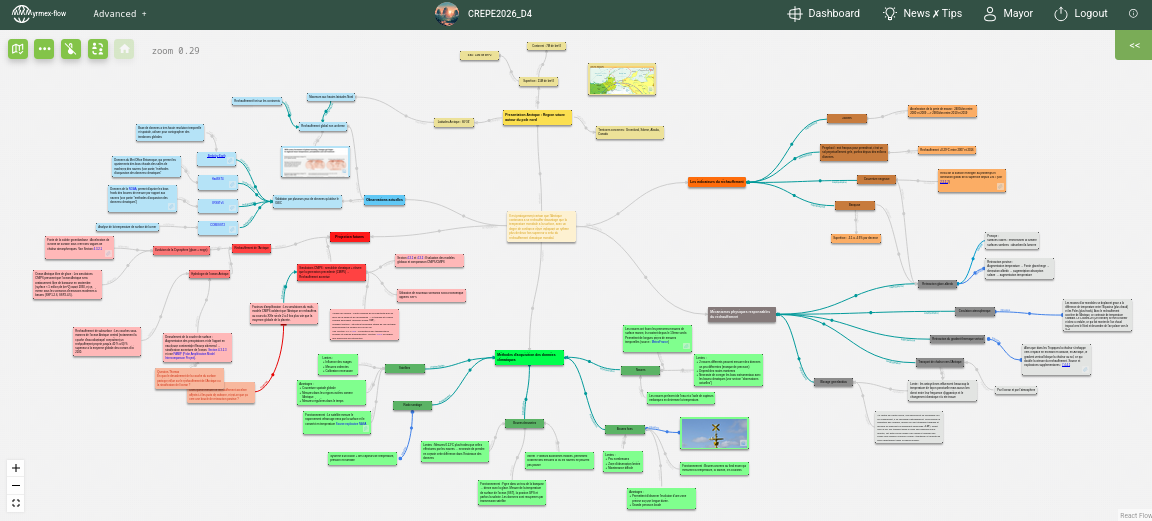}
    \caption{CREPE 2026 group D4 workspace: an example of epistemic structure expressed
        primarily through visual styling and layout rather than MMM edge type conventions.}
    \label{fig:D4}
\end{figure}
}

\begin{figure}[htbp]
    \centering    \parbox{0.49\textwidth}{
        \includegraphics[width=0.492\textwidth]{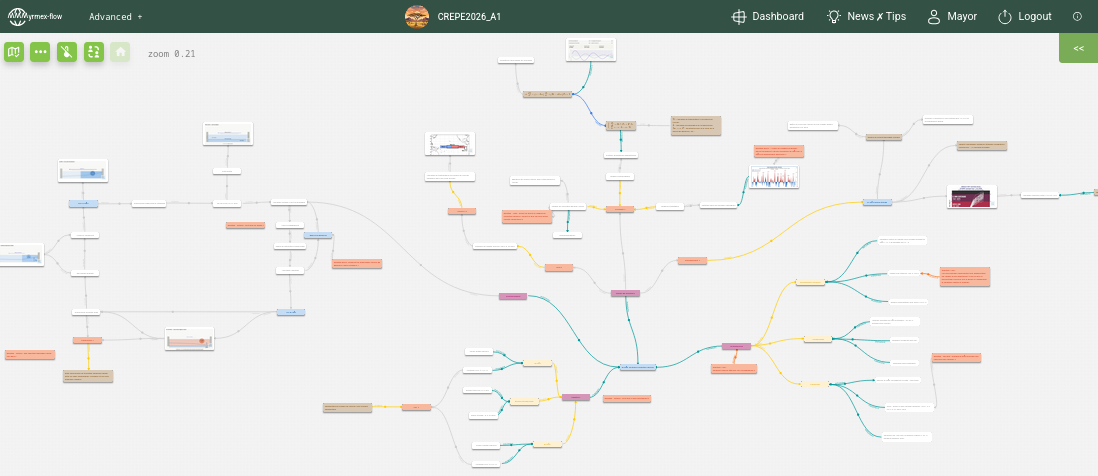}\\ 
        \includegraphics[trim={0 25mm 0 0},clip,width=0.492\textwidth]{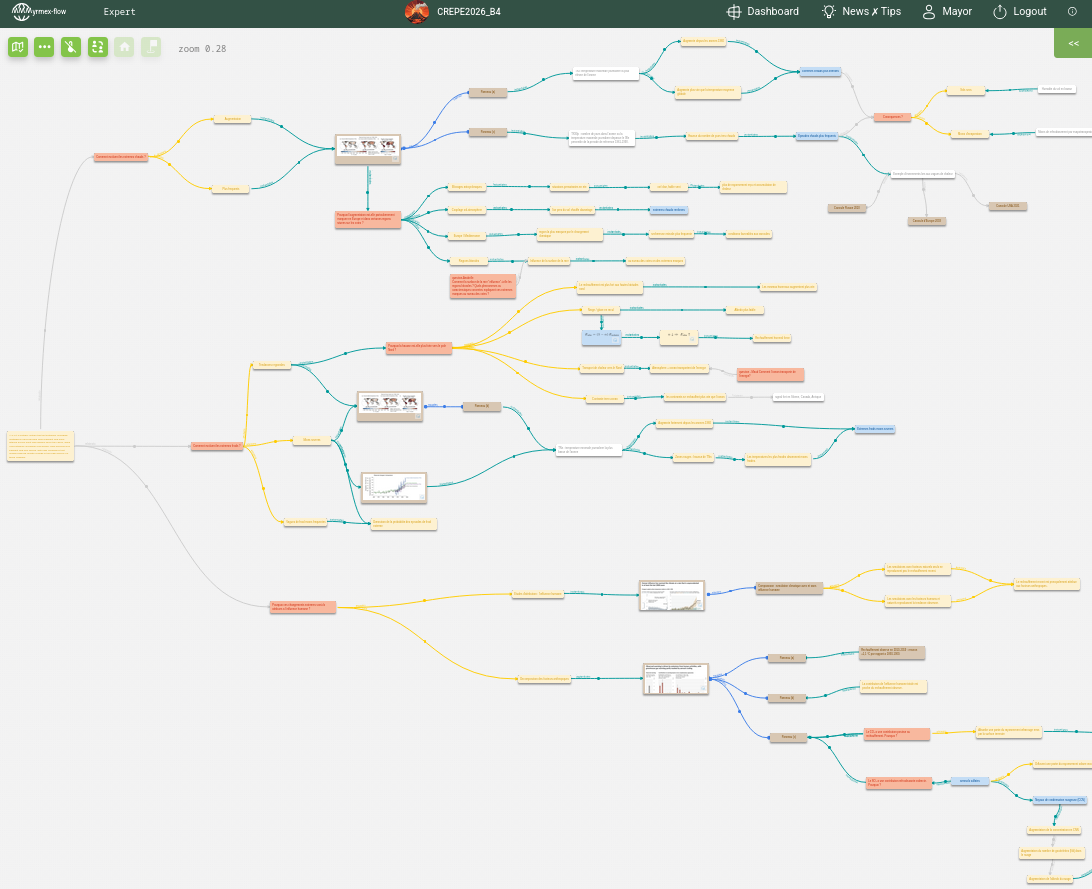}\\
        \includegraphics[trim={0 12mm 0 0},clip,width=0.492\textwidth]{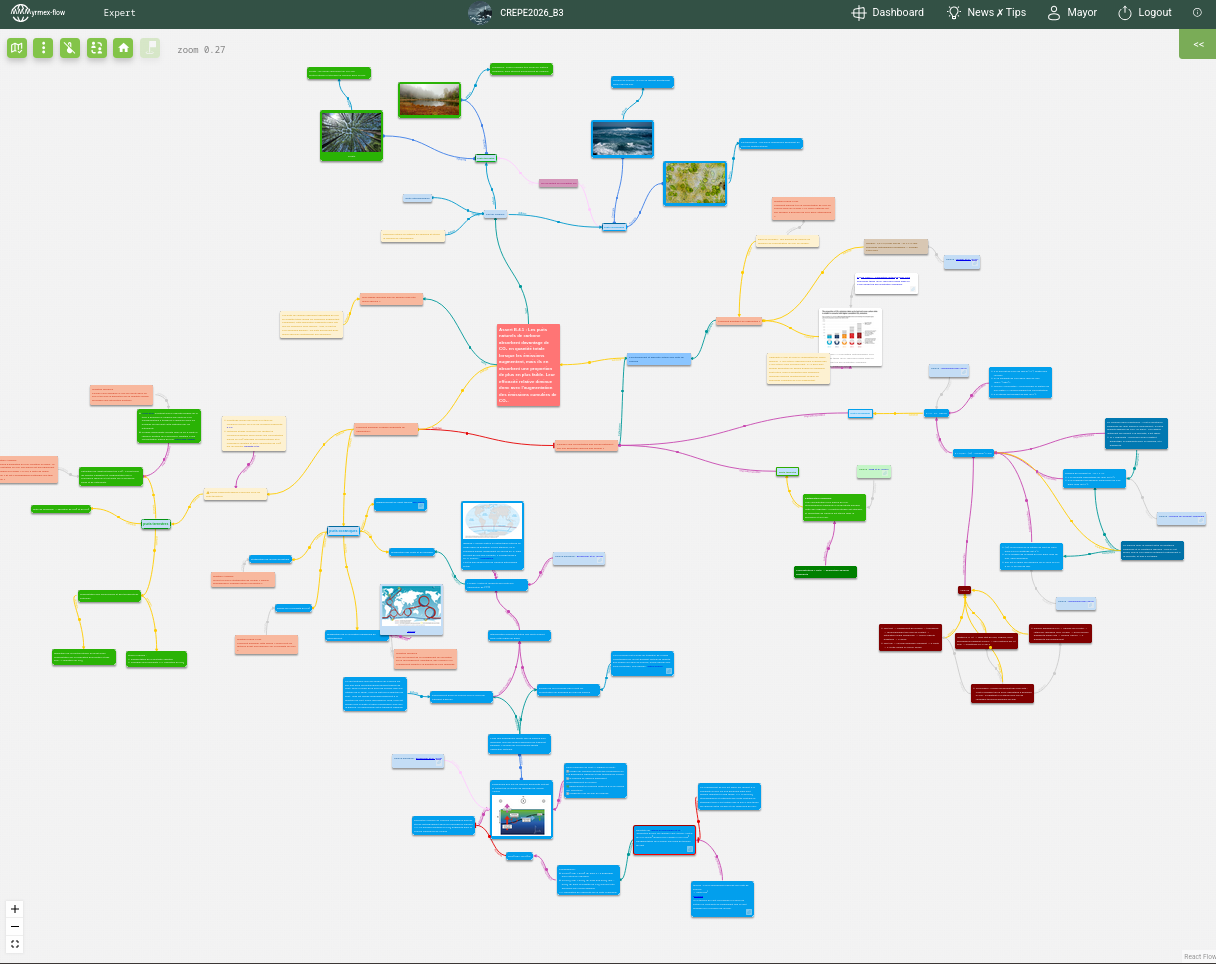} }\hspace{0pt}
    \parbox{0.49\textwidth}{
        \includegraphics[width=0.508\textwidth]{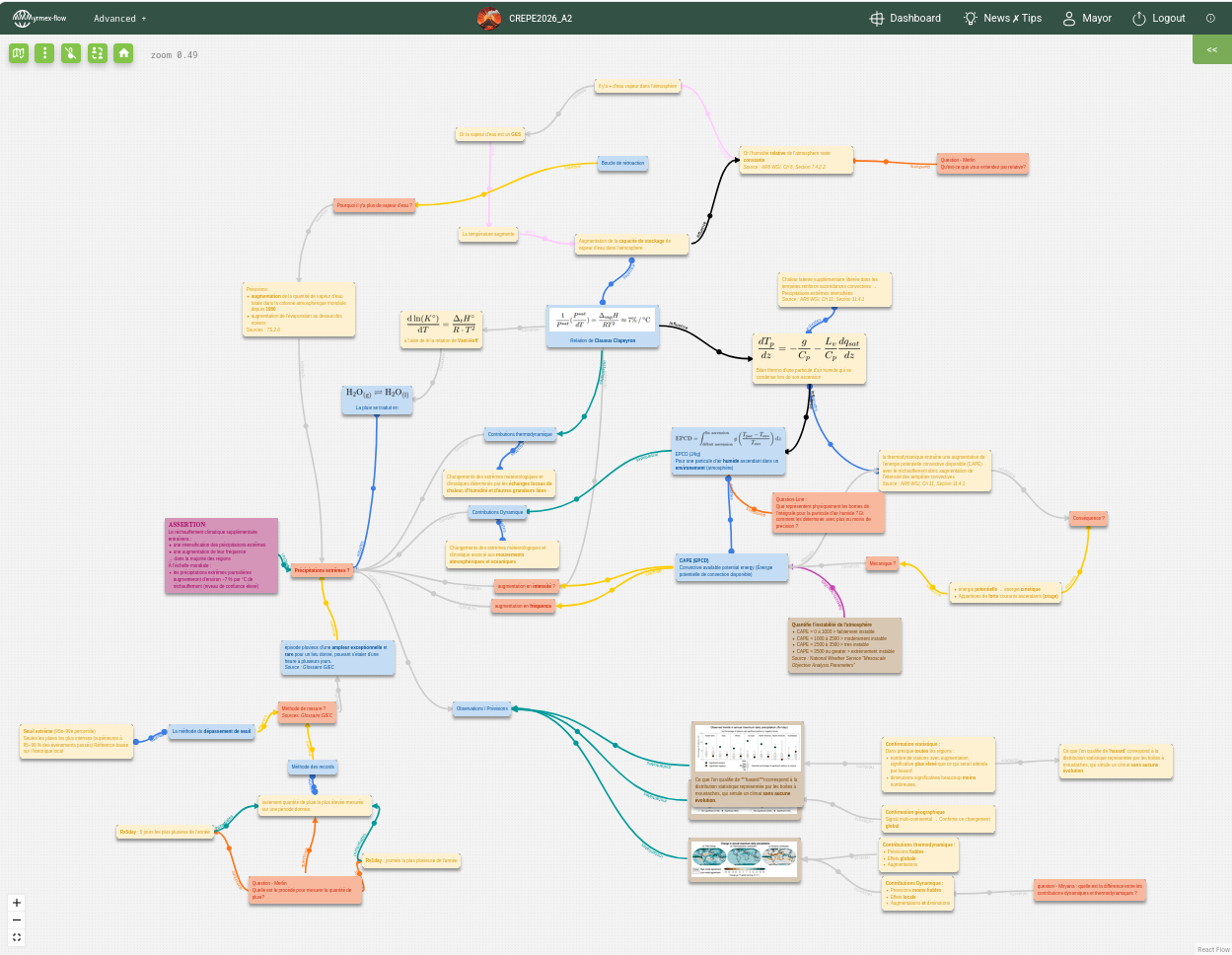}\\ 
        \includegraphics[trim={0 0 2mm 0},clip,width=0.508\textwidth]{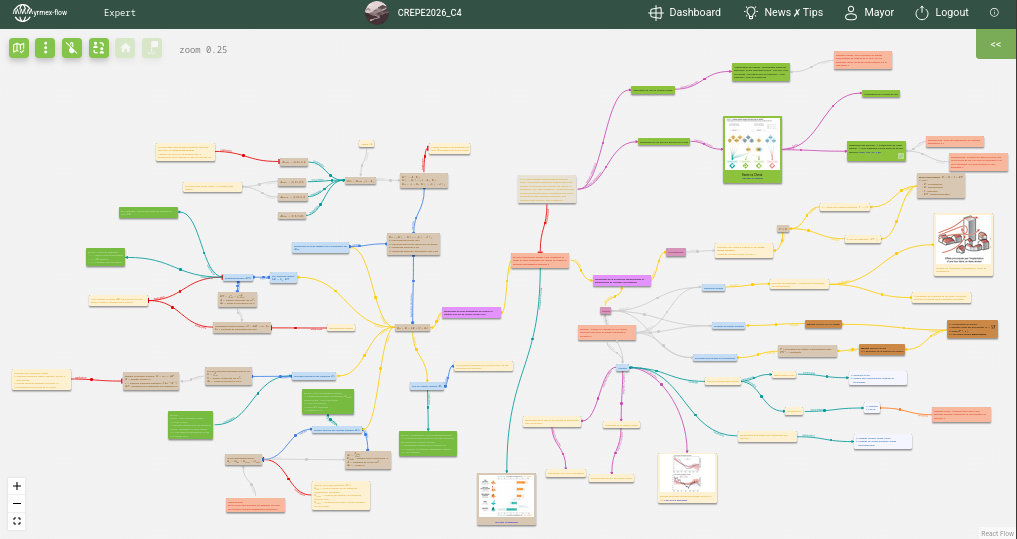} \\
    \includegraphics[width=0.508\textwidth]{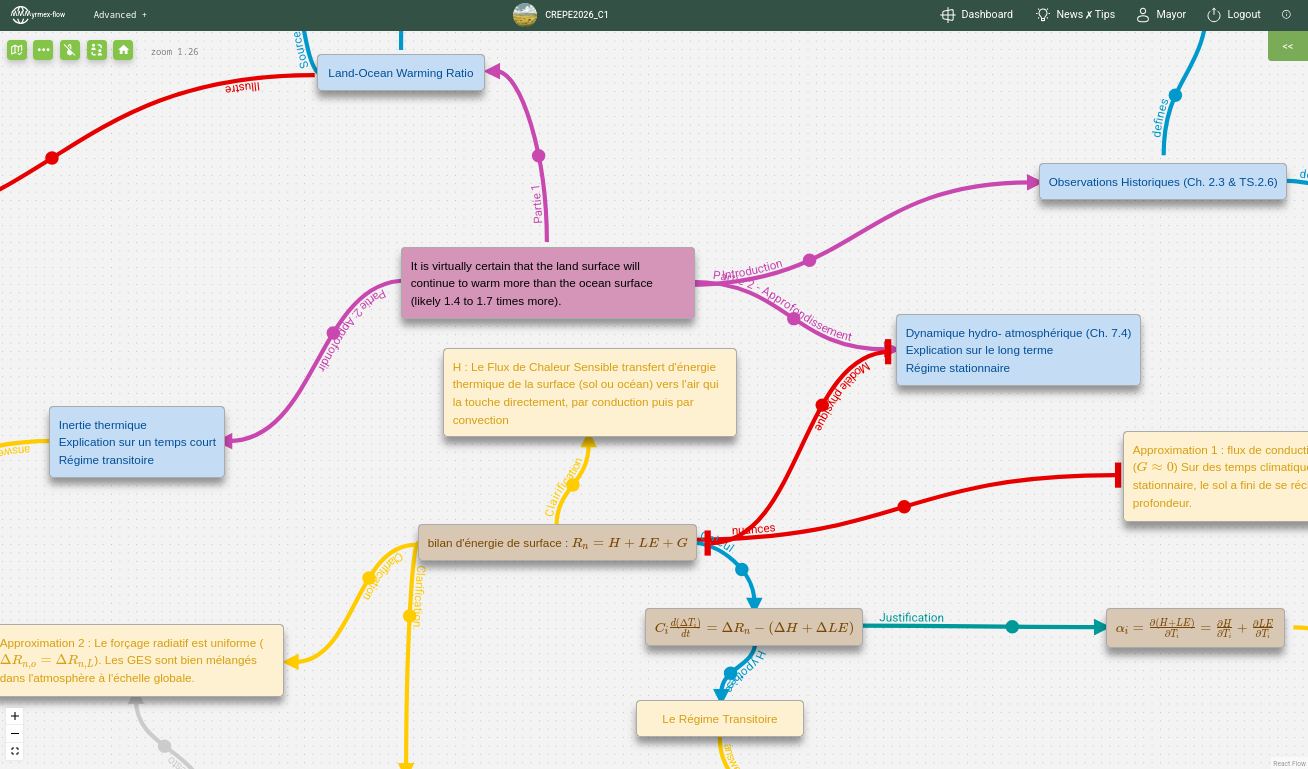}}
     
    \caption{A selection of CREPE 2026 student group workspaces illustrating the diversity of documentation practices that emerge within a single MMM-based exercise,  showing that expressive visual organisation
        and  MMM typing can coexist.   Groups vary in their use of spatial layout, colour coding, and      typing to structure the information   -- all valid choices  within MMM'S  flexible contribution framework. Workspaces at the top used   the default application styling, without any personalisation.  {\it Bottom right:}
        {Detail of a student workspace centred on the primary IPCC claim, with    edges radiating outward    irrespective of the normative directional semantics of typed edges (\S\ref{mmmtypes}).}
    }
    \label{fig:crepe}
\end{figure}

\paragraph{Idiosyncrasies.}
The per-workspace data also illustrates the flexibility the data model explicitly permits. Fig.~\ref{fig:crepe} illustrates that most authors make do with the default per-type Myrmex styling of contributions. However, Group D4 (CREPE 2026) exemplifies another end of the spectrum: 63\% of their edges are of fallback type, while their vertex type distribution is comparable to other groups. Inspection of their workspace (Fig.~\ref{fig:D4}) reveals a graph whose epistemic structure is conveyed primarily through visual organisation and custom styling -- colour, position, and layout -- rather than through MMM edge type conventions, producing a workspace that is visually legible and internally coherent while making limited use of the typing vocabulary.

Another tendency observed in some student graphs, concerns the orientation of edges. They sometimes follow the flow of information as students encountered the information during their research, and the succession of their reasoning steps -- typically radiating outward from the central IPCC claim (cf~Fig.\ref{fig:crepe} bottom right) -- rather than according to the directional semantics of typed edges (cf \S~\ref{mmmtypes}). For instance, a \texttt{Questions} edge, which should be oriented from the questioning contribution toward the contribution it questions, was sometimes reversed (cf~Fig.\ref{fig:B1}).

Some groups relied on edge labels and colour coding to convey relational meaning rather than on the normative type and orientation of edges. In the most problematic cases, poor typing carries a risk of conveying misleading epistemic information and misrepresenting the author's reasoning.
However, these idiosyncratic usage patterns are entirely consistent with MMM's design intent. In line with the \dimname{Human Primacy}, \dimname{Expression Intent}, and \dimname{Immediate \& Local Value} dimensions, the data model does not impose correctness on end users (cf~\S\ref{norms}): these graphs fully served their local purpose as the design priority is resolutely on \dimname{Immediate \& Local Value}. The trade-off is one of scope and scale rather than local quality. Precise typing and correct edge orientation are what enable cross-workspace querying, external readability, and future reuse (\dimname{interoperability}).
We see this as primarily a user experience challenge: nudging users gently toward respecting MMM conventions  
so that locally produced data is more readily reusable at wider scale, without making such conventions a barrier to participation.

\paragraph{Epistemically specific edge types.}
Beyond the fallback types, the distribution of specific edge types confirms that users
across all contexts actively structure their contributions rather than merely accumulate
disconnected vertices. 
\texttt{Pertains} 
accounts for 27.4\% of all edges globally and for 59.1\% of Researcher~1's edges, coinciding with the intended use of this edge type as default epistemically meaningful edge type.
\texttt{Instantiates},
\texttt{Answers}, \texttt{Questions}, \texttt{Challenges}, and \texttt{Substantiates} together
account for a further 30.1\% of edges globally, and are present in meaningful proportions
across all usage contexts.
The CREPE 2026 cohort in particular shows strong adoption of \texttt{Instantiates} (21.8\%
of edges, up from 7.0\% in 2025) and \texttt{Answers} (15.6\% up from 13.5\%) edge types. Compared to 2025,  usage of \texttt{Equates}, \texttt{\texttt{Challenges}}, and \texttt{Substantiates}  also increased.  
The second cohort exploited a broader range of the model's expressive capabilities. 

\subsection{Limitations}
The deployment evidence reported in this section is preliminary and subject to several limitations.  No systematic comparison was made with other tools.  Although the author's involvement was episodic, it cannot be entirely ruled out as a facilitating factor in some contexts. The deployment is small in scale.
The usage contexts reported are limited to research documentation and engineering education. Whether MMM's design is equally suitable for other domains -- such as journalism, law, or policy -- remains to be investigated. 
All data was produced within a single deployment of a single implementation: structural interoperability across independently governed deployments remains to be demonstrated. 

 \afterpage{
    \newgeometry{left=2.8cm,right=2.8cm} 

\begin{table}[htbp]
\centering
\small
\begin{tabular}{|>{\columncolor{gray!20}\bfseries\arraybackslash\vspace{2pt}}m{1.63in}|
     >{\centering}m{0.31in}| %
    m{10cm}|}
\hline
\textbf{Dimension} & {\tiny\textbf{Status}} & \textbf{Evidence or basis} \\
\hline

\multicolumn{3}{|l|}{\cellcolor{gray!20}\statusdem \textit{Proven (mostly) by deployment data in practice}} \\
\hline

Expression Intent 
& \cellcolor{provencolor}\statusdem
& Examples of contributions
expressing individual doubt, challenge, and situated interpretation (cf\ \S\ref{sec:pilot})
\\\hline

Redundancy-Friendly 
& \cellcolor{provencolor}\statusdem
& Overlapping and near-identical contributions in different graphs  (cf \S\ref{sec:pilot})
 \\\hline

Homogeneous Space 
& \cellcolor{provencolor}\statusdem
& Comments  structurally indistinguishable from primary content (cf \S\ref{observations}) \\\hline

Immediate \& Local 
& \cellcolor{provencolor}\statusdem
&  8 substantial graphs produced by solo users, including the biggest graph of size 905 (cf \S\ref{pastor})\\\hline

Persistent Production 
& \cellcolor{provencolor}\statusdem\statusdes
& Durability via identifier and field immutability (cf \S\ref{norms}). Graph fed a peer-reviewed publication (cf \S\ref{pastor}) +  2025 contributions still accessible and referenceable in 2026
 \\\hline

Adoption \& Accessibility 
& \cellcolor{provencolor}\statusdem
& Data model hidden behind graph UI in reference implementation (cf \S\ref{sec:pilot}). 350 landmarks created within 2 hours by 1st-year students without tutorial (cf \S\ref{crepe})\\\hline

\multicolumn{3}{|l|}{\cellcolor{gray!20}\statuspar \textit{Partially demonstrated}} \\\hline

Continual Improvement 
& \cellcolor{partialcolor}\statuspar
& MMM graph iteratively worked into a peer-reviewed publication; instructor and cross-group questions enrich graphs by surfacing gaps and inviting precision (cf \S\ref{sec:pilot}); \texttt{Pit}-based censureless mechanism support quality signalling (cf\ \S\ref{norms});  no graph-level quality metric yet defined to formally characterise improvement over time
\\\hline

Interoperability 
& \cellcolor{partialcolor}\statuspar
& Demonstrated across heterogeneous application paradigms via prototypes (cf Fig.~\ref{fig:interop-apps}, \S\ref{refapp}), cross-deployment interoperability pending open-source release and community uptake\\\hline

\multicolumn{3}{|l|}{\cellcolor{gray!20}\statusdes \textit{Demonstrated (mostly) by design}} \\\hline

{Universal Scope} 
& \cellcolor{designcolor}\statusdes\statusdem
& Free-text labels and attachment marks give MMM contributions the expressive range of documents with no restriction on topic, epistemic form, or formalisation degree (cf\ \S\ref{contributions}); 
confirmed in practice  across  mathematics, signal processing, systems biology, climatology, and mechanics,  spanning KaTeX formulae, natural language, and bibliographic references, by researchers, students, and teachers (cf\ \S\ref{sec:pilot})
\\\hline

Human Primacy 
& \cellcolor{designcolor}\statusdes\statusdem
& Natural language label field (cf S\ref{contributions}) + default/optional and forgiving typing (cf \S\ref{mmmtypes}, \S\ref{norms}, \S\ref{crepe}, e.g. Fig.~\ref{fig:D4}) 
  \\\hline

Strong Common Rules 
& \cellcolor{designcolor}\statusdes
& cf Section \ref{mmm} and in particular \S\ref{norms} \\\hline

Disagreement Express.
& \cellcolor{designcolor}\statusdes
& 
Native mandatory representation of logical absurdty (\texttt{Pit}) +  \texttt{Challenges} as a first-class edge type to highlight contradiction  
\\\hline

No Convergence Intent 
& \cellcolor{designcolor}\statusdes
& \dimname{Disagreement  Expression} 
 + field immutability ensures contradictions coexist permanently (cf \S\ref{contributions}, \S\ref{mmmtypes}, \S\ref{norms}).
 \\\hline

No Global Consistency 
& \cellcolor{designcolor}\statusdes
& Contribution validity is purely syntactic (cf \S\ref{contributions}); no consistency constraints are declared and no enforcement mechanism exists (cf \S\ref{norms}).
 \\\hline

Min. Ontol. Commitment 
& \cellcolor{designcolor}\statusdes
& Small fixed epistemic type vocabulary  (cf \S\ref{mmmtypes}), and  schema-free text labels    (cf \S\ref{contributions})  \\\hline

Normative Data Model 
& \cellcolor{designcolor}\statusdes
& cf Section \ref{mmm} \\\hline

Formal Typing 
& \cellcolor{designcolor}\statusdes
&
cf \S\ref{mmmtypes} 
\\\hline

Post-Document 
& \cellcolor{designcolor}\statusdes
&  UUIDs and the structure described in  \S\ref{contributions} 
  \\\hline

1st-Class Relationships 
& \cellcolor{designcolor}\statusdes
& Edges are full-fledged landmarks, cf \S\ref{contributions}  \\\hline

Knowledge System 
& \cellcolor{designcolor}\statusdes\statusdem
& \dimname{Post-document organisation} + \dimname{formal typing}  (cf.\ \S\ref{contributions}, \S\ref{mmmtypes}); confirmed in practice: MMM's epistemic types actively used  (cf\ \S\ref{sec:pilot})
\\\hline

Contextual Enrichment 
& \cellcolor{designcolor}\statusdes
& 1st-class edges (cf \S\ref{contributions}) + in practice: 53\% (resp. 34\%) of contributions are edges (resp. epistemically meaningful edges), cf \S\ref{sec:pilot} \\\hline

\multicolumn{3}{|l|}{\cellcolor{gray!20}\statusfut \textit{Dependent on external conditions, not yet demonstrable, future work}} \\
\hline

Enclosure Resistance 
& \cellcolor{futurecolor}\statusfut
&  No institutional custodianship arrangement yet in place (cf Section \ref{futurework})
\\\hline

Write Decentralisability 
& \cellcolor{futurecolor}\statusfut
& CRDT merge  not yet formalised, and
current deployment is centralised\\\hline

Wide-Scale Collab.
& \cellcolor{futurecolor}\statusfut
&  Maximum number of collaborators observed so far on the same graph is 9 (cf \S\ref{sec:pilot}) \\\hline

Emergent Coll. Benefits
& \cellcolor{futurecolor}\statusfut
& Spontaneous adoption at scale not yet reached  
\\\hline

\end{tabular}
\smallskip
\caption{Summary of empirical and 
architectural evidence for each dimension defined in Section~\ref{dimensions}. }
\label{tab:dimensions-evidence}
\end{table}
\clearpage
\restoregeometry
}

\section{Conclusion}

Table~\ref{taxonomytable} recalled
that the combinations of properties exhibited by existing information systems leave room for intermediate positions in the design space between widely adopted permissive systems and powerful but demanding formal ones.
Section~\ref{mmm} proposed the MMM data model  as a basis for one such position: slightly more structured than the WWW, preserving the full expressivity of natural language and  the possibility of accommodating knowledge in the making, yet with just enough structure to support meaningful reuse.

MMM approaches  \dimname{write decentralisability} and \dimname{interoperability}  by starting from a different architectural premise than the Semantic Web: since semantics are never purely formal and systems whose interoperability depends on shared semantics cannot be genuinely decentralised, MMM proposes that independent deployments agree on how knowledge is packaged and identified, but not on what it means. By abandoning semantic convergence requirements, MMM avoids the governance burden of maintaining shared semantics while preserving interoperability at the structural level.

MMM is intended as an infrastructural substrate rather than an end-user application.  Its relevance is tied to
the user experience provided by applications built on top of the data model  -- their ease of use, and the \dimname{immediate value} they represent for users who may already be locked into other applications.
MMM's flexibility with respect to free text  is intended to ensure  that MMM-based tools can provide immediate standalone value, similarly to PKM systems, before any network effects emerge.
In principle, any knowledge creation software could be reimplemented on top of MMM
without loss of usability, as long as their content can be represented as free text
or  files -- whether doing so would be worthwhile depends on the value the application provides to its users.
Research prototypes have been developed toward demonstrating this.
Early testing with scientists and students at small scale suggests that the approach is promising.  

Whether MMM can be meaningfully taken up depends on whether suitable social and institutional conditions can be established around it. 
Network effects arise from shared practices and  committed communities rather than formalism alone.  Historical examples such as OpenStreetMap, Wikipedia, and early Linux development illustrate the importance of mapping parties, writing groups, meetups, and other founding communities and events in bootstrapping content, conventions and contributor onboarding \cite{Hristova2013MappingParties,olson2000distance,raymond1999cathedral,wikipediaGoodFaith,osm_mapping_parties}.

As with many infrastructural technologies, MMM can ultimately only be assessed through its situated deployment and the practices that emerge around it.

\section{Limitations}

MMM is a design proposal at this stage. While the core data model is specified and a reference implementation exists, 
certain components --  in particular the set of unidirectional edge types and pen types, and their semantics --  remain provisional and subject to stabilisation, which itself depends on sufficient uptake to inform final definitions. 

\dimname{write decentralisability} of MMM remains partially theoretical until  CRDT merge mechanism   is formalised.
\dimname{Interoperability} beyond a single deployment also remains to be demonstrated through independent developer uptake.

The deployment evidence is preliminary: small in scale, limited to academic use cases, and produced within a single implementation. 

Several dimensions figuring in Table \ref{taxonomytable} are satisfied by MMM by design of the data model or can be verified at small scale with the reference application (e.g. \dimname{immediate individual value},  cf Table~\ref{tab:dimensions-evidence}). Others  -- \dimname{Wide-Scale Collaboration} and \dimname{adoption},   \dimname{enclosure resistance}, and \dimname{emergent collective benefits}   --  depend at least in part on external conditions, including  community formation, 
tool ecosystem development, 
network effects, 
and institutional arrangements, 
that the current work  and the small pilot deployment
cannot address. 
And in particular, until anti-enclosure mechanisms are devised and implemented, the system offers no guarantees of long-term distributed custodianship.

\section{Future work and Vision}
\label{futurework}

Future work spans several directions.

\subsection{Finalisation}
First, the MMM data model itself requires finalisation of CRDT merge semantics for deterministic decentralised synchronisation, stabilisation of underspecified pen and edge types, definition of governance mechanisms for evolving the MMM specification, and further development of tools, interfaces, and further pilot deployments in research and education contexts.
Concurrently, we are actively refactoring and cleaning up the reference codebase, with the intention of releasing a substantial subset of it as open-source.

\begin{figure}
    \includegraphics[width=1\textwidth]{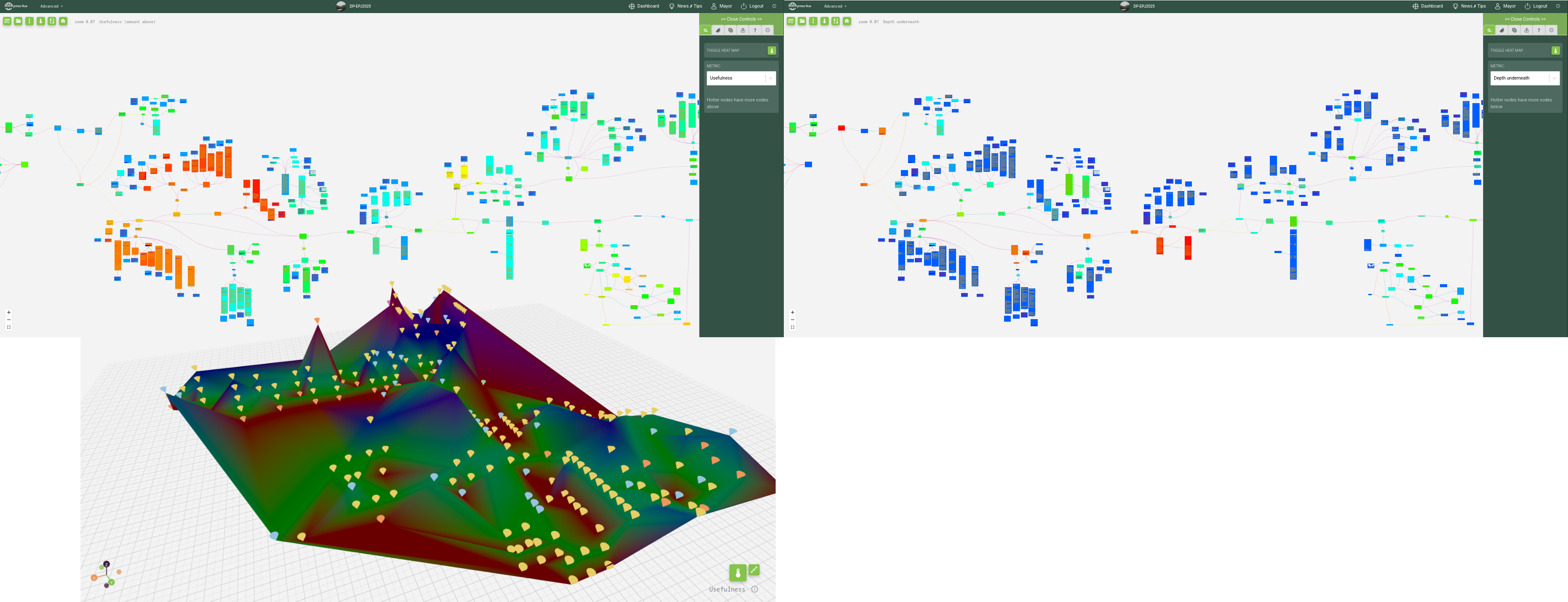}
    \caption{MMM formatted data can be qualified using formal metrics and visualised accordingly. Our reference app Myrmex demonstrates this through a small  set of naive metrics including so-called "usefulness" and "depth" (number of contributions along an outgoing/incoming path). Courtesy of D. Pastor. \medskip
    }
    \label{fig:metrics}
\end{figure}

\subsection{Metrics}
\label{metrics}

An additional line of work concerns MMM-based metrics to qualify knowledge.
MMM's formal structure allows  leveraging  graph-theoretic properties (e.g., path length, vertex degree)  and MMM contribution types to characterise information qualities relative to single contributions or to areas of an MMM graph (e.g., usefulness, interdisciplinarity,  nuance,  how well contextualised, challenged or corroborated, or the bias of a contribution in terms of its position relative to some specific other contribution, or its overuse of specific epistemic patterns). Such metrics could be used for multiple purposes, including   filtering information deemed unreliable or irrelevant according to finely specified quality criteria,  and supporting diverse representations of the epistemic landscape, making qualitative differences in information more intuitively legible,
as demonstrated by a 3D prototype in   \href{https://myrmex.app}{the reference implementation}  (cf Fig.~\ref{fig:metrics}). Domain and purpose oriented metrics need to be conceived, tested and represented visually. No single generic metric should be expected to reliably qualify information.

\subsection{Insight from Graph Structure}
\label{futureresearch}
A further application of MMM-based metrics   and more generally of graph-structural analysis may be to derive global insights, such as identifying under-developed lines of inquiry,  or tracking sentiments and their evolution across a body of documented work.

Assuming a definition of "compatibility" between contributions in terms of  structural criteria  (e.g. reachability without traversing {\tt Challenges} edges), then starting from a landmark representing a specific assumption (e.g. P=NP), all 
contributions compatible with that assumption could be identified and all others temporarily  hidden (at the application level).
MMM metrics could then be computed over the remaining
visible subgraph to characterise its structure, and identify underdeveloped regions (e.g. areas sparse in {\tt Question} vertices).
Repeating the process under an alternative assumption (e.g. P$\neq$NP) may allow the resulting subgraphs and their metrics to be compared. Asymmetries between the two perspectives could reveal gaps in  corresponding lines of inquiry,
expose questions that naturally arise under one assumption but not the other, and suggest contingently overlooked but hopefully  tractable avenues of investigation.

MMM graphs could also hypothetically support analogy-making across disciplines and lines of inquiry by enabling structural comparison between independently documented research programmes, with the aim of revealing useful
higher-level correspondences. The author's  background includes work on Boolean automata networks, investigating recurring structural motifs (e.g. negative and positive feedback loops, cycle intersections, and branching structures), interpreted in functional terms (e.g. generation of genuinely new information or of distinct time steps, memory, combination or destruction of information)
and studied as candidate building blocks for characterising the complexity of transitory and asymptotic network behaviours, in an implicit parallel with the role of primitive recursive functions in computability theory. In interdisciplinary settings, such  methodological analogies may remain implicit until enough formal results accumulate to make them legible to outsiders with different scientific references and culture. An navigable MMM graph documenting not only individual results but also the surrounding constellation of questions, intuitions, hypotheses, and dependencies to existing work could make an emerging research approach legible and   disputable across disciplinary traditions, before it admits a rigorous formal description and produces enough results to be appreciated as an intentional method. Comparing such a graph with an independent one charting the structured knowledge of an established field (e.g. computability theory) could then  expose similarities in the organisation of reasoning (or the tenuousness of suspected similarities),  that would otherwise be difficult to articulate.

\subsection{Epistemic Time Travel}

Since MMM contributions are append-only, each contribution can be treated as an event in time,
enabling frontends that reconstruct, visualise, and replay the evolution of epistemic landscapes over time, navigating
the sequence of knowledge acts back and forth, rewinding, fast-forwarding, or
comparing different temporal snapshots of the same graph.
Combined with MMM based metrics, and with temporal metrics (e.g. growth rates of qualified patterns or subgraphs) this opens up to a form of  perspective on the development and dynamics of knowledge:
visualising how understanding evolved, identifying areas of sustained focus, and surfacing historically, but contingently neglected regions.
Exploring these temporal analytics and visualisation layers constitutes a  direction for future work.

\subsection{Epistemic Versioning}

MMM's generally append-only philosophy (outside of small, locally managed settings where contributions are editable and consequences of edits can be handled manually)  means that correcting, updating or replacing a contribution can usually not be done by in-place mutation. Corrections must be expressed by creating a new contribution linked to the obsoleted one.  But the obsoleted one may have accumulated edges from other contributions that pertain to it, answer it, instantiate it, {\it etc}. If a {\tt Narrative} vertex reading  "{\it the trousers are red}" must be superseded by one reading  "{\it the trousers are blue}",  some adjacent contributions   may transfer unproblematically to the new version -- e.g. a {\tt Question} asking "{\it where were the trousers bought?}"  -- while others may not -- e.g. an {\tt Existence} vertex labelled "{\it red}", linked via a {\it Pertains} edge to the original {\tt Narrative}. Edges that can  safely be duplicated to the new version must be distinguished from those that 
must instead  be submitted for manual review, in a way analogous to conflict markers in version control systems such as Git.
Local structural patterns, MMM types, and label comparison 
may inform the distinction.
Non-transferability may also propagate along specific structural patterns to  contributions depending on an already identified non-transferable ones  -- e.g. a {\tt Question}  reading "{\it are the trousers uniformly red?}" linked both to the original {\tt Narrative} and to the {\tt Existence} vertex labelled "{\it red}".
Future work will investigate 
how edges incident to a superseded contribution and its surrounding subgraph should be treated when a new version is created,  how    to surface a broader set of likely-affected  candidates from a small number of initial local judgements, and how to present these candidates and any resolution suggestions for efficient human review.
Such \dimname{post-document} versioning would allow contributors to localise the consequences of revisions in their ongoing work, directing attention only to the parts of the graph genuinely affected by a change. 

\subsection{Persistence and Archival}

Research is still  needed on the persistence and long-term stewardship of MMM data.
This includes investigating database designs and storage architectures that are best suited to MMM-formatted data, and its access patterns in practice.

Because of the \dimname{enclosure resistance} dimension that our scientific research oriented solution aims at (cf \S\ref{whiteboard}), work is also needed on infrastructures for the long-term preservation and access of MMM data, especially public (e.g. scientific)  MMM data. A two-tier storage architecture could
combine (i)  "cold" slow-access archival storage, designed for infrequent access and long-term preservation (inspired by the Internet Archive's Wayback Machine, but distributed and structured), storing snapshots of MMM data (e.g. rarely accessed historical material), with (ii) a "warm", faster access layer (local-first storage and P2P sync when online).
This entails conceiving mechanisms  for synchronisation, migration, and unified querying across warm and cold storage layers, enabling both efficient day-to-day access and durable long-term preservation.

\subsection{Surfacing Conventions through Graph Analysis}

MMM has no \dimname{Semantic Convergence Intent}, so it does not face the specific write  decentralisability challenges discussed in section \ref{framework}.
But it could serve as infrastructure for observing when convergence is emerging organically, without imposing it. A rich discussion graph could reveal, through graph-theoretic properties, which questions have been discussed extensively enough that a stable pattern has emerged (e.g. high agreement density, low rates of new objections), or narrow the scope of the remaining uncertainty. Without eliminating the need for human judgment, this could reduce the workload and influence of central standard-setting bodies by ensuring that final conventions are established in light of a broad and transparent body of evidence and reasoning.

MMM may provide a foundation for researching decentralised, domain-independent merge mechanisms. Unlike purely syntactic merge mechanisms such as Git's, which operate on text structure alone, MMM-based merge could be informed by the epistemic structure of contributions.  Decisions could then be guided by measurable indicators of consensus, stability, and justification derived from the graph itself (MMM based metrics), enabling convergence processes that remain expert-validated,  flexible and auditable.
\medskip

Future work involves interfacing MMM data with more formal knowledge systems, including  progressive formalisation pathways through which stable, consensual subsets of MMM content can be lifted into more specialised formalisms such as RDF/OWL, 
allowing MMM to act as connective tissue between  systems optimised for specific formal representation and reuse purposes, and the vast middle ground of general human knowledge they do not currently cover.

\subsection{Exploration and Discovery}

Conversely,  external semantic structures  -- biological taxonomies, library classifications, SKOS concept schemes, domain ontologies and knowledge graphs -- may be selectively imported   into MMM and used as “skeletons”    for perspective-dependent  routing and exploration of distributed knowledge. This would offer an alternative to centralised search and AI assistants (Google, ChatGPT) that each enforce their own single proprietary, opaque relevance logic uniformly across users \cite{brereton2022nextgoogle,introna2000shaping,lindemann2025chatbots}. A user query could be mapped  by a local engine, to one or more vertices in a selected skeleton/perspective  (e.g., legal, biological, general purpose). Traversal of the MMM network could  be constrained to relevant subgraphs surrounding those vertices, rather than requiring a global crawl. Each skeleton vertex could be associated with one or several preferred  MMM deployments (e.g. a deployment maintained by an institution such as a local or partner university). Query propagation would be constrained to one or several of those specific deployments. Each receiving deployment would match the query against its own skeleton, identify relevant skeleton vertices and surrounding relevant MMM material, and, if necessary,  forward the query only to its own associated designated deployments documented as being relevant for material relative to the skeleton vertex in question.
While the MMM data model provides a foundation for such perspective-dependent, decentralised knowledge exploration,
the network-level protocols and formalisation pipelines required to instantiate it -- including 
skeleton import and alignment, query-mapping, LLM relaying  at the tips of skeleton branches to identify relevant areas 
 and inter-deployment query propagation  -- constitute future research and engineering work.  

Another related line of research work concerns publish-subscribe mechanisms, including  subscription scoping, allocation of responsibilities between publishers and subscribers,  and the design of access-control policies governing subscription eligibility (who may subscribe to what). Subscribers could register interest in specific topics or contribution patterns (e.g. as specified by previously mentioned metrics) with multiple  MMM deployments (e.g., universities, news media, local government) and receive notifications of material newly available on these deployments.

\subsection{Distributed House-keeping}

In a decentralised ecosystem, information maintenance mechanisms are needed, in particular to manage redundancy. MMM is designed to be \dimname{Redundancy friendly} but not all redundancy is constructive.  Shortening lengths of paths between related contributions may improve navigation and accessibility of each piece of knowledge.
The exact means by which equivalence and  transitive relations between MMM contributions are defined and detected is planned future work. Likewise, the question of which actors (public institutions, community curators, or search providers) take responsibility for crawling  and identifying equivalence across the network remains an open governance problem.

MMM search may  provide incentives for contributors to  appropriately connect their contributions using well typed edges reflecting what they add to the existing epistemic landscape. The more embedded a contribution is within the surrounding epistemic landscape, the more formal reasons MMM metric-based search engines may have to access it. Conversely, disconnected, poorly connected, or poorly typed contributions may suffer from reduced visibility.

As part of upcoming frontend development, {\t Pit}-based quality control will be implemented and tested in the reference application.  

\subsection{Complementarity between MMM and LLMs} 

The structured, typed nature of MMM graphs opens several complementary directions at the intersection of knowledge representation and large language models.

A first direction concerns MMM-LLM integration for documentation assistance. An integrated LLM prompting functionality was implemented in the reference prototype, allowing users to send individual or multiple contributions as prompts and collect responses as connected {\tt Narrative} contributions.
Our next step is to fine-tune a lightweight open-source model to generate and parse MMM-JSON for a range of tasks -- e.g. suggesting contribution types, converting unstructured documents into MMM graphs, and generating (ephemeral) output documents (slides, questionnaires, technical reports) from selected MMM subgraphs -- improving on the existing rule-based implementations that  convert selected MMM graph portions into LaTeX Beamer slide decks or glossaries.

A second direction concerns MMM as a structured persistence layer for LLM interactions more generally. Building on the existing prompting functionality, the goal is to use MMM systematically to store conversation threads, prompts, context, and responses as typed contributions, supporting navigation of prompt-response histories, precise context assembly, collaborative exploration,  branching,  replay and continuation of past interactions, which may be particularly useful for orchestration of AI-generated code as part of new programming workflows.

A third direction concerns MMM-informed prompting strategies and retrieval. Stored prompt-response pairs could serve as a cache, to resolve identical or similar queries from existing contributions rather than recompute them, and to avoid queries    that previously led to  unproductive reasoning paths and dead ends by consulting the surrounding graph structure. MMM-derived graph metrics (involving e.g. vertex type, centrality, distance to solved examples, edge-type distribution) could further guide prompt engineering without additional training: assembling relevant context and examples from graph proximity or structural similarity, generating prompt templates from MMM subgraphs, and adapting temperature (e.g. lower in highly technical or formal regions, higher in more abstract or exploratory ones, exploiting the orientation of unidirectional MMM edges from concrete specifics to abstract generalities, cf \S\ref{mmmtypes}).
Query routing could also exploit these signals to select appropriate models following the cost-efficiency principles of FrugalGPT \cite{chen2023frugalgptuselargelanguage}.
For example, a {\tt Question} located deep within a specialised graph with no nearby answers could be routed to a more expensive model, whereas one close to solved examples could be handled locally via a cheaper model or retrieval.
Furthermore, MMM graphs explicitly capture reasoning steps and inferential relations as vertices and edges,  making them structured, reusable, and accountable reasoning scaffolds for Graph-of-Thought-style prompting \cite{GoT2024,GoT2026}. 
And MMM-based graph traversal could potentially offer a more explicit and auditable alternative to vector-similarity retrieval for RAG, replacing similarity search over unstructured documents with structured traversal over typed MMM contributions.

Finally, MMM's filtering capabilities -- already implemented in the reference application (cf Fig~\ref{fig:filters}) -- provide a configurable  mechanism for curating   training subsets
across domains, expertise levels, and confidentiality constraints  (e.g. public, shared, private contributions). MMM contributions functioning as annotations on existing material, together with MMM-based metrics characterising a contribution's position in the broader network and epistemic usage patterns,
may serve as fine-grained structured supervision signals during fine-tuning.
This would enable models to learn epistemic distinctions such as challenged vs. corroborated claims, obsolete vs. current information,
and degrees of bias, and to calibrate outputs accordingly.

\subsection{Centralisation}

Institutional governance plays a central role. Systems that achieve durable commons (\dimname{enclosure resistance}) typically rely on some form of organisational structure and licensing frameworks to preserve accessibility and continuity of shared content.
In the case of MMM, decentralisability requires complementary institutional arrangements that ensure the resulting information space remains globally navigable -- rather than fragmenting into disconnected local datasets.  The aim is an \dimname{emergent collective benefit} analogous to the continuous coverage of OpenStreetMap: users should be able to traverse the contribution landscape  without encountering significant discontinuities.

A relevant historical pattern is that decentralised systems are notoriously prone to re-centralise in practice in certain functions,
not necessarily through hostile enclosure but through the inherent pull of coordination costs, infrastructure maintenance, and the emergence of dominant implementations \cite{Schneider2019Decentralization}. A pragmatic approach is therefore to anticipate  it by design: to decide explicitly and in advance    which forms of partial centralisation are acceptable or even  necessary  because they serve the commons rather than enclose it. Designing for acceptable centralisation upfront is preferable to discovering imposed centralisation after the fact.

For long-term commons  durability,  we envisage public institutions helping maintain and curate MMM-based knowledge as a public good,  taking responsibility for preserving what they classify as part of the public good,  enabling the  collective benefit to emerge from distributed contributions rather than leaving integration entirely to private coordination. But in line with  Ostrom's defiance of the "sterile dichotomy" between private and public governance, arguing for mixed institutional designs \cite{ostrom1990governing}, any MMM deployment -- private or public -- may persist and  expose any public MMM contributions. Each institution would apply its own criteria, resulting in overlapping but non-identical sets of preserved material. This pluralistic federation of diverse deployments with diverse priorities and expertise would avoid a single authoritative curator while maintaining interoperable knowledge bases.

\subsubsection*{Acknowledgements.~} 

~I thank Steve Coast, founder of OpenStreetMap, for his careful reading of a draft of this article. His generous list of detailed comments and observations highlighted a number of additional perspectives and points for consideration. 
I remain solely  responsible for all claims and any remaining errors.
I am  grateful to Dominique Pastor and the CapECL students and instructors for their engagement with early versions of the MMM data model through early research prototypes.
Thanks to Telmo Menezes for proposing the research direction of exploring MMM graphs by "turning the light on and off" on regions compatible with different assumptions, mentioned in \S\ref{futureresearch}. I also thank numerous colleagues and collaborators    with whom I have discussed these ideas over the years, particularly in  Germany and France. Their questions, critiques, and perspectives have collectively shaped and refined the framework presented here. While it is not possible to name them all individually, their contributions are deeply embedded in the   work presented here.
Finally, I thank  Adrien Richard for having shared frustrations at the whiteboard that led to our initial formulation of the core ideas underlying MMM.

\bibliography{bib}

@article{dp2026,
	author = {{Pastor, Dominique} and {Fernandez, Jonas} and {Thomas-Vaslin, V\'eronique}},
	title = {The Generic Sensory Automaton (GENSA) for modeling affinity-based dynamics and interactions in biology},
	DOI= "10.1051/epjconf/202636601009",
	OPTurl= "https://doi.org/10.1051/epjconf/202636601009",
	journal = {EPJ Web Conf.},
	year = 2026,
	volume = 366,
	pages = "01009",
}

@article{bush1945we,
  title={As we may think},
  author={Bush, Vannevar},
  journal={The atlantic monthly},
  volume={176},
  number={1},
  pages={101--108},
  year={1945},
  OPTpublisher={[S. l.]}
}

@inbook{RN2,
   author = {IPCC},
   title = {Summary for Policymakers},
   booktitle = {Climate Change 2021: The Physical Science Basis. Contribution of Working Group I to the Sixth Assessment Report of the Intergovernmental Panel on Climate Change},
   OPTeditor = {Masson-Delmotte, V.  and Zhai, P. and Pirani, A. and Connors, S.L. and Péan, C. and Berger, S. and Caud, N. and Chen, Y. and Goldfarb, L. and Gomis, M.I. and Huang, M. and Leitzell, K. and Lonnoy, E. and Matthews, J.B.R. and Maycock, T.K. and Waterfield, T. and Yelekçi, O. and Yu, R. and Zhou, B.},
   publisher = {Cambridge University Press},
   OPTaddress = {Cambridge, United Kingdom and New York, NY, USA},
   pages = {3--32},
   OPTDOI = {10.1017/9781009157896.001},
   year = {2021},
   type = {Book Section}
}

@misc{huang2023generativeaidigitalcommons,
  title         = {Generative AI and the Digital Commons},
  author        = {Saffron Huang and Divya Siddarth},
  year          = {2023},
  eprint        = {2303.11074},
  archiveprefix = {arXiv},
  primaryclass  = {cs.CY},
  url           = {https://arxiv.org/abs/2303.11074}
}

@article{noroozian2025generative,
  title   = {Generative AI and the Future of the Digital Commons: Five Open Questions and Knowledge Gaps},
  author  = {Noroozian, Arman and Aldana, Lorena and Arisi, Marta and Asghari, Hadi and Avila, Renata and Bizzaro, Pietro Giovanni and Chandrasekhar, Ramya and Consonni, Cristian and De Angelis, Deborah and De Chiara, Francesca and others},
  journal = {arXiv preprint arXiv:2508.06470},
  year    = {2025}
}

@misc{sun2026aiimprovesanswersslows,
      title={When AI Improves Answers but Slows Knowledge Creation: Matching and Dynamic Knowledge Creation in Digital Public Goods}, 
      author={Keh-Kuan Sun},
      year={2026},
      eprint={2604.00468},
      archivePrefix={arXiv},
      primaryClass={econ.GN},
      url={https://arxiv.org/abs/2604.00468}, 
}

@article{del2024large,
  title     = {Large language models reduce public knowledge sharing on online Q\&A platforms},
  author    = {del Rio-Chanona, R Maria and Laurentsyeva, Nadzeya and Wachs, Johannes},
  journal   = {PNAS nexus},
  volume    = {3},
  number    = {9},
  pages     = {pgae400},
  year      = {2024},
  publisher = {Oxford University Press US}
}

@misc{bommasani2022opportunitiesrisksfoundationmodels,
      title={On the Opportunities and Risks of Foundation Models}, 
      year={2022},
      eprint={2108.07258},
      archivePrefix={arXiv},
      primaryClass={cs.LG},
      url={https://arxiv.org/abs/2108.07258}, 
      author={Rishi Bommasani and Drew A. Hudson and Ehsan Adeli and Russ Altman and Simran Arora and Sydney von Arx and Michael S. Bernstein and Jeannette Bohg and Antoine Bosselut and Emma Brunskill and Erik Brynjolfsson and Shyamal Buch and Dallas Card and Rodrigo Castellon and Niladri Chatterji and Annie Chen and Kathleen Creel and Jared Quincy Davis and Dora Demszky and Chris Donahue and Moussa Doumbouya and Esin Durmus and Stefano Ermon and John Etchemendy and Kawin Ethayarajh and Li Fei-Fei and Chelsea Finn and Trevor Gale and Lauren Gillespie and Karan Goel and Noah Goodman and Shelby Grossman and Neel Guha and Tatsunori Hashimoto and Peter Henderson and John Hewitt and Daniel E. Ho and Jenny Hong and Kyle Hsu and Jing Huang and Thomas Icard and Saahil Jain and Dan Jurafsky and Pratyusha Kalluri and Siddharth Karamcheti and Geoff Keeling and Fereshte Khani and Omar Khattab and Pang Wei Koh and Mark Krass and Ranjay Krishna and Rohith Kuditipudi and Ananya Kumar and Faisal Ladhak and Mina Lee and Tony Lee and Jure Leskovec and Isabelle Levent and Xiang Lisa Li and Xuechen Li and Tengyu Ma and Ali Malik and Christopher D. Manning and Suvir Mirchandani and Eric Mitchell and Zanele Munyikwa and Suraj Nair and Avanika Narayan and Deepak Narayanan and Ben Newman and Allen Nie and Juan Carlos Niebles and Hamed Nilforoshan and Julian Nyarko and Giray Ogut and Laurel Orr and Isabel Papadimitriou and Joon Sung Park and Chris Piech and Eva Portelance and Christopher Potts and Aditi Raghunathan and Rob Reich and Hongyu Ren and Frieda Rong and Yusuf Roohani and Camilo Ruiz and Jack Ryan and Christopher Ré and Dorsa Sadigh and Shiori Sagawa and Keshav Santhanam and Andy Shih and Krishnan Srinivasan and Alex Tamkin and Rohan Taori and Armin W. Thomas and Florian Tramèr and Rose E. Wang and William Wang and Bohan Wu and Jiajun Wu and Yuhuai Wu and Sang Michael Xie and Michihiro Yasunaga and Jiaxuan You and Matei Zaharia and Michael Zhang and Tianyi Zhang and Xikun Zhang and Yuhui Zhang and Lucia Zheng and Kaitlyn Zhou and Percy Liang}
}

@misc{krakauer2025emergence,
  author  = {Krakauer, David C. and Krakauer, John W. and Mitchell, Melanie},
  title   = {Large Language Models and Emergence: A Complex Systems Perspective},
  OPTjournal = {arXiv preprint arXiv:2506.11135},
  year    = {2025},
  doi     = {10.48550/arXiv.2506.11135}
}

@article{wei2022emergent,
  title={Emergent abilities of large language models},
  author={Wei, Jason and Tay, Yi and Bommasani, Rishi and Raffel, Colin and Zoph, Barret and Borgeaud, Sebastian and Yogatama, Dani and Bosma, Maarten and Zhou, Denny and Metzler, Donald and others},
  journal={arXiv preprint arXiv:2206.07682},
  year={2022}
}

@misc{underlay,
  author    = {Hillis, Danny  and  Klein, Samuel and  Rich, Travis},
  year      = {2018},
  title     = {\href{https://notes.knowledgefutures.org/pub/h67iji6d/release/1}{Underlay: A First Description}}}

@misc{originTrail,
  author    = {{Trace Labs -- OriginTrail Core developers}},
  year      = {2024},
  OPTpublisher={Trace Labs},
  OPTurl={https://origintrail.io/documents/Verifiable_Internet_for_Artificial_Intelligence_whitepaper_v3_pre_publication.pdf},
  title     = {\href{https://origintrail.io/documents/Verifiable_Internet_for_Artificial_Intelligence_whitepaper_v3_pre_publication.pdf}{Verifiable Internet for Artificial Intelligence: The Convergence of Crypto, Internet and AI}}
}

@inproceedings{bollacker2008freebase,
  title     = {Freebase: a collaboratively created graph database for structuring human knowledge},
  author    = {Bollacker, Kurt and Evans, Colin and Paritosh, Praveen and Sturge, Tim and Taylor, Jamie},
  booktitle = {Proceedings of the 2008 ACM SIGMOD international conference on Management of data},
  pages     = {1247--1250},
  series    = {SIGMOD '08},
  year      = {2008}
}

@article{wikidata,
  author  = {Vrande\v{c}i\'{c}, Denny and Kr\"{o}tzsch, Markus},
  title   = {Wikidata: a free collaborative knowledgebase},
  year    = {2014},
  volume  = {57},
  number  = {10},
  journal = {Communications of the ACM},
  pages   = {78--85}
}

@misc{jeuris2020socratreesexploringdesignargument,
      title={Socratrees: Exploring the Design of Argument Technology for Layman Users}, 
      author={Steven Jeuris},
      year={2020},
      eprint={1812.04478},
      archivePrefix={arXiv},
      primaryClass={cs.HC},
      url={https://arxiv.org/abs/1812.04478}, 
}

@misc{debatemap,
  title  = {Argdown Syntax Specification},
  author = {Wicklund, Stephen {\it et al.}},
  year   = {2013},
  url    = {https://debatemap.app}
}

@misc{story2015contradiction,
  author = {Henry Story},
  title = {contradiction and controversy - was: Is it necessary for the semantic web to be self contradictory?},
  howpublished = {W3C Public Mailing List Archive (public-philoweb)},
  year = {2015},
  month = may,
  day = {27},
  url = {https://lists.w3.org/Archives/Public/public-philoweb/2015May/0004.html},
  note = {Message-Id: <9F05F606-8D4F-4C63-86E3-F7765C83D8C1@bblfish.net>}
}

@misc{noualpersp,
  author       = {Mathilde Noual},
  title        = {Perspectives and Networks},
  OPTjournal      = {CoRR},
  OPTvolume       = {abs/1610.08765},
  year         = {2016},
  url          = {http://arxiv.org/abs/1610.08765},
  OPTeprinttype   = {arXiv},
  primaryclass  = {cs.OH},
  archiveprefix = {arXiv},
  eprint       = {1610.08765},
  OPTtimestamp    = {Mon, 13 Aug 2018 16:46:35 +0200},
}

@article{noualBAN,
  title     = {About non-monotony in Boolean automata networks},
  author    = {Noual, Mathilde and Regnault, Damien and Sen{\'e}, Sylvain},
  journal   = {Theoretical Computer Science},
  volume    = {504},
  pages     = {12--25},
  year      = {2013},
  publisher = {Elsevier}
}

@misc{brereton2022nextgoogle,
  author   = {Brereton, Dmitri},
  title    = {The Next {Google}},
  year     = {2022},
  optmonth = {April},
  url      = {https://dkb.blog/p/the-next-google},
  note     = {DKB Blog}
}

@article{introna2000shaping,
  title     = {Shaping the Web: Why the politics of search engines matters},
  author    = {Introna, Lucas D and Nissenbaum, Helen},
  journal   = {The information society},
  volume    = {16},
  number    = {3},
  pages     = {169--185},
  year      = {2000},
  publisher = {Taylor \& Francis}
}

@article{zhu2024llms,
  title={Llms for knowledge graph construction and reasoning: Recent capabilities and future opportunities},
  author={Zhu, Yuqi and Wang, Xiaohan and Chen, Jing and Qiao, Shuofei and Ou, Yixin and Yao, Yunzhi and Deng, Shumin and Chen, Huajun and Zhang, Ningyu},
  journal={World Wide Web},
  volume={27},
  number={5},
  pages={58},
  year={2024},
  publisher={Springer}
}

@article{lindemann2025chatbots,
  title     = {Chatbots, search engines, and the sealing of knowledges},
  author    = {Lindemann, Nora Freya},
  journal   = {AI \& Society},
  volume    = {40},
  number    = {6},
  pages     = {5063--5076},
  year      = {2025},
  publisher = {Springer}
}

@article{GoT2024,
  title   = {Graph of Thoughts: Solving Elaborate Problems with Large Language Models},
  volume  = {38},
  opturl  = {https://ojs.aaai.org/index.php/AAAI/article/view/29720},
  optdoi  = {10.1609/aaai.v38i16.29720},
  number  = {16},
  journal = {Proceedings of the AAAI Conference on Artificial Intelligence},
  author  = {Besta, Maciej and Blach, Nils and Kubicek, Ales and Gerstenberger, Robert and Podstawski, Michal and Gianinazzi, Lukas and Gajda, Joanna and Lehmann, Tomasz and Niewiadomski, Hubert and Nyczyk, Piotr and  Hoefler, Torsten },
  year    = {2024},
  pages   = {17682--17690}
}

@article{GoT2026,
  author  = {Besta, Maciej and Memedi, Florim and Zhang, Zhenyu and Gerstenberger, Robert and Piao, Guangyuan and Blach, Nils and Nyczyk, Piotr and Copik, Marcin and Kwaśniewski, Grzegorz and Müller, Jürgen and Gianinazzi, Lukas and Kubicek, Ales and Niewiadomski, Hubert and O'Mahony, Aidan and Mutlu, Onur and Hoefler, Torsten},
  journal = {IEEE Transactions on Pattern Analysis and Machine Intelligence},
  title   = {Demystifying Chains, Trees, and Graphs of Thoughts},
  year    = {2025},
  volume  = {47},
  number  = {12},
  pages   = {10967--10989},
  optdoi  = {10.1109/TPAMI.2025.3598182}
}

@misc{chen2023frugalgptuselargelanguage,
  title         = {FrugalGPT: How to Use Large Language Models While Reducing Cost and Improving Performance},
  author        = {Lingjiao Chen and Matei Zaharia and James Zou},
  year          = {2023},
  eprint        = {2305.05176},
  archiveprefix = {arXiv},
  primaryclass  = {cs.LG},
  url           = {https://arxiv.org/abs/2305.05176}
}

@misc{zhao2023survey,
  title         = {A survey of large language models},
  author        = {Zhao, W. X. and Zhou, K. and Li, J. and Tang, T. and Wang, X. and Hou, Y. and Min, Y. and Zhang, B. and Zhang, J. and Dong, Z. and Du, Y. and Yang, C. and Chen, Y. and Chen, Z. and Jiang, J. and Ren, R. and Li, Y. and Tang, X. and Liu, Z. and Liu, P. and Nie, J.-Y. and Wen, J.-R.},
  optauthor     = {Zhao, Wayne Xin and Zhou, Kun and Li, Junyi and Tang, Tianyi and Wang, Xiaolei and Hou, Yupeng and Min, Yingqian and Zhang, Beichen and Zhang, Junjie and Dong, Zican and Yifan Du and Chen Yang and Yushuo Chen and Zhipeng Chen and Jinhao Jiang and Ruiyang Ren and Yifan Li and Xinyu Tang and Zikang Liu and Peiyu Liu and Jian-Yun Nie and Ji-Rong Wen{\it et al}},
  optauthor     = {Wayne Xin Zhao and Kun Zhou and Junyi Li and Tianyi Tang and Xiaolei Wang and Yupeng Hou and Yingqian Min and Beichen Zhang and Junjie Zhang and Zican Dong and Yifan Du and Chen Yang and Yushuo Chen and Zhipeng Chen and Jinhao Jiang and Ruiyang Ren and Yifan Li and Xinyu Tang and Zikang Liu and Peiyu Liu and Jian-Yun Nie and Ji-Rong Wen},
  optjournal    = {arXiv preprint arXiv:2303.18223},
  volume        = {1},
  number        = {2},
  pages         = {1--124},
  optyear       = {2023},
  year          = {2026},
  eprint        = {2303.18223},
  archiveprefix = {arXiv},
  primaryclass  = {cs.CL},
  opturl        = {https://arxiv.org/abs/2303.18223}
}

@inproceedings{authorship,
  author    = {Kiermer, V. and Adams, Sofia   and  Bibbins-Domingo, Kirsten  and  Flores Bueso,Yensi  and Jamieson, Kathleen  Hall   and  Heber, Joerg  and  Hosseini, Mohammad  and Marušić, Ana   and Nielsen, Beau   and Skipper, Magdalena  and Swamy, Geeta K.   and Wolf, Susan M.  },
  optauthor = {Kiermer, V. and Adams, Sofia   and  Bibbins-Domingo, Kirsten  and  Flores Bueso,Yensi  {\it et al.}  },
  title     = {Creating a responsible authorship culture in science: Anchoring authorship practices in principles of transparency, credit, and accountability},
  booktitle= {Proc. Natl. Acad. Sci. U.S.A.},
  volume    = {123},
  OPTnumber    = {12},
  year      = {2026}
}

@incollection{foucault2003author,
  title     = {What is an Author?},
  author    = {Foucault, Michel},
  booktitle = {Reading architectural history},
  pages     = {71--81},
  year      = {2003},
  publisher = {Routledge}
}

@inproceedings{OT,
  author       = {Ellis, C. A. and Gibbs, S. J.},
  title        = {Concurrency control in groupware systems},
  year         = {1989},
  isbn         = {0897913175},
  optpublisher = {Association for Computing Machinery},
  opturl       = {https://doi.org/10.1145/67544.66963},
  optdoi       = {10.1145/67544.66963},
  booktitle    = {ACM SIGMOD International Conference on Management of Data},
  pages        = {399--407},
  optnumpages  = {9},
  optlocation  = {Portland, Oregon, USA},
  series       = {SIGMOD '89}
}

@article{BernersLeeVision,
  title   = {The Semantic Web},
  author  = {Berners-Lee, Tim and Hendler, James and Lassila, Ora},
  journal = {Scientific American},
  volume  = {284},
  number  = {5},
  pages   = {34--43},
  year    = {2001}
}

@article{SemwebVision,
  author  = {Shadbolt, N. and Berners-Lee, T. and Hall, W.},
  journal = {IEEE Intelligent Systems},
  title   = {The Semantic Web Revisited},
  year    = {2006},
  volume  = {21},
  number  = {3},
  pages   = {96-101}
}

@book{BaaderDL,
  place     = {Cambridge},
  author    = {Baader, Franz and Calvanese, Diego and McGuinness, 
               Deborah and Nardi, Daniele and Patel-Schneider, Peter},
  edition   = {2},
  title     = {The Description Logic Handbook: Theory, Implementation and Applications. 2nd ed.},
  publisher = {Cambridge University Press},
  year      = {2007}
}

@misc{rdfsemantics,
  author       = {Hayes, Patrick and Patel-Schneider, Peter F.},
  title        = {{RDF} 1.1 Semantics},
  howpublished = {W3C Recommendation},
  year         = {2014},
  url          = {https://www.w3.org/TR/rdf11-mt/}
}

@misc{RDF11,
  author       = {  Cyganiak, Richard  and    Wood, David  and 
                  Lanthaler, Markus },
  howpublished = {W3C Recommendation},
  title        = {{RDF 1.1 Concepts and Abstract Syntax}},
  url          = {http://www.w3.org/TR/2014/REC-rdf11-concepts-20140225/},
  year         = 2014
}

@misc{OWL2W,
  title        = {OWL 2 Web Ontology Language Primer (Second Edition)},
  author       = {Hitzler, Pascal  and Kr{\"o}tzsch, Markus  and Parsia, Bijan  and Patel-Schneider, Peter F.  and  Rudolph, Sebastian},
  howpublished = {W3C Recommendation},
  year         = {2012},
  url          = {https://www.w3.org/TR/owl2-primer/}
}

@article{semwebadoption,
  author   = {Haller, Armin and Polleres, Axel},
  editor   = {Pascal Hitzler and Krzysztof Janowicz},
  title    = {Are we better off with just one ontology on the Web?},
  year     = {2020},
  address  = {NLD},
  volume   = {11},
  number   = {1},
  issn     = {1570-0844},
  journal  = {Semantic Web},
  pages    = {87--99},
  numpages = {13}
}

@article{semwebhard,
  author   = {Hitzler, Pascal},
  title    = {A review of the semantic web field},
  year     = {2021},
  volume   = {64},
  number   = {2},
  journal  = {Communications of the ACM},
  pages    = {76--83},
  numpages = {8}
}

@article{hogan2021knowledge,
  author    = {Hogan, Aidan and Blomqvist, Eva and Cochez, Michael and D’amato, Claudia and Melo, Gerard De and Gutierrez, Claudio and Kirrane, Sabrina and Gayo, Jos\'{e} Emilio Labra and Navigli, Roberto and Neumaier, Sebastian and Ngomo, Axel-Cyrille Ngonga and Polleres, Axel and Rashid, Sabbir M. and Rula, Anisa and Schmelzeisen, Lukas and Sequeda, Juan and Staab, Steffen and Zimmermann, Antoine},
  title     = {Knowledge Graphs},
  year      = {2021},
  volume    = {54},
  number    = {4},
  issn      = {0360-0300},
  journal   = {ACM Comput. Surv.},
  articleno = {71}
}

@article{bizer2009linkeddata,
  title   = {Linked Data-The Story So Far},
  author  = {Bizer, Christian and Heath, Tom and Berners-Lee, Tim},
  journal = {International Journal on Semantic Web and Information Systems},
  volume  = {5},
  number  = {3},
  pages   = {1--22},
  year    = {2009}
}

@article{davis2015commonsense,
  title     = {Commonsense reasoning and commonsense knowledge in artificial intelligence},
  author    = {Davis, Ernest and Marcus, Gary},
  journal   = {Communications of the ACM},
  volume    = {58},
  number    = {9},
  pages     = {92--103},
  year      = {2015},
  publisher = {ACM New York, NY, USA}
}

@book{nelson1981literary,
  author    = {Nelson, Theodor Holm},
  title     = {Literary Machines},
  publisher = {Mindful Press},
  year      = {1981},
  address   = {Sausalito, CA}
}

@incollection{Zettelkasten,
  title     = {Kommunikation mit Zettelk{\"a}sten: Ein Erfahrungsbericht},
  author    = {Luhmann, Niklas},
  booktitle = {{\"O}ffentliche Meinung und sozialer Wandel/Public Opinion and Social Change},
  pages     = {222--228},
  year      = {1981},
  publisher = {Springer}
}

@inproceedings{derose1997problems,
  author    = {DeRose, Steven J. and Maden, Christopher R.},
  title     = {Problems with Dynamically Assembled Document Portions, 
               and Some Solutions},
  booktitle = {SGML/XML},
  year      = {1997}
}

@techreport{zittrain2021paper,
  author      = {Zittrain, Jonathan and Bowers, John and Stanton, Clare},
  title       = {The Paper of Record Meets an Ephemeral Web: 
                 An Examination of Linkrot and Content Drift 
                 within {The New York Times}},
  institution = {Library Innovation Lab, Harvard Law School},
  year        = {2021}
}

@phdthesis{zhu2025linkrot,
  author = {Zhu, Jingyuan},
  title  = {Solutions for Link Rot on the Modern Web},
  school = {University of Michigan},
  year   = {2025},
  type   = {PhD dissertation}
}

@article{haklay2008osm,
  title   = {OpenStreetMap: User-Generated Street Maps},
  author  = {Haklay, Mordechai and Weber, Patrick},
  journal = {IEEE Pervasive Computing},
  volume  = {7},
  number  = {4},
  pages   = {12--18},
  year    = {2008}
}

@misc{osm_api,
  author = {{OpenStreetMap Contributors}},
  title  = {{OpenStreetMap API Usage Policy}},
  url    = {https://wiki.openstreetmap.org/wiki/API},
  year   = {2024}
}

@article{neis2012analyzing,
  title     = {Analyzing the contributor activity of a volunteered geographic information project--The case of OpenStreetMap},
  author    = {Neis, Pascal and Zipf, Alexander},
  journal   = {ISPRS International Journal of Geo-Information},
  volume    = {1},
  number    = {2},
  pages     = {146--165},
  year      = {2012},
  publisher = {Molecular Diversity Preservation International}
}

@article{osmcomplete,
  author    = {Barrington-Leigh, Christopher AND Millard-Ball, Adam},
  journal   = {PLOS ONE},
  publisher = {Public Library of Science},
  title     = {The world's user-generated road map is more than 80\% complete},
  year      = {2017},
  volume    = {12},
  pages     = {1--20},
  number    = {8}
}

@article{mooney2012characteristics,
  author    = {Mooney, Peter and Corcoran, Padraig},
  title     = {Characteristics of heavily edited objects in                {OpenStreetMap}},
  journal   = {Future Internet},
  volume    = {4},
  number    = {1},
  pages     = {285--305},
  year      = {2012},
  publisher = {MDPI},
  optnote   = {Concurrent edits to the same elements are common in practice (e.g. multiple contributors mapping the same road This paper directly studies heavily edited OSM objects -- roads, buildings -- and documents the frequency and patterns of concurrent editing, which supports your claim.}
}

@inproceedings{osm_rdf,
  author    = {Auer, S{\"o}ren and Lehmann, Jens and Hellmann, Sebastian},
  editor    = {Bernstein, Abraham
               and Karger, David R.
               and Heath, Tom
               and Feigenbaum, Lee
               and Maynard, Diana
               and Motta, Enrico
               and Thirunarayan, Krishnaprasad},
  title     = {LinkedGeoData: Adding a Spatial Dimension to the Web of Data},
  booktitle = {The Semantic Web - ISWC},
  year      = {2009},
  publisher = {Springer Berlin Heidelberg},
  pages     = {731--746},
  isbn      = {978-3-642-04930-9}
}

@inproceedings{Hristova2013MappingParties,
  author    = {Hristova, Desislava and Quattrone, Giovanni and Mashhadi, Afra and Capra, Licia},
  title     = {The Life of the Party: Impact of Social Mapping in {OpenStreetMap}},
  booktitle = {International AAAI Conference on Web and Social Media (ICWSM)},
  volume    = {7},
  pages     = {234--243},
  year      = {2013}
}

@misc{osm_mapping_parties,
  author  = {OpenStreetMap Wiki},
  title   = {Mapping parties --- OpenStreetMap Wiki{,} },
  year    = {2023},
  url     = {https://wiki.openstreetmap.org/w/index.php?title=Mapping_parties&oldid=2615609},
  optnote = {[Online; accessed 5-May-2026]}
}

@book{wikipediaGoodFaith,
  author    = {Reagle, Joseph},
  title     = {Good Faith Collaboration: The Culture of Wikipedia},
  publisher = {The MIT Press},
  year      = {2010},
  month     = {08},
  isbn      = {9780262289719}
}

@inproceedings{teblunthuis2018revisiting,
  title     = {Revisiting "The rise and decline" in a population of peer production projects},
  author    = {TeBlunthuis, Nathan and Shaw, Aaron and Hill, Benjamin Mako},
  booktitle = {CHI Conference on Human Factors in Computing Systems},
  pages     = {1--7},
  year      = {2018}
}

@article{Wikipedia,
  journal = {Journal of the Association for Information Science \& Technology},
  author  = {Mesgari, Mostafa  and Okoli, Chitu  and Mehdi, Mohamad  and Nielsen, Finn A.  and Lanam\"aki, Arto },
  title   = {The sum of all human knowledge: A systematic review of scholarly research on the content of Wikipedia},
  year    = {2015},
  pages   = {219-245},
  volume  = {66},
  number  = {2}
}

@article{raymond1999cathedral,
  author    = {Raymond, Eric S.},
  title     = {The Cathedral and the Bazaar: Musings on {Linux} and 
               Open Source by an Accidental Revolutionary},
  journal   = {Knowledge, Technology \& Policy},
  volume    = {12},
  number    = {3},
  pages     = {23--49},
  year      = {1999},
  publisher = {Springer}
}

@article{Schneider2019Decentralization,
  author    = {Nathan Schneider},
  title     = {Decentralization: an incomplete ambition},
  journal   = {Journal of Cultural Economy},
  volume    = {12},
  number    = {4},
  pages     = {265--285},
  year      = {2019},
  publisher = {Routledge}
}

@article{olson2000distance,
  title     = {Distance matters},
  author    = {Olson, Gary M and Olson, Judith S},
  journal   = {Human--computer interaction},
  volume    = {15},
  number    = {2-3},
  pages     = {139--178},
  year      = {2000},
  publisher = {Taylor \& Francis}
}

@article{benkler2006commons,
  title   = {The Wealth of Networks: How Social Production Transforms 
             Markets and Freedom},
  author  = {Benkler, Yochai},
  journal = {Yale University Press},
  year    = {2006}
}

@book{ostrom1990governing,
  author    = {Ostrom, Elinor},
  title     = {Governing the Commons: The Evolution of Institutions 
               for Collective Action},
  publisher = {Cambridge University Press},
  year      = {1990},
  address   = {Cambridge}
}

@techreport{ibis1970,
  title       = {Issues as Elements of Information Systems  (Working paper)},
  author      = {Kunz, Werner and Rittel, Horst W.},
  institution = {Berkeley: Institute of Urban and Regional Development},
  year        = {1970}
}

@article{chesnevar2006aif,
  title   = {Towards an Argument Interchange Format},
  author  = {Chesnevar, Carlos Iv\'an and McGinnis, Jarred and Modgil, Sanjay and Rahwan, Iyad and Reed, Chris and Simari, Guillermo and South, Matthew and Vreeswijk, Gerard and Willmott, Steven},
  journal = {The Knowledge Engineering Review},
  volume  = {21},
  number  = {4},
  pages   = {293--316},
  year    = {2006}
}

@article{scheuer2010argument,
  title   = {Computer-supported argumentation: A review of the state of the art},
  author  = {Scheuer, Oliver and Loll, Frank and Pinkwart, Niels and McLaren, Bruce M.},
  journal = {Computer Supported Learning},
  volume  = {5},
  pages   = {43--102},
  year    = {2010}
}

@techreport{shapiro2011crdt,
  title       = {{A comprehensive study of Convergent and Commutative Replicated Data Types}},
  author      = {Shapiro, Marc and Pregui{\c c}a, Nuno and Baquero, Carlos and Zawirski, Marek},
  opttype     = {Research Report},
  number      = {RR-7506},
  pages       = {50},
  institution = {{INRIA}},
  year        = {2011}
}

@article{kleppmann2017conflict,
  author  = {Kleppmann, Martin and Beresford, Alastair R.},
  title   = {A Conflict-Free Replicated {JSON} Datatype},
  journal = {IEEE Transactions on Parallel and Distributed Systems},
  volume  = {28},
  number  = {10},
  pages   = {2733--2746},
  year    = {2017}
}

@book{latour1987science,
  title     = {Science in action: How to follow scientists and engineers through society},
  author    = {Latour, Bruno},
  year      = {1987},
  publisher = {Harvard university press}
}

@book{latour1986laboratory,
  title     = {Laboratory Life: The Construction of Scientific Facts},
  author    = {Latour, Bruno and Woolgar, Steve},
  year      = {1986},
  publisher = {Princeton University Press}
}

@book{kuhn1962structure,
  title     = {The Structure of Scientific Revolutions},
  author    = {Kuhn, Thomas S.},
  year      = {1962},
  publisher = {University of Chicago Press}
}

@book{fleck1981genesis,
  title     = {Genesis and development of a scientific fact},
  author    = {Fleck, Ludwik},
  year      = {1981},
  publisher = {University of Chicago Press},
  note      = {Originally published in German, 1935. Translated by F. Bradley and T. J. Trenn}
}

\end{document}